\documentclass{article} 
\usepackage{iclr2026_conference,times}

\usepackage{amsmath,amsfonts,bm}

\def\eqref#1{equation~\ref{#1}}

\def\1{\bm{1}}

\DeclareMathAlphabet{\mathsfit}{\encodingdefault}{\sfdefault}{m}{sl}
\SetMathAlphabet{\mathsfit}{bold}{\encodingdefault}{\sfdefault}{bx}{n}

\DeclareMathOperator*{\argmax}{arg\,max}

\usepackage{hyperref}
\usepackage{url}
\usepackage{times,latexsym}
\usepackage{hyperref}
\usepackage{booktabs}
\usepackage{multirow}
\usepackage{amsmath}
\usepackage{amssymb}
\usepackage{graphicx}
\usepackage{bbding}
\usepackage{arydshln}
\usepackage{wrapfig}
\usepackage{xcolor}         
\definecolor{ao}{rgb}{0.0, 0.5, 0.0}
\definecolor{forestgreen}{RGB}{0, 150, 0}

\title{Bridging Fairness and Explainability: Can Input-Based Explanations Promote Fairness in Hate Speech Detection?}

\author{\textbf{Yifan Wang\textsuperscript{1}, }
        \textbf{Mayank Jobanputra\textsuperscript{1}, }
        \textbf{Ji-Ung Lee\textsuperscript{1}, }
        \textbf{Soyoung Oh\textsuperscript{1}, }
        \\
        \textbf{Isabel Valera\textsuperscript{1,3}, }
        \textbf{Vera Demberg\textsuperscript{1,2}} 
        \\
        \textsuperscript{1} Saarland University, Germany \\
        \textsuperscript{2} Max Planck Institute for Informatics, Saarland Informatics Campus, Germany \\
        \textsuperscript{3} Max Planck Institute for Software Systems, Saarland Informatics Campus, Germany \\
        \texttt{\{yifwang,mayank,soyoung,vera\}@lst.uni-saarland.de}, \\
        \texttt{ji-ung.lee@uni-saarland.de},
        \texttt{ivalera@cs.uni-saarland.de}
}

\iclrfinalcopy 
\begin{document}

\maketitle

\begin{abstract}
Natural language processing (NLP) models often replicate or amplify social bias from training data, raising concerns about fairness.
At the same time, their black-box nature makes it difficult for users to recognize biased predictions and for developers to effectively mitigate them.
While some studies suggest that input-based explanations can help detect and mitigate bias, others question their reliability in ensuring fairness.
Existing research on explainability in fair NLP has been predominantly qualitative, with limited large-scale quantitative analysis.
In this work, we conduct the first systematic study of the relationship between explainability and fairness in hate speech detection, focusing on both encoder- and decoder-only models.
We examine three key dimensions: (1) identifying biased predictions, (2) selecting fair models, and (3) mitigating bias during model training.
Our findings show that input-based explanations can effectively detect biased predictions and serve as useful supervision for reducing bias during training, but they are unreliable for selecting fair models among candidates.
Our code is available at \href{https://github.com/Ewanwong/fairness_x_explainability}{https://github.com/Ewanwong/fairness\_x\_explainability}.
\end{abstract}

\section{Introduction}

Language models (LMs) pre-trained on large-scale natural language datasets have 
shown great capacities in various NLP tasks~\citep{wang-etal-2018-glue, Eval-2023-Harness}.
However, previous studies have shown that they can replicate and amplify stereotypes and social bias present in their training data and demonstrate biased behaviors~\citep{sheng-etal-2021-societal, gupta2024calm, gallegos-etal-2024-bias}.
Such behaviors risk the underrepresentation of marginalized groups and the unfair allocation of resources, raising serious concerns in critical applications~\citep{blodgett-etal-2020-language}.

Meanwhile, current NLP models are mostly based on black-box neural networks.
Despite their strong capacities, the complex architecture and large number of parameters of these models make it hard for humans to understand their behaviors~\citep{Bommasani-2021-FoundationModels}.
To understand neural NLP models, different types of explanations have been devised, such as input-based explanations~\citep{yin-neubig-2022-interpreting, deiseroth-etal-2023-atman, madsen-etal-2024-self, bcos-wang-2025}, natural language explanations~\citep{ramnath2024tailoring, wang-etal-2025-cross}, and concept-based explanations~\citep{yu-etal-2024-latent, raman-etal-2024-understanding}.
Among these, input-based explanations, often referred to as rationales, indicate the contribution of each token to models' predictions, and thus provide the most direct insights into models' behaviors~\citep{arras-etal-2019-evaluating, atanasova2022diagnostics, lyu-etal-2024-towards}.

Explainability has long been deemed critical to improving fairness. 
Researchers believe that if the use of sensitive features is evidenced by model explanations, then they can easily detect biased predictions and impose fairness constraints by guiding models to avoid such faulty reasoning~\citep{meng2022interpretability, sogancioglu-etal-2023-using-explainability}. 
However, recent studies have challenged this assumption, suggesting that the relationship between explainability and fairness is complex and that explanations may not always reliably detect or mitigate bias~\citep{dimanov2020you, fooling-lime, pruthi-etal-2020-learning}.
Unfortunately, to the best of our knowledge, current studies are mostly limited to qualitative analysis on a small set of explanation methods~\citep{balkir-etal-2022-challenges, deck-etal-2024-critical-survey}. 
Our work takes a step toward bridging explainability and fairness by providing the first comprehensive quantitative analysis in the context of hate speech detection, a task where both fairness and explainability are particularly critical.
Specifically, we address the following three research questions to investigate the role of explainability in promoting fairness within the task of hate speech detection:
\begin{itemize}
    \item \textbf{RQ1: Can input-based explanations be used to identify biased predictions?}
    \item \textbf{RQ2: Can input-based explanations be used to automatically select fair models?}
    \item \textbf{RQ3: Can input-based explanations be used to mitigate bias during model training?}
\end{itemize}
Our experiments demonstrate that input-based explanations can effectively detect biased predictions (RQ1), are less reliable for automatic fair model selection (RQ2), and can help reduce bias during model training (RQ3).
Furthermore, our analyses indicate that explanation-based bias detection remains robust even when models are trained to reduce reliance on sensitive features, and that these explanations outperform LLM judgments in identifying bias.

\section{Related Work}

\paragraph{Bias in NLP}

The presence of social bias and stereotypes has significantly shaped human language and LMs trained on it~\citep{blodgett-etal-2020-language, sheng-etal-2021-societal}. 
As a result, these models often exhibit biased behaviors~\citep{gallegos-etal-2024-bias}, such as stereotypical geographical relations in the embedding space~\citep{bolukbasi2016man, may2019measuring} and stereotypical associations between social groups and certain concepts in the model outputs~\citep{fang2024bias, wan-chang-2025-white}. 
More critically, disparities in model predictions and performance across social groups~\citep{zhao-etal-2018-gender, sheng-etal-2019-woman} can significantly compromise user experiences of marginalized groups and risk amplifying bias against them, therefore drawing great concerns in critical use cases. 


\paragraph{Input-based Model Explanations}

Input-based explanations in NLP models aim to attribute model predictions to each input token~\citep{lyu-etal-2024-towards}.
They can be broadly categorized based on how they generate explanations: gradient-based~\citep{simonyan-etal-2014, kindermans-etal-2016, Sundararajan-2017-IntegratedGrad, enguehard-2023-sequential}, propagation-based~\citep{bach2015pixel, shrikumar2017learning, ferrando-etal-2022-measuring, modarressi-etal-2022-globenc, modarressi-etal-2023-decompx}, perturbation-based~\citep{occlusion, Ribeiro-2016-LIME, Lundberg-2017-SHAP, deiseroth-etal-2023-atman}, and attention-based methods~\citep{attention, abnar-zuidema-2020-quantifying}. 
While most prior work has focused on encoder-only models, recent studies have also explored explaining the behaviors of generative models~\citep{yin-neubig-2022-interpreting, ferrando-etal-2022-measuring, enouen-etal-2024-textgenshap, cohen-etal-2024-contextcite}.


\paragraph{Bridging Explainability and Fairness}

Explainability is often considered essential for achieving fairness in machine learning systems~\citep{balkir-etal-2022-challenges, deck-etal-2024-critical-survey}. 
One line of research investigates model bias by analyzing explanations~\citep{prabhakaran-etal-2019-perturbation, jeyaraj-delany-2024-explainable, sogancioglu-etal-2023-using-explainability}. 
For instance, \citet{muntasir-etal-2025-explainable-ai} shows that a biased model relied on gendered words as key features in its predictions, as revealed by LIME explanations.
Similarly, \citet{stevens-etal-2020-explainability-and-fairness} demonstrates that biased models often place high importance on gender and race features when examined with SHAP explanations. 
Extending this line of evidence, \citet{meng2022interpretability} finds that features with higher importance scores are associated with larger disparities in model performance on a synthetic medical dataset using deep learning models.

Another line of research focuses on mitigating bias with explanations~\citep{dimanov2020you, kennedy-etal-2020-contextualizing, rao2023studying, devil-liu-2024}.
For example, \citet{hickey2020fairness} improves fairness by reducing reliance on sensitive features during training with SHAP explanations.
\citet{bhargava2020limeout} and \citet{enhancing-santiago-2025} first identify predictive sensitive features using LIME and SHAP, respectively, and then remove them prior to model training.
Relatedly, \citet{grabowicz-etal-2022-marrying} traces unfairness metrics to input features and adjusts them to mitigate bias.

However, recent research has challenged the assumption that input-based explanations can be reliably used to detect and mitigate bias.
First, current explanation methods may be unfaithful, meaning that they may not always reflect the true decision-making process of models~\citep{jain-wallace-2019-attention, ye-etal-2025-input}. 
This makes it difficult to reliably detect the use of sensitive features in predictions.
Second, efforts to reduce the influence of sensitive features can lead to unintended consequences, sometimes degrading both task performance and fairness of models~\citep{dimanov2020you}.
Finally, models can be deliberately trained to assign lower importance to sensitive features, thereby masking biased predictions when explanations are inspected~\citep{fooling-lime, pruthi-etal-2020-learning}.


Despite growing interest in this topic, most existing work remains qualitative or restricted to limited setups~\citep{balkir-etal-2022-challenges, deck-etal-2024-critical-survey}.
To the best of our knowledge, this is the first study to quantitatively and comprehensively examine the relationship between explainability and fairness in NLP models.
We focus on hate speech detection as a particularly critical application.
Prior research has shown that biased NLP models often rely on demographic information such as race and gender, leading to inferior performance on marginalized groups in this task~\citep{sap-etal-2019-risk, hatexplain}. 
Detecting and mitigating such biased behaviors are therefore essential for equitable social media participation across all groups.
We define hate speech and social bias and review related fairness and explainability work in hate speech detection in Appendix~\ref{appendix:definition_and_motivation}.


\begin{figure}[t]
    \centering
    \includegraphics[width=\linewidth]{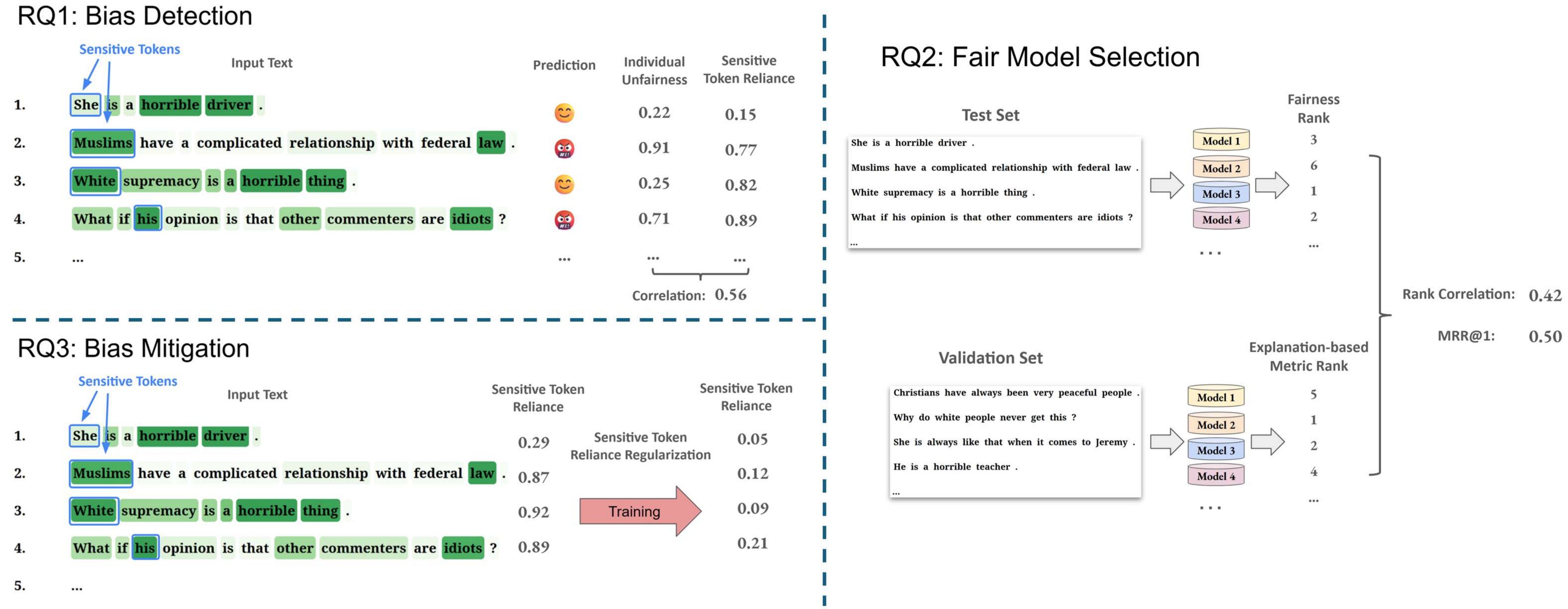}
    \caption{Workflow diagram illustrating the processes used to address each research question. Sensitive tokens are shown in \textcolor{blue}{blue} boxes, and the intensity of the \textcolor{forestgreen}{green} shading reflects each word’s contribution to the model’s prediction.}
    \label{fig:workflow}
\end{figure}
\section{Experimental Setup}

\paragraph{Notations} 



Let an input text $\mathbf{x}$ consist of tokens ${t_1, t_2, \dots, t_n}$. 
The task of hate speech detection is to predict a binary label $\hat{y} \in \{\text{toxic}, \text{non-toxic}\}$. 
A classifier outputs the probability of class $c$ as $f_c(x)$, where $f$ is implemented by a neural model.

In the context of social bias, we assume that a bias type (e.g., race) involves a set of social groups $G$ (e.g., ${\text{black}, \text{white}}, ...$). 
A subset of tokens ${t_{g_1}, t_{g_2}, \dots, t_{g_m}}$ in $\mathbf{x}$ denotes the sensitive feature $g \in G$ of the speaker or target. 
We refer to these tokens as \textit{sensitive tokens}.
By replacing the sensitive tokens of group $g$ with those of another group $g'$, we obtain a counterfactual version of $\mathbf{x}$ that refers to $g'$, denoted as $\mathbf{x}^{(g')}$.

An input-based explanation assigns an attribution score to each token in $\mathbf{x}$ for class $c$: ${a_1^{c}, a_2^{c}, \dots, a_n^{c}}$, indicating their contribution to the prediction of class $c$. 
Following~\citet{dimanov2020you}, we compute attribution scores on the sensitive tokens, ${a_{g_1}^{c}, a_{g_2}^{c}, \dots, a_{g_m}^{c}}$, which we refer to as the \textit{sensitive token reliance} scores.
To handle cases where multiple sensitive tokens appear in the same sentence, we take the maximum absolute attribution value as the reliance score for that example\footnote{We have experimented with normalizing feature importance scores but found that using raw scores yielded the best results. We also evaluated sum and average aggregation methods beyond taking the max absolute value and observed similar outcomes.}:
$$
\text{sensitive token reliance}(\mathbf{x}, c) = a^{c}_{j^*}, 
\text{where } 
j^* = \argmax_{j \in \{g_1, \ldots, g_m\}} \left| a^{c}_{j} \right|
$$
\paragraph{Datasets and Vocabulary} We use two hate speech detection datasets: Civil Comments~\citep{borkan2019nuanced} and Jigsaw~\citep{jigsaw-unintended-bias-in-toxicity-classification}. 
To ensure coverage, we focus on three bias types and their associated groups: race (black/white), gender (female/male), and religion (Christian/Muslim/Jewish). 
We include examples containing identity-marking terms but exclude those with derogatory or slur-based references, as the latter can reasonably serve as direct evidence for toxic predictions.
The sensitive token vocabulary is derived from~\citet{caliskan2017semantics} and~\citet{wang-demberg-2024-parameter}. 
Further details on dataset pre-processing are provided in Appendix~\ref{appendix: experimental details}.

\paragraph{Models} 
We evaluate two major classes of NLP models: encoder-only models (BERT~\citep{devlin-etal-2019-bert} and RoBERTa~\citep{roberta}) and decoder-only large language models (Llama3.2-3B-Instruct~\citep{llama3}, Qwen3-4B, and Qwen3-8B~\citep{qwen3}, all of which are aligned to human values).
We fine-tune encoder-only models on data subsets that either target a single bias type or combine all bias types.
For decoder-only models, we use an instruction-based setup where the model is prompted to decide whether a test example contains hate speech. 
The prompt includes the definition of hate speech, the test example, and a corresponding question.
As a baseline, we adopt the zero-shot setting as the default configuration.

Beyond conventional fine-tuning and prompting, we also investigate the interaction between explainability and fairness in debiased models.
For encoder-only models, we apply pre-processing techniques such as group balance~\citep{kamiran2012data}, group-class balance~\citep{dixon-measuring-and-mitigating-2018}, and counterfactual data augmentation (CDA, \citealp{zmigrod-etal-2019-counterfactual}), as well as in-processing techniques including dropout~\citep{dropout-debiasing}, attention entropy~\citep{attanasio-etal-2022-entropy}, and causal debias~\citep{zhou-etal-2023-causal}.
For decoder-only models, we incorporate bias reduction through prompt design, including few-shot, fairness imagination~\citep{chen-etal-2025-causally}, and fairness instruction prompting~\citep{chen-etal-2025-causally}.
We do not include reasoning models and chain-of-thought prompting, as we find that their predictions are primarily attributed to intermediate reasoning steps rather than the input text, which complicates analysis and falls beyond the scope of this work.
Further details are provided in Appendix~\ref{appendix: experimental details}.

\paragraph{Fairness Metrics} 
We evaluate fairness in model predictions using two categories of metrics: \textbf{group fairness} and \textbf{individual fairness}.
Group fairness metrics capture disparities in performance across demographic groups:
\begin{align*}
    \text{Disp}_\text{metric} = \sum_{g\in G} |\text{metric}_g - \overline{\text{metric}_{G}}| ,
\end{align*}
where $\overline{\text{metric}_{G}}$ is the average metric value across all groups $G$ in a bias type. 
We specifically measure disparities in accuracy (ACC), false positive rate (FPR), and false negative rate (FNR).

Individual fairness measures the extent to which a model’s prediction for a given example changes when the associated social group is altered. 
To maintain consistency with the direction of group fairness metrics, we compute the individual unfairness (IU) score of $\mathbf{x}_i$ and the predicted class $\hat{y_i}$:
\begin{align*}
    \text{IU}(\mathbf{x}_i)= |f_{\hat{y}_i}(\mathbf{x}_i) - \frac{1}{|G \backslash \{g_i\}|} \sum_{g'\in G \backslash \{g_i\}} f_{\hat{y}_i}(\mathbf{x}_i^{(g')})|
\end{align*}
The Average IU score ($\text{Avg}_{\text{iu}}$) is then computed over a dataset to reflect the overall level of individual unfairness in a model.

For both types of metrics, higher scores indicate more bias in model predictions. 
It is worth noting that individual unfairness can be evaluated at the level of each example, whereas group fairness metrics are defined over sets of validation or test examples.
To compute the fairness metrics, we randomly sample a subset of examples for each bias type such that each social group contributes an equal number of examples. 
Further details on test set sampling are provided in Appendix~\ref{appendix: experimental details}.

\paragraph{Explanation Methods} 
We employ 16 variants of commonly used input-based post-hoc explanation methods, selected to represent a diverse range of methodological categories: Attention~\citep{attention}, Attention rollout (Attn rollout,~\citealp{abnar-zuidema-2020-quantifying}), Attention flow (Attn flow,~\citealp{abnar-zuidema-2020-quantifying}), Gradient (Grad, ~\citealp{simonyan-etal-2014}), Input x Gradient (IxG,~\citealp{kindermans-etal-2016}), Integrated Gradients (IntGrad, ~\citealp{Sundararajan-2017-IntegratedGrad}), Occlusion~\citep{occlusion}, DeepLift~\citep{shrikumar2017learning}, KernelSHAP~\citep{Lundberg-2017-SHAP}, DecompX~\citep{modarressi-etal-2023-decompx}, and Progressive Inference (ProgInfer,~\citealp{pmlr-v235-kariyappa24a})\footnote{We apply DecompX only to encoder-only models and Progressive Inference only to decoder-only models, following the setups of the original papers.}.
For methods that attribute predictions to embeddings, we aggregate attribution scores into a single feature importance value using either the mean or the L2 norm. 
For Occlusion, we additionally report results obtained by taking the absolute value of each attribution score prior to computing sensitive token reliance scores (denoted as Occlusion abs).
The time and GPU memory costs for each method are shown in Appendix~\ref{appendix:explanation-efficiency}.
We also study rationales generated by LLMs and find that they are not as reliable as input-based explanations in detecting bias (Section \ref{sec:llm}).
\begin{table*}[!h]
    \centering
         \caption{Task performance and fairness of default and debiased models on Civil Comments.
     Results are provided for race/gender/religion biases.
    \textcolor{forestgreen}{Green} (\textcolor{red}{red}) indicates the results are \textcolor{forestgreen}{better} (\textcolor{red}{worse}) than the default/zero-shot models. No debiasing method consistently reduces bias across all metrics and bias types.\\}
     
     \resizebox{\textwidth}{!}{
     
     \begin{tabular}{llccccc}
     \toprule
        Model  & Method & Accuracy ($\uparrow$) & $\text{Disp}_\text{acc} (\downarrow)$ & $\text{Disp}_\text{fpr} (\downarrow)$ & $\text{Disp}_\text{fnr} (\downarrow)$ & $\text{Avg}_\text{iu} (\downarrow)$ \\
    \midrule
    \multirow{7}{*}{\shortstack{BERT}} & Default & 78.38/88.05/85.93 & 2.05/3.30/18.07 & 0.50/0.03/5.77 & 10.04/11.98/30.9  & 3.17/0.66/1.27   \\
     & Group balance & \textcolor{forestgreen}{79.25}/\textcolor{red}{87.25}/\textcolor{forestgreen}{86.83} & \textcolor{red}{3.10}/\textcolor{forestgreen}{2.80}/\textcolor{forestgreen}{13.53} & \textcolor{forestgreen}{0.25}/\textcolor{red}{1.73}/\textcolor{red}{11.53} & \textcolor{red}{10.46}/\textcolor{forestgreen}{5.38}/\textcolor{forestgreen}{30.31} & \textcolor{red}{3.79}/\textcolor{forestgreen}{0.42}/\textcolor{red}{2.01}  \\
     & Group-class balancing & \textcolor{red}{78.00}/\textcolor{red}{87.02}/\textcolor{red}{85.77} & \textcolor{forestgreen}{1.80}/\textcolor{forestgreen}{2.75}/\textcolor{forestgreen}{14.73} & \textcolor{red}{2.42}/\textcolor{red}{0.99}/\textcolor{forestgreen}{3.09} & \textcolor{red}{10.63}/\textcolor{forestgreen}{7.26}/\textcolor{red}{33.14} & \textcolor{red}{4.43}/\textcolor{red}{0.98}/\textcolor{forestgreen}{0.71}   \\
     & CDA & \textcolor{red}{76.83}/\textcolor{red}{86.70}/\textcolor{red}{84.83} & \textcolor{red}{2.35}/\textcolor{red}{3.60}/\textcolor{forestgreen}{14.13} & \textcolor{red}{5.88}/\textcolor{red}{2.00}/\textcolor{forestgreen}{5.67} & \textcolor{red}{18.45}/\textcolor{forestgreen}{7.57}/\textcolor{forestgreen}{24.12} & \textcolor{forestgreen}{0.50}/\textcolor{forestgreen}{0.50}/\textcolor{forestgreen}{0.90}  \\
     & Dropout & \textcolor{forestgreen}{78.53}/\textcolor{forestgreen}{88.20}/\textcolor{red}{85.03} & \textcolor{red}{2.25}/\textcolor{forestgreen}{2.10}/\textcolor{forestgreen}{15.67} & \textcolor{red}{0.78}/\textcolor{red}{1.46}/\textcolor{red}{5.93} & \textcolor{red}{10.82}/\textcolor{forestgreen}{3.50}/\textcolor{forestgreen}{27.16} & \textcolor{red}{3.43}/\textcolor{forestgreen}{0.52}/\textcolor{red}{1.51}   \\
     & Attention entropy &  \textcolor{forestgreen}{79.15}/\textcolor{red}{87.67}/\textcolor{red}{84.93} & \textcolor{red}{2.60}/\textcolor{forestgreen}{2.05}/\textcolor{forestgreen}{15.07} & \textcolor{red}{0.99}/\textcolor{red}{0.10}/\textcolor{forestgreen}{4.99} & \textcolor{red}{11.71}/\textcolor{forestgreen}{7.11}/\textcolor{forestgreen}{26.52} & \textcolor{forestgreen}{2.95}/\textcolor{red}{0.67}/\textcolor{red}{1.58}  \\
     & Causal debias & \textcolor{forestgreen}{78.80}/\textcolor{red}{86.17}/\textcolor{forestgreen}{86.40} & \textcolor{forestgreen}{0.00}/\textcolor{forestgreen}{2.65}/\textcolor{forestgreen}{16.40} & \textcolor{red}{3.90}/\textcolor{red}{0.46}/\textcolor{red}{8.82} & \textcolor{forestgreen}{7.98}/\textcolor{forestgreen}{10.67}/\textcolor{forestgreen}{30.46} & \textcolor{red}{3.83}/\textcolor{forestgreen}{0.48}/\textcolor{red}{2.10}   \\
    \midrule
    \multirow{4}{*}{Qwen3-4B} & Zero-shot & 69.55/79.75/77.50 & 0.60/0.00/17.40 & 7.13/1.40/21.07 & 13.25/3.71/5.17 & 2.55/2.41/3.32 \\
     & Few-shot & \textcolor{forestgreen}{70.15}/\textcolor{forestgreen}{80.73}/\textcolor{forestgreen}{79.53} & \textcolor{red}{1.80}/\textcolor{red}{0.65}/\textcolor{red}{18.93} & \textcolor{red}{10.02}/\textcolor{red}{2.50}/\textcolor{forestgreen}{19.31} & \textcolor{forestgreen}{11.89}/\textcolor{red}{9.15}/\textcolor{red}{5.57} & \textcolor{red}{3.18}/\textcolor{red}{3.34}/\textcolor{red}{3.76}  \\
     & Fairness imagination & \textcolor{forestgreen}{71.23}/\textcolor{forestgreen}{80.40}/\textcolor{forestgreen}{80.83} & \textcolor{red}{0.85}/\textcolor{red}{1.00}/\textcolor{red}{18.27} & \textcolor{forestgreen}{4.03}/\textcolor{red}{2.11}/\textcolor{forestgreen}{10.51} & \textcolor{forestgreen}{11.62}/\textcolor{red}{9.21}/\textcolor{forestgreen}{4.28} & \textcolor{red}{2.98}/\textcolor{red}{3.16}/\textcolor{forestgreen}{2.20} 
  \\ 
     & Fairness instruction & \textcolor{forestgreen}{70.40}/\textcolor{forestgreen}{79.77}/\textcolor{forestgreen}{80.47} & 0.60/\textcolor{red}{1.35}/\textcolor{red}{19.33} & \textcolor{forestgreen}{4.30}/\textcolor{forestgreen}{0.39}/\textcolor{forestgreen}{4.67} & \textcolor{forestgreen}{11.11}/\textcolor{red}{5.24}/\textcolor{forestgreen}{5.08} & \textcolor{forestgreen}{2.02}/\textcolor{forestgreen}{1.83}/\textcolor{forestgreen}{1.71} 
  \\

    \bottomrule
     \end{tabular}
     }

     \label{tab:performance and fairness}
\end{table*}

\section{Quantitative Analyses of Fairness and Explainability}

To comprehensively understand the relationship between explainability and fairness in NLP models, we examine three ways in which model explanations can be applied to promote fairness. 
The subsequent sections detail the experimental setups for each application and report the corresponding results. 
The workflow for our research questions is shown in Figure~\ref{fig:workflow}.  
For brevity, we report results on Civil Comments using BERT trained on single bias types and Qwen3-4B. 
We further demonstrate that our findings extend to other models and the Jigsaw dataset (Appendix~\ref{appendix: performance and fairness}–\ref{appendix: rq3}), and that they generalize across tasks, model alignment types, and sensitive token sets beyond those considered in the main experiments (Appendix~\ref{appendix:generalization}).

\subsection{Model Performance and Fairness}
\label{sec: fairness and performance}
As a prerequisite, we first summarize the performance and fairness of the evaluated models. 
The results in Table~\ref{tab:performance and fairness} show that no single debiasing method consistently improves all fairness metrics. For BERT and Qwen3-4B, CDA and fairness instruction achieve the largest reductions in individual unfairness, yet they may simultaneously amplify biases on other metrics.
Other debiasing methods show a similar pattern: they reduce bias for a specific metric or bias type, but the improvement does not generalize across different setups. 
These limitations underscore the importance of leveraging explanations for bias detection and mitigation.
We find similar results for other models and for Jigsaw, which we provide in Appendix~\ref{appendix: performance and fairness} along with a discussion on model performance and fairness. 

\subsection{RQ1: Explanations For Bias Detection}
\label{sec:rq1}
Our first research question asks whether explanations can be used to detect biased predictions. 
We address the question through three steps: (1) obtain model predictions and compute individual unfairness scores; (2) generate input-based explanations for the predictions; and (3) compute sensitive token reliance scores and evaluate their Pearson correlation with individual unfairness, which we refer to as \textit{fairness correlation}.
A higher fairness correlation indicates that the explanation method is more effective in identifying predictions with high individual unfairness. 
To ensure robustness, we compute the fairness correlation separately for each prediction class–group pair and report the average absolute score as the final result for each explanation method.

We present results for default and debiased models where individual unfairness remains high after debiasing, as bias detection is particularly critical in these cases.
Specifically, we report results for models with the highest average $\text{Avg}_\text{iu}$ scores across bias types, namely default, group balance, and causal debias for BERT, and zero-shot, few-shot, and fairness imagination prompting for Qwen3-4B. 
Results for religion as well as other models and the Jigsaw dataset are provided in Appendix~\ref{appendix: rq1}.

\paragraph{Results}
\begin{figure*}[h!]  
    \centering
    \includegraphics[width=\linewidth]{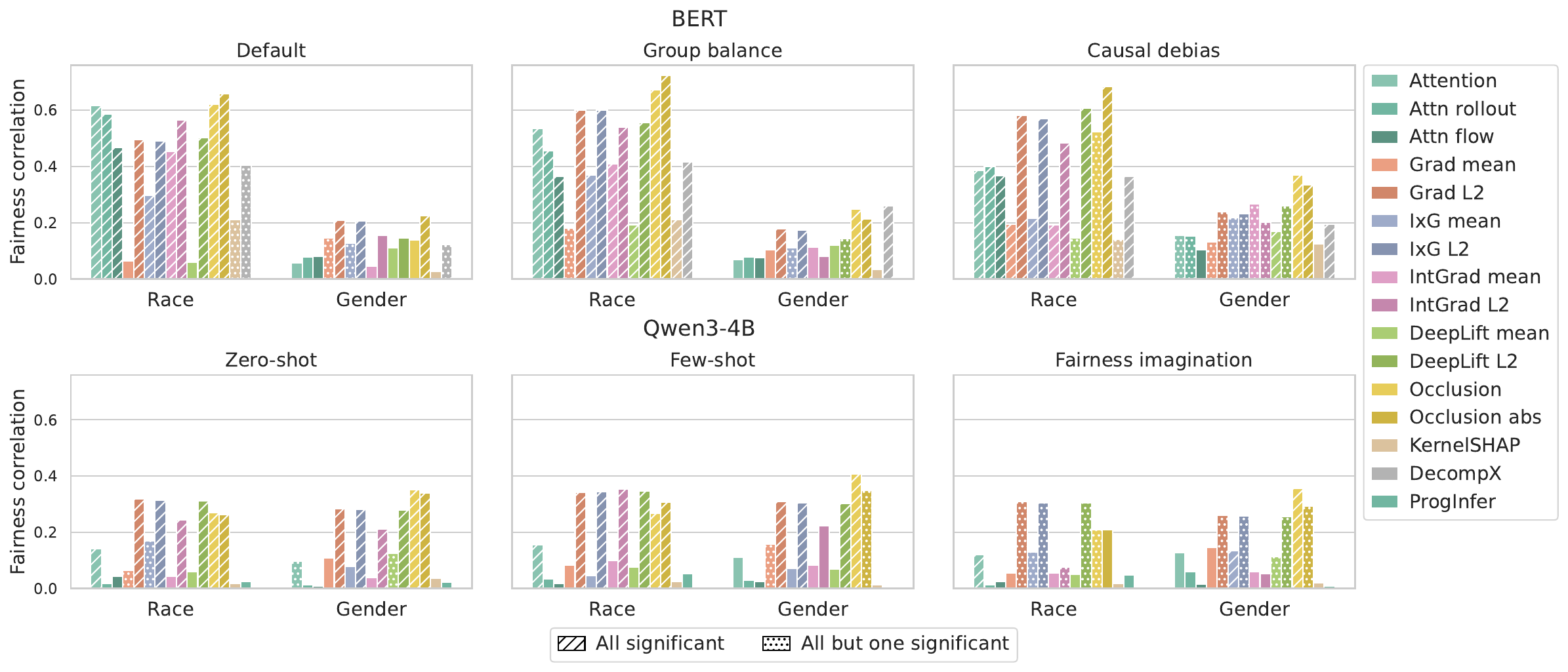}
    \caption{Fairness correlation results for each explanation method. 
    Occlusion- and L2-based explanations are effective for bias detection across different bias types and models.}
    \label{fig:fairness correlation}
    
\end{figure*}

Figure~\ref{fig:fairness correlation} shows that the best-performing explanation methods, such as Grad L2, IxG L2, DeepLift L2, Occlusion, and Occlusion abs, generally achieve high fairness correlations across different models and bias types, indicating a strong ability to detect biased predictions. 
Besides, their fairness correlations are mostly statistically significant ($p < \alpha=0.05)$ in all, or in all but one, class-group categories, which confirms their reliability.
Among these methods, Occlusion and Occlusion abs perform best with BERT models, whereas the L2-based methods Grad L2, IxG L2, and DeepLift L2 are most effective with Qwen3-4B.

When comparing different variants of the same explanation family, mean-based approaches perform considerably worse than their L2-based counterparts, and also underperform compared to undirected attention-based methods.
We attribute this limitation to their dependence on accurately determining the direction of each token’s contribution, a challenge that attention- and L2-based explanations do not face.
Our analysis in Appendix~\ref{appendix:faithfulness} further shows that the effectiveness of explanation-based bias detection is not determined by explanation faithfulness, underscoring the need for careful evaluation when selecting methods for bias identification.
In addition, our human study in Appendix~\ref{appendix:human fairness auditing} shows that explanation-based signals can assist human fairness auditing.


\vspace{-0.5em}
\begin{center}
\colorbox{gray!15}{%
  \parbox{0.98\linewidth}{\textbf{Takeaway}: Input-based explanation methods, particularly Occlusion- and L2-based ones, are effective for identifying biased predictions at inference time.}} 
\end{center}

\subsection{RQ2: Explanations for Model Selection}
\label{sec:rq2}

Given that explanations can detect biased predictions (RQ1), we next investigate whether they can also be used to select fair models among candidates.
Prior work has demonstrated that input-based explanations on validation examples can help humans identify spurious correlations in models~\citep{lertvittayakumjorn-toni-2021-explanation, pezeshkpour-etal-2022-combining}.
Extending this idea, we examine whether explanations can be leveraged for automatic fair model selection, thereby removing the need for human intervention.

Our experiments consist of three steps: (1) for all default and debiased models (seven encoder-only and four decoder-only), we generate predictions on a validation set and compute explanation-based metrics; (2) we compute fairness metrics on the test set for each model; and (3) we evaluate model selection ability using two measures: Spearman’s rank correlation ($\rho$) between validation set explanation-based metrics and test set fairness metrics, which reflects the ability to rank models, and mean reciprocal rank of the fairest model (MRR@1), which reflects the ability to select the fairest model.
Higher rank correlations and MRR@1 indicate that an explanation method is useful for ranking models and selecting the fairest one.
Specifically, we use the average absolute sensitive token reliance on the validation set as the explanation-based metric to rank and select models based on average individual unfairness on the test set.\footnote{We have evaluated other metrics to predict group fairness outcomes. However, neither explanation-based metrics nor validation set fairness achieved rank correlations beyond random chance with the test set results. The full set of evaluated metrics is provided in Appendix~\ref{appendix: rq2}.} 
\paragraph{Results}
\begin{figure*}[h!]  
    \centering
    \includegraphics[width=\linewidth]{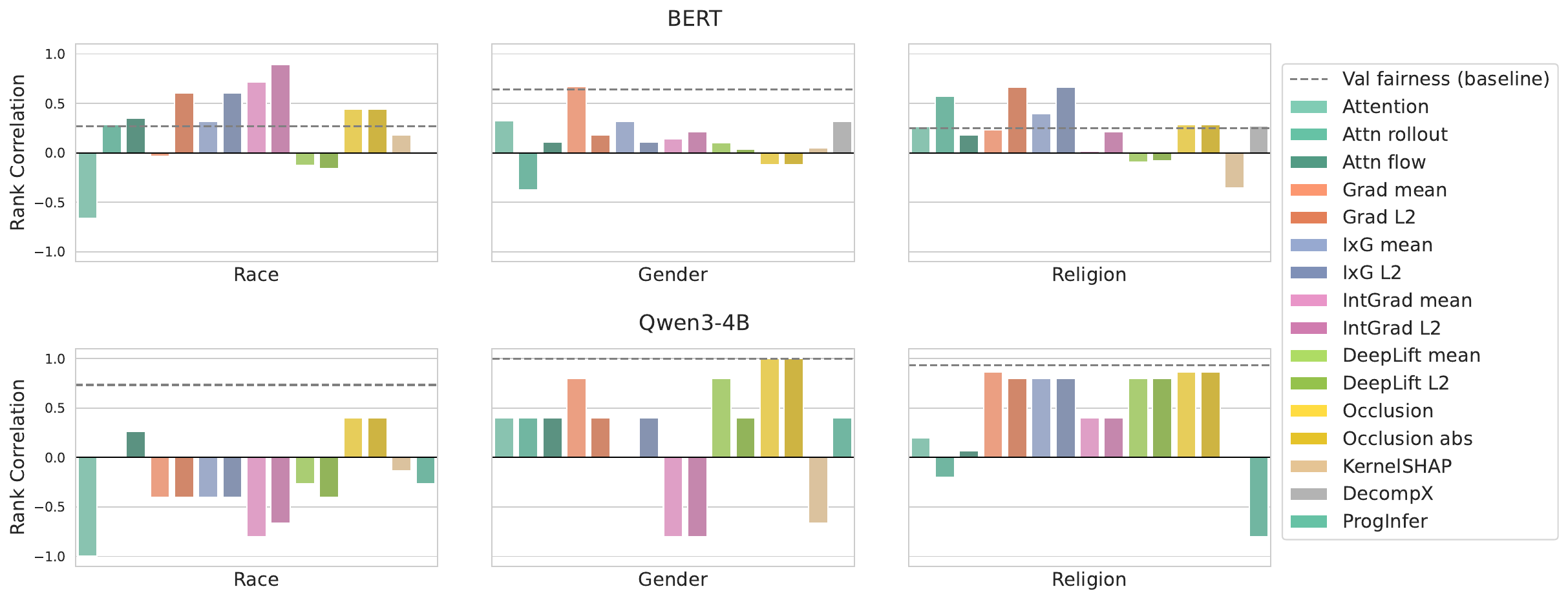}
    \caption{Rank correlations between validation set average absolute sensitive token reliance and test set individual unfairness. 
    The validation set sizes are 500 for race and gender, and 200 for religion. 
    None of the explanation methods consistently achieve performance on par with the baseline.}
    \label{fig:model selection}
    
\end{figure*}

\begin{wrapfigure}{r}{0.48\textwidth} 
    \vspace{-5mm}
    \centering
    \includegraphics[width=\linewidth]{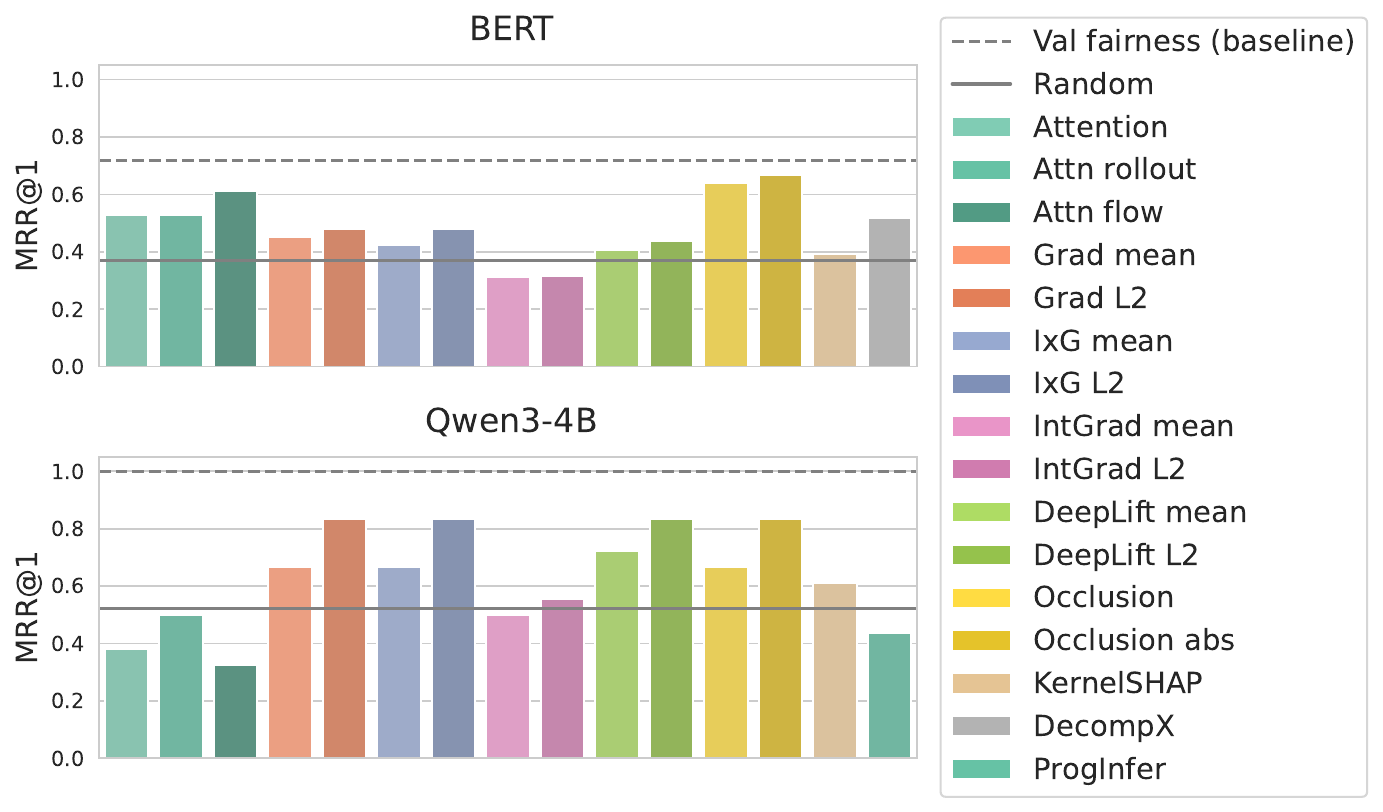}
    \caption{Average MRR@1 across bias types. Explanation methods perform worse than the baseline in identifying the fairest models.}
    \label{fig:mrr}
    \vspace{-5mm}
\end{wrapfigure}
As a baseline, we report results of using the validation set average individual unfairness as the predictor of test set fairness performance.
The results are averaged over six and three random validation set selections for encoder- and decoder-only models, respectively. 
Results for more models and the Jigsaw dataset are presented in Appendix~\ref{appendix: rq2}.
\looseness=-1

The results in Figures~\ref{fig:model selection} and~\ref{fig:mrr} highlight the limitations of using explanations for model selection. 
Although some methods occasionally show high rank correlations (e.g., Grad L2 for race and religion biases in BERT and Occlusion-based methods for gender and religion biases in Qwen3-4B), none of them consistently reach the baseline of using the individual unfairness on the validation set.
This limitation is particularly evident in decoder-only models, where the baseline achieves a perfect rank correlation of 1.
Similarly, the baseline consistently achieves the highest MRR@1 scores, further showing the limited effectiveness of explanation-based methods in selecting the fairest models.
Considering that these explanations are often more computationally expensive to generate than evaluating validation set fairness, they are not practically useful as a model fairness indicator.
Therefore, we do not recommend explanation-based model selection, especially in decoder-only models. 
The difference in findings between RQ1 and RQ2 may stem from the fact that debiasing methods can alter model behaviors and thereby affect explanation attributions.
As a result, comparing explanations across default and debiased models is less reliable, whereas comparing explanations within the same model remains effective for detecting biased predictions.

\vspace{-1.5em}
\begin{center}
\colorbox{gray!15}{
  \parbox{0.98\linewidth}{\textbf{Takeaway}: Input-based explanation methods are not reliable tools for selecting fair models.}}
\end{center}

\subsection{RQ3: Explanations for Bias Mitigation}
\label{sec:rq3}

Having shown that explanations can reliably reveal biased predictions (RQ1), we now investigate whether they can also be leveraged to mitigate model bias.
Building on prior work demonstrating that explanation regularization can reduce spurious correlations while also improving performance and generalization~\citep{kennedy-etal-2020-contextualizing, rao2023studying}, we investigate bias mitigation by minimizing sensitive token reliance during training.
Following~\citet{dimanov2020you}, we define a debiasing regularization term, $\text{L}_{\text{debias}}$, which penalizes the average sensitive token reliance of all such tokens in an input, in addition to the task loss:
\begin{align*}
    \text{L} = \text{L}_\text{task} + \alpha \text{L}_\text{debias}
\end{align*}
Here, $\alpha$ is a hyperparameter that controls the strength of sensitive token reliance reduction. For embedding-level attributions, we apply either an L1 or L2 norm penalty, corresponding to minimizing mean- or L2-based reliance scores, respectively.

While~\citet{dimanov2020you} tune hyperparameters based on task accuracy, we search $\alpha \in \{0.01, 0.1, 1, 10, 100\}$ using a fairness-balanced metric (the harmonic mean of accuracy and 100–unfairness) on the validation set\footnote{As Occlusion is sensitive to the debiasing strength, we use $\alpha \in \{0.002, 0.004, 0.006, 0.008, 0.01\}$.}. 
Models are selected separately for each fairness criterion and results are averaged over three runs.
Due to computational cost, we restrict training to single bias types.
We exclude DeepLift, DecompX, and KernelSHAP, as they are not easily differentiable and thus cannot be incorporated into model training.
Integrated Gradients is substantially more expensive in time and memory for generating explanations and tracking gradients, sp we apply them only to race bias mitigation in BERT and report the results in Table~\ref{tab: intgrad bias mitigation} in Appendix~\ref{appendix: rq3}.
More implementation details are provided in Appendix~\ref{appendix: experimental details}.
\begin{figure*}[!h]
\centering
\includegraphics[width=\linewidth]{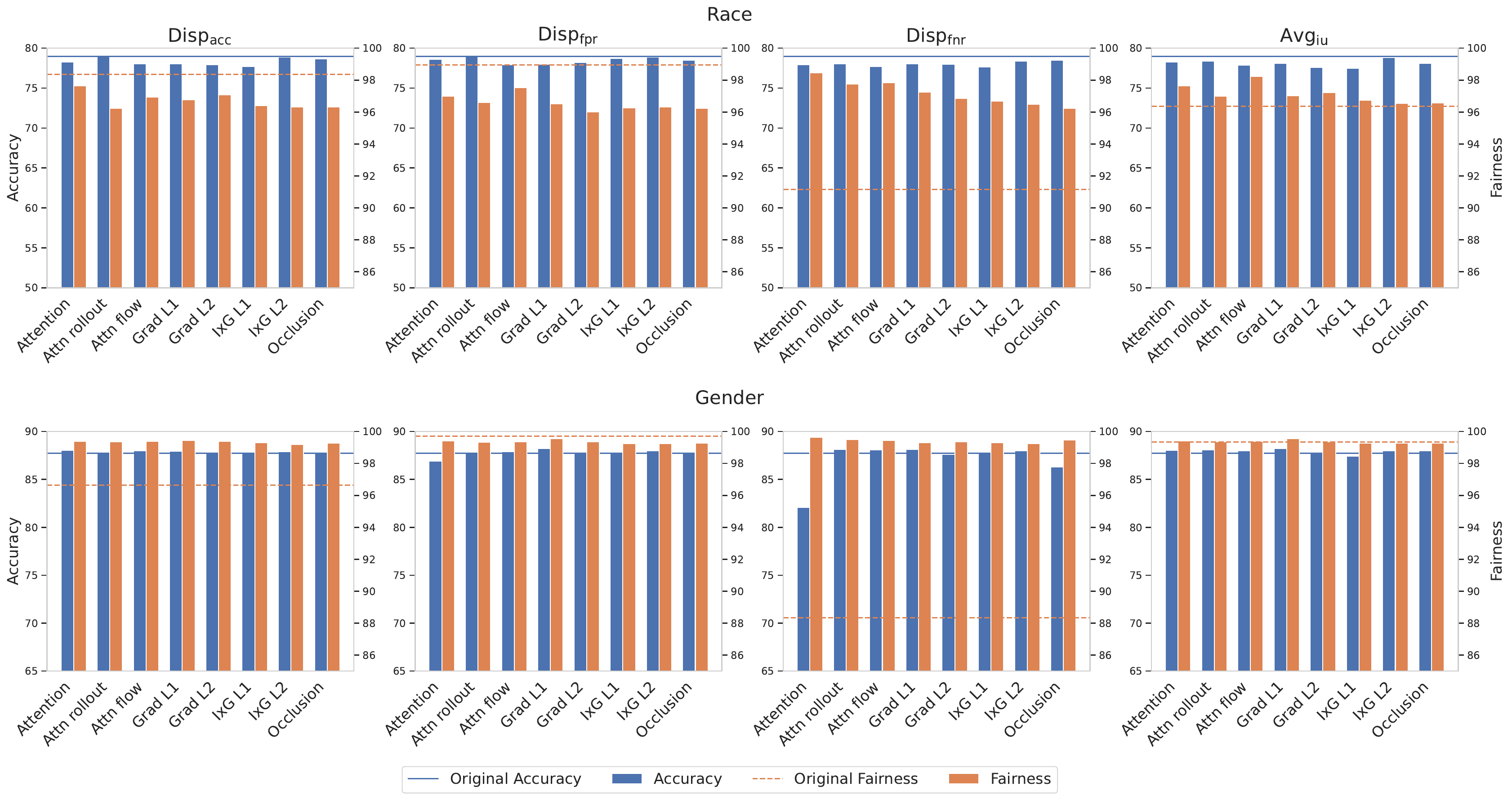}
\caption{Accuracy and fairness results for bias mitigation using different explanation methods. 
Each column corresponds to models selected by maximizing the fairness-balanced metric with respect to the indicated bias metric. We find that explanation methods can improve fairness across many metrics while maintaining reasonable task accuracy.}
\label{fig:bias mitigation}
\end{figure*}
\paragraph{Results}
In Figure~\ref{fig:bias mitigation}, we present race and gender bias mitigation results. For
consistency with accuracy, fairness results are reported as $100-\{\text{Disp}_\text{acc}, \text{Disp}_\text{fpr}, \text{Disp}_\text{fnr}, \text{Avg}_\text{iu}\}$, so that higher
values indicate lower bias.
We find that explanation-based bias mitigation effectively improves fairness across multiple metrics.
Most notably, it consistently and substantially reduces $\text{Disp}_\text{fnr}$ for all bias types.
For gender bias, it also yields considerable reductions in $\text{Disp}_\text{acc}$, and $\text{Avg}_\text{iu}$ is mitigated for race bias.
Moreover, as shown in Figure \ref{fig:appendix bias mitigation civil bert}, all group fairness disparity metrics decrease for religion bias.
The bias mitigation effects are consistent across all models and are also observed on the Jigsaw dataset (see Figures \ref{fig:appendix bias mitigation civil bert}, \ref{fig:appendix bias mitigation civil roberta}, \ref{fig:appendix bias mitigation jigsaw bert}, \ref{fig:appendix bias mitigation jigsaw roberta} in Appendix~\ref{appendix: rq3}).

At the same time, explanation-based debiasing maintains a good balance between fairness and accuracy.
For example, Grad L1 both increases accuracy and reduces $\text{Disp}_\text{acc}$, $\text{Disp}_\text{fnr}$, and $\text{Avg}_\text{iu}$ for gender bias, while most other explanation methods also achieve better $\text{Disp}_\text{acc}$ and $\text{Disp}_\text{fnr}$ with marginal or no accuracy loss.
Our harmonic fairness–accuracy mean results (Figures \ref{fig:appendix harmonic mean bert civil}, \ref{fig:appendix harmonic mean roberta civil}, \ref{fig:appendix harmonic mean bert jigsaw}, \ref{fig:appendix harmonic mean roberta jigsaw}) further confirm this by showing that explanation-based debiasing almost always achieves comparable or higher harmonic means than both default models and traditional debiasing methods.

Among individual explanation methods, attention and attn flow achieve strong debiasing performance on BERT, while IxG L1 and L2 consistently yield a good balance between accuracy and fairness across models.
Overall, IxG L2 and attention-based methods provide robust debiasing while maintaining a favorable fairness–accuracy trade-off across bias types, models, and datasets, as reflected in the harmonic mean results.
Our findings differ from those of~\citet{dimanov2020you}, which we attribute to our fairness-based hyperparameter tuning strategy.
We further show in Appendix~\ref{appendix:hybrid debiasing} that combining traditional and explanation-based debiasing methods outperforms either approach in isolation.
\vspace{-1.5em}
\begin{center}
\colorbox{gray!15}{
  \parbox{0.98\linewidth}{\textbf{Takeaway}: Input-based explanations can provide effective supervision for mitigating model bias during training while maintaining a good fairness–performance trade-off. In particular, IxG L2 and attention-based methods achieve robust debiasing with strong overall balance.}}
\end{center}

\section{Bias Detection in Explanation-Debiased Models}
\label{sec: fairwashing}
\begin{wrapfigure}{r}{0.6\textwidth}
\vspace{-5mm}
    \centering
    \includegraphics[width=\linewidth]{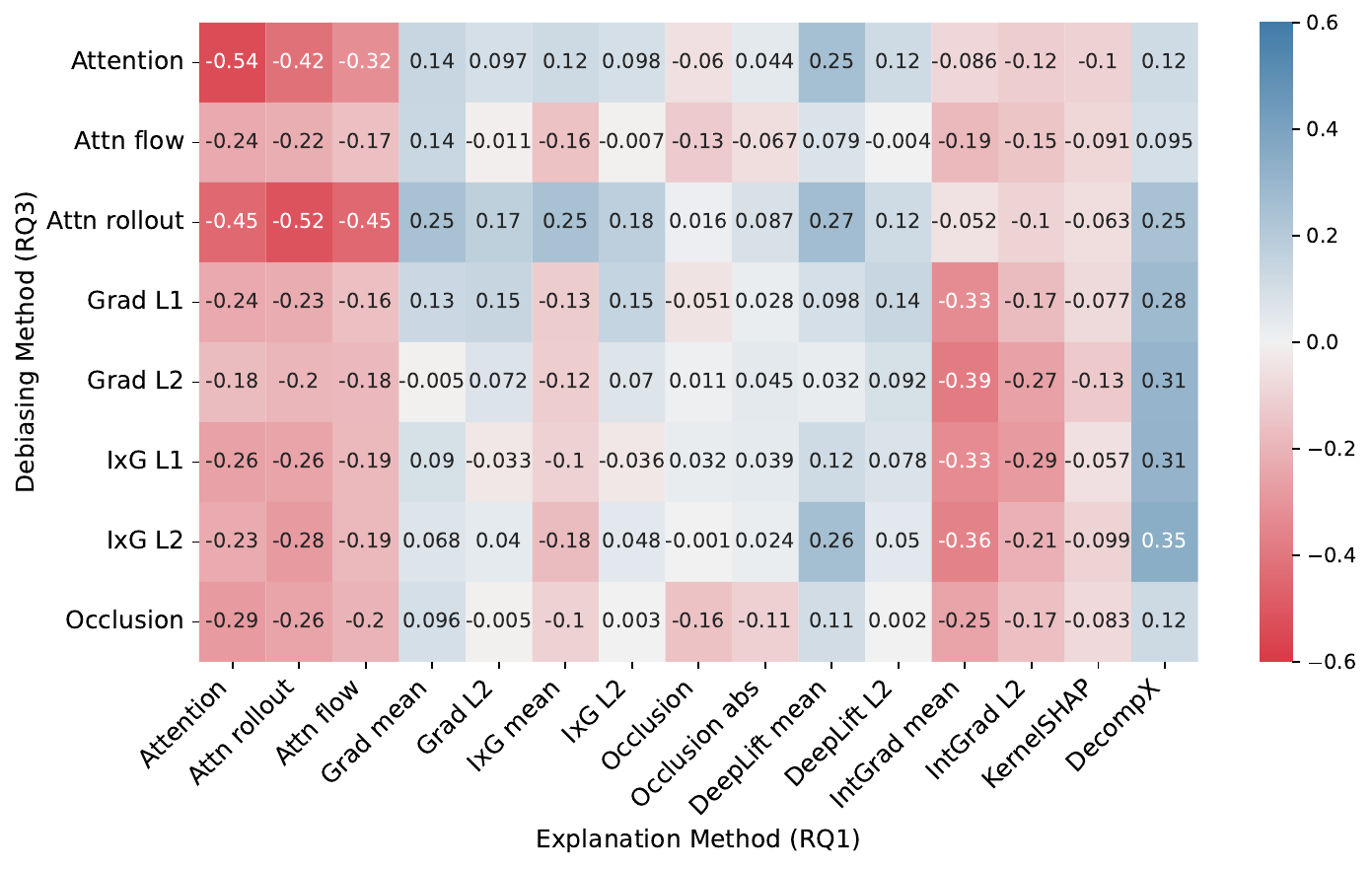}
    \caption{Fairness correlation differences between default and explanation-debiased BERT. Occlusion- and L2-based explanations (except IntGrad L2) are less affected by explanation-based debiasing and remain effective for bias detection.}
\vspace{-5mm}
    \label{fig:fairwashing diff}
\end{wrapfigure}
While explanation-based methods are effective in reducing bias (RQ3), their suppression of attributions on sensitive tokens could potentially mislead users into believing that model predictions are unbiased~\citep{dimanov2020you, fooling-lime, pruthi-etal-2020-learning}.
To investigate this concern, we reapply the bias detection procedure from RQ1 to explanation-debiased models and compare their fairness correlations with those from the corresponding default models. 
For this analysis, we use the models debiased for race bias with respect to individual unfairness, as described in RQ3.

The fairness correlation differences from default models are shown in Figure~\ref{fig:fairwashing diff}. We observe that the impact of explanation-based debiasing on fairness correlations depends on both the explanations used for debiasing and those used for bias detection. 
Some approaches, such as Grad mean/L2, IxG L2, DeepLift mean/L2, Occlusion, and Occlusion abs, are only marginally, or even positively, affected by debiasing.
Their fairness correlation scores (see Figure~\ref{fig:appendix fairwashing race} in Appendix~\ref{appendix: fair-washing}) further indicate that Occlusion-  and L2-based methods (except IntGrad L2) remain reliable for revealing bias in explanation-debiased models.
In contrast, attention-based explanations experience substantial drops, particularly when the models themselves are debiased using attention-based methods.
Similarly, IntGrad-based explanations show a reduced bias detection ability when the debiasing procedure is also gradient-based.
Overall, these findings demonstrate that certain input-based explanations remain effective for detecting biased predictions even in explanation-debiased models.
Our results are different from those of \citet{dimanov2020you}, likely because their analysis focused solely on attribution magnitudes without considering their relationship to fairness metrics.

\section{Explanation-Based Bias Detection vs. LLM-as-a-Judge}
\label{sec:llm}
Existing research suggests that LLMs could identify and correct biased model outputs~\citep{constitutional_ai, furniturewala-etal-2024-thinking}.
In this section, we compare the bias detection ability of input-based explanations against LLMs' judgments under two paradigms: (1) LLM decision, where LLMs are asked to indicate whether a model's prediction rely on bias or stereotypes, and (2) LLM attribution, where LLMs choose a K-word rationale from the input, which we then examine for the presence of sensitive tokens. 
We conduct this analysis using two LLMs, Qwen3-4B and GPT-OSS-120B, on predictions made by Qwen3-4B on the race subset of Civil Comments (see Appendix~\ref{appendix: experimental details} for the prompts used).

Table~\ref{tab:llm explanation} shows the results of LLM-as-a-judge for bias detection. 
Under the LLM decision setup, Qwen3-4B is extremely conservative: it flags only 86 out of 4000 predictions as biased, and all of them correspond to toxic predictions.
Moreover, the predictions labeled as biased by the model exhibit lower average individual unfairness than those labeled as non-biased, indicating poor precision as well.
Under LLM attribution, Qwen3-4B performs slightly better: predictions whose rationales contain sensitive tokens show higher average individual unfairness than those without. However, this still falls short of a simple input-based explanation baseline that flags the top 50\% of predictions ranked by absolute Grad L2 reliance scores (Grad L2 Binary).
The larger GPT-OSS-120B exhibits improved bias detection ability in the LLM decision setting, but its performance under LLM attribution remains comparable to Qwen3-4B and still substantially worse than input-based explanations.
Overall, we conclude that input-based explanations are more reliable than LLM-as-a-judge for bias detection. This finding aligns with the observations of~\citet{rethinking-yang-2025}, who also report that LLM-as-a-judge is unreliable for bias detection.


\begin{table}[h]
    \centering
    \caption{Results of LLM-as-a-judge for bias detection using Qwen3-4B and GPT-OSS-120B. Predictions come from Qwen3-4B on race-related Civil Comments examples. "Biased/Unbiased" denotes whether an example is judged as biased or unbiased by the LLM through LLM decision or LLM attribution. If the judgments are reliable, $\text{Avg}_\text{iu}$ should be higher for biased examples than unbiased ones. For LLM decision with Qwen3-4B, fairness correlation cannot be computed because the model labels no non-toxic predictions as biased. Input-based explanations reveal bias more reliably than LLM-as-a-judge. \\}
    \resizebox{0.8\textwidth}{!}{
    \begin{tabular}{llccc}
    \toprule
     LLM & Method & \# Biased/Unbiased & $\text{Avg}_\text{iu}$ (Biased/Unbiased)  & Fairness Correlation \\
    \midrule
    \multirow{3}{*}{Qwen3-4B} &  LLM decision  & 86/3914 & 0.065/2.59 & - \\
     &  LLM attribution (K=5) & 2063/1904 & 3.55/1.49 & 0.104 \\
     &  LLM attribution (K=10) & 2176/1474 & 2.93/1.56 & 0.070 \\
    \midrule
    \multirow{3}{*}{GPT-OSS-120B} &LLM decision  & 399/3601 & 4.42/2.35 & 0.051 \\
     &  LLM attribution (K=5) & 2153/1843 & 3.33/1.65 & 0.092 \\
     &  LLM attribution (K=10) & 2729/1238 & 2.88/1.74 & 0.063 \\
    \midrule
      \multicolumn{1}{c}{$\text{---}$} & Grad L2 Binary & 2000/2000 & 5.02/0.09 & 0.194 \\
    \bottomrule
    \end{tabular}
    }
    \label{tab:llm explanation}
\end{table}


\section{Conclusion}

In this work, we present the first comprehensive study linking input-based explanations and fairness in hate speech detection.
Our experiments show that (1) input-based explanations can effectively identify biased predictions, (2) they are not reliable for selecting fair models, and (3) they can serve as effective supervision signals during training, mitigating bias while preserving a strong balance between fairness and task performance.
We further provide practical recommendations on which explanation methods are best suited for bias detection and bias mitigation.
Finally, our analyses demonstrate that explanation-based bias detection remains effective in explanation-debiased models, and they outperforms LLM-as-a-judge in identifying biased predictions\footnote{Limitations and future directions are discussed in Appendix~\ref{appendix:limitations}.
}.

\section*{Ethics Statement}

This work investigates explainability and fairness in hate speech detection. 
Despite the diverse experimental setups explored and the additional generalization tests in Appendix~\ref{appendix:generalization}, the findings are still constrained by the specific configurations considered here. As such, the results may not fully generalize across demographic groups, domains, or tasks, and they may remain vulnerable to adversarial manipulation.
We further caution that explanation methods and debiasing techniques cannot fully eliminate residual harms, and that LLM-generated bias judgments are unreliable for bias detection. 
We hope that our study will contribute to the development of NLP systems that are more transparent, reliable, and fair.

\section*{Reproducibility Statement}

We include full implementation details in the main text and appendix, covering data pre-processing details, model architectures, training procedures, and hyperparameters. 
Our code and scripts for data pre-processing and experiments are available at \href{https://github.com/Ewanwong/fairness_x_explainability}{https://github.com/Ewanwong/fairness\_x\_explainability}.

\section*{Acknowledgements}
This work was funded by the DFG project GRK 2853 "Neuroexplicit Models of Language, Vision, and Action" (project number 471607914).

We gratefully acknowledge the computing time made available to them on the high-performance computer at the NHR Center of TU Dresden. This center is jointly supported by the Federal Ministry of Research, Technology and Space of Germany and the state governments participating in the NHR (www.nhr-verein.de/unsere-partner).

Mayank Jobanputra is supported by Deutsche Forschungsgemeinschaft (DFG, German Research Foundation) Project ID 389792660-TRR 248 (Foundations of Perspicuous Software Systems).

We also sincerely thank the reviewers and area chairs for their exceptionally detailed and constructive feedback.

\bibliography{iclr2026_conference}

@inproceedings{balkir-etal-2022-challenges,
    title = "Challenges in Applying Explainability Methods to Improve the Fairness of {NLP} Models",
    author = "Balkir, Esma  and
      Kiritchenko, Svetlana  and
      Nejadgholi, Isar  and
      Fraser, Kathleen",
    editor = "Verma, Apurv  and
      Pruksachatkun, Yada  and
      Chang, Kai-Wei  and
      Galstyan, Aram  and
      Dhamala, Jwala  and
      Cao, Yang Trista",
    booktitle = "Proceedings of the 2nd Workshop on Trustworthy Natural Language Processing (TrustNLP 2022)",
    month = jul,
    year = "2022",
    address = "Seattle, U.S.A.",
    publisher = "Association for Computational Linguistics",
    url = "https://aclanthology.org/2022.trustnlp-1.8/",
    doi = "10.18653/v1/2022.trustnlp-1.8",
    pages = "80--92"
}

@inproceedings{deck-etal-2024-critical-survey,
author = {Deck, Luca and Schoeffer, Jakob and De-Arteaga, Maria and K\"{u}hl, Niklas},
title = {A Critical Survey on Fairness Benefits of Explainable AI},
year = {2024},
isbn = {9798400704505},
publisher = {Association for Computing Machinery},
address = {New York, NY, USA},
url = {https://doi.org/10.1145/3630106.3658990},
doi = {10.1145/3630106.3658990},
booktitle = {Proceedings of the 2024 ACM Conference on Fairness, Accountability, and Transparency},
pages = {1579–1595},
numpages = {17},
location = {Rio de Janeiro, Brazil},
series = {FAccT '24}
}

@inproceedings{garg-etal-2019-counterfactual-fairness,
author = {Garg, Sahaj and Perot, Vincent and Limtiaco, Nicole and Taly, Ankur and Chi, Ed H. and Beutel, Alex},
title = {Counterfactual Fairness in Text Classification through Robustness},
year = {2019},
isbn = {9781450363242},
publisher = {Association for Computing Machinery},
address = {New York, NY, USA},
url = {https://doi.org/10.1145/3306618.3317950},
doi = {10.1145/3306618.3317950},
booktitle = {Proceedings of the 2019 AAAI/ACM Conference on AI, Ethics, and Society},
pages = {219–226},
numpages = {8},
location = {Honolulu, HI, USA},
series = {AIES '19}
}

@inproceedings{jeyaraj-delany-2024-explainable,
    title = "An Explainable Approach to Understanding Gender Stereotype Text",
    author = "Jeyaraj, Manuela  and
      Delany, Sarah",
    editor = "Fale{\'n}ska, Agnieszka  and
      Basta, Christine  and
      Costa-juss{\`a}, Marta  and
      Goldfarb-Tarrant, Seraphina  and
      Nozza, Debora",
    booktitle = "Proceedings of the 5th Workshop on Gender Bias in Natural Language Processing (GeBNLP)",
    month = aug,
    year = "2024",
    address = "Bangkok, Thailand",
    publisher = "Association for Computational Linguistics",
    url = "https://aclanthology.org/2024.gebnlp-1.4/",
    doi = "10.18653/v1/2024.gebnlp-1.4",
    pages = "45--59"
}

@inproceedings{muntasir-etal-2025-explainable-ai,
author = {Muntasir, Fahim and Noor, Jannatun},
title = {Explainable AI Discloses Gender Bias in Sexism Detection Algorithm},
year = {2025},
isbn = {9798400711589},
publisher = {Association for Computing Machinery},
address = {New York, NY, USA},
url = {https://doi.org/10.1145/3704522.3704524},
doi = {10.1145/3704522.3704524},
booktitle = {Proceedings of the 11th International Conference on Networking, Systems, and Security},
pages = {120–127},
numpages = {8},
location = {
},
series = {NSysS '24}
}

@inproceedings{hickey2020fairness,
  title={Fairness by explicability and adversarial SHAP learning},
  author={Hickey, James M and Di Stefano, Pietro G and Vasileiou, Vlasios},
  booktitle={Joint European Conference on Machine Learning and Knowledge Discovery in Databases},
  pages={174--190},
  year={2020},
  organization={Springer}
}

@INPROCEEDINGS{stevens-etal-2020-explainability-and-fairness,
  author={Stevens, Alexander and Deruyck, Peter and Veldhoven, Ziboud Van and Vanthienen, Jan},
  booktitle={2020 IEEE Symposium Series on Computational Intelligence (SSCI)}, 
  title={Explainability and Fairness in Machine Learning: Improve Fair End-to-end Lending for Kiva}, 
  year={2020},
  volume={},
  number={},
  pages={1241-1248},
  doi={10.1109/SSCI47803.2020.9308371}}

@article{meng2022interpretability,
  title={Interpretability and fairness evaluation of deep learning models on MIMIC-IV dataset},
  author={Meng, Chuizheng and Trinh, Loc and Xu, Nan and Enouen, James and Liu, Yan},
  journal={Scientific Reports},
  volume={12},
  number={1},
  pages={7166},
  year={2022},
  publisher={Nature Publishing Group UK London}
}

@inproceedings{prabhakaran-etal-2019-perturbation,
    title = "Perturbation Sensitivity Analysis to Detect Unintended Model Biases",
    author = "Prabhakaran, Vinodkumar  and
      Hutchinson, Ben  and
      Mitchell, Margaret",
    editor = "Inui, Kentaro  and
      Jiang, Jing  and
      Ng, Vincent  and
      Wan, Xiaojun",
    booktitle = "Proceedings of the 2019 Conference on Empirical Methods in Natural Language Processing and the 9th International Joint Conference on Natural Language Processing (EMNLP-IJCNLP)",
    month = nov,
    year = "2019",
    address = "Hong Kong, China",
    publisher = "Association for Computational Linguistics",
    url = "https://aclanthology.org/D19-1578/",
    doi = "10.18653/v1/D19-1578",
    pages = "5740--5745"
}

@inproceedings{kennedy-etal-2020-contextualizing,
    title = "Contextualizing Hate Speech Classifiers with Post-hoc Explanation",
    author = "Kennedy, Brendan  and
      Jin, Xisen  and
      Mostafazadeh Davani, Aida  and
      Dehghani, Morteza  and
      Ren, Xiang",
    editor = "Jurafsky, Dan  and
      Chai, Joyce  and
      Schluter, Natalie  and
      Tetreault, Joel",
    booktitle = "Proceedings of the 58th Annual Meeting of the Association for Computational Linguistics",
    month = jul,
    year = "2020",
    address = "Online",
    publisher = "Association for Computational Linguistics",
    url = "https://aclanthology.org/2020.acl-main.483/",
    doi = "10.18653/v1/2020.acl-main.483",
    pages = "5435--5442"
}

@inproceedings{rao2023studying,
  title={Studying how to efficiently and effectively guide models with explanations},
  author={Rao, Sukrut and B{\"o}hle, Moritz and Parchami-Araghi, Amin and Schiele, Bernt},
  booktitle={Proceedings of the IEEE/CVF International Conference on Computer Vision},
  pages={1922--1933},
  year={2023}
}

@incollection{dimanov2020you,
  title={You shouldn’t trust me: Learning models which conceal unfairness from multiple explanation methods},
  author={Dimanov, Botty and Bhatt, Umang and Jamnik, Mateja and Weller, Adrian},
  booktitle={ECAI 2020},
  pages={2473--2480},
  year={2020},
  publisher={IOS Press}
}

@inproceedings{pruthi-etal-2020-learning,
    title = "Learning to Deceive with Attention-Based Explanations",
    author = "Pruthi, Danish  and
      Gupta, Mansi  and
      Dhingra, Bhuwan  and
      Neubig, Graham  and
      Lipton, Zachary C.",
    editor = "Jurafsky, Dan  and
      Chai, Joyce  and
      Schluter, Natalie  and
      Tetreault, Joel",
    booktitle = "Proceedings of the 58th Annual Meeting of the Association for Computational Linguistics",
    month = jul,
    year = "2020",
    address = "Online",
    publisher = "Association for Computational Linguistics",
    url = "https://aclanthology.org/2020.acl-main.432/",
    doi = "10.18653/v1/2020.acl-main.432",
    pages = "4782--4793"
}

@inproceedings{grabowicz-etal-2022-marrying,
author = {Grabowicz, Przemyslaw A. and Perello, Nicholas and Mishra, Aarshee},
title = {Marrying Fairness and Explainability in Supervised Learning},
year = {2022},
isbn = {9781450393522},
publisher = {Association for Computing Machinery},
address = {New York, NY, USA},
url = {https://doi.org/10.1145/3531146.3533236},
doi = {10.1145/3531146.3533236},
booktitle = {Proceedings of the 2022 ACM Conference on Fairness, Accountability, and Transparency},
pages = {1905–1916},
numpages = {12},
location = {Seoul, Republic of Korea},
series = {FAccT '22}
}

@inproceedings{bhargava2020limeout,
  title={Limeout: An ensemble approach to improve process fairness},
  author={Bhargava, Vaishnavi and Couceiro, Miguel and Napoli, Amedeo},
  booktitle={Joint European conference on machine learning and knowledge discovery in databases},
  pages={475--491},
  year={2020},
  organization={Springer}
}

@inproceedings{sogancioglu-etal-2023-using-explainability,
author = {Sogancioglu, Gizem and Kaya, Heysem and Salah, Albert Ali},
title = {Using Explainability for Bias Mitigation: A Case Study for Fair Recruitment Assessment},
year = {2023},
isbn = {9798400700552},
publisher = {Association for Computing Machinery},
address = {New York, NY, USA},
url = {https://doi.org/10.1145/3577190.3614170},
doi = {10.1145/3577190.3614170},
booktitle = {Proceedings of the 25th International Conference on Multimodal Interaction},
pages = {631–639},
numpages = {9},
location = {Paris, France},
series = {ICMI '23}
}

@inproceedings{ye-etal-2025-input,
    title = "Can Input Attributions Explain Inductive Reasoning in In-Context Learning?",
    author = "Ye, Mengyu  and
      Kuribayashi, Tatsuki  and
      Kobayashi, Goro  and
      Suzuki, Jun",
    editor = "Che, Wanxiang  and
      Nabende, Joyce  and
      Shutova, Ekaterina  and
      Pilehvar, Mohammad Taher",
    booktitle = "Findings of the Association for Computational Linguistics: ACL 2025",
    month = jul,
    year = "2025",
    address = "Vienna, Austria",
    publisher = "Association for Computational Linguistics",
    url = "https://aclanthology.org/2025.findings-acl.1092/",
    doi = "10.18653/v1/2025.findings-acl.1092",
    pages = "21199--21225",
    ISBN = "979-8-89176-256-5"
}

@inproceedings{chen-etal-2025-causally,
    title = "Causally Testing Gender Bias in {LLM}s: A Case Study on Occupational Bias",
    author = "Chen, Yuen  and
      Raghuram, Vethavikashini Chithrra  and
      Mattern, Justus  and
      Mihalcea, Rada  and
      Jin, Zhijing",
    editor = "Chiruzzo, Luis  and
      Ritter, Alan  and
      Wang, Lu",
    booktitle = "Findings of the Association for Computational Linguistics: NAACL 2025",
    month = apr,
    year = "2025",
    address = "Albuquerque, New Mexico",
    publisher = "Association for Computational Linguistics",
    url = "https://aclanthology.org/2025.findings-naacl.281/",
    doi = "10.18653/v1/2025.findings-naacl.281",
    pages = "4984--5004",
    ISBN = "979-8-89176-195-7"
}

@inproceedings{furniturewala-etal-2024-thinking,
    title = "``Thinking'' Fair and Slow: On the Efficacy of Structured Prompts for Debiasing Language Models",
    author = "Furniturewala, Shaz  and
      Jandial, Surgan  and
      Java, Abhinav  and
      Banerjee, Pragyan  and
      Shahid, Simra  and
      Bhatia, Sumit  and
      Jaidka, Kokil",
    editor = "Al-Onaizan, Yaser  and
      Bansal, Mohit  and
      Chen, Yun-Nung",
    booktitle = "Proceedings of the 2024 Conference on Empirical Methods in Natural Language Processing",
    month = nov,
    year = "2024",
    address = "Miami, Florida, USA",
    publisher = "Association for Computational Linguistics",
    url = "https://aclanthology.org/2024.emnlp-main.13/",
    doi = "10.18653/v1/2024.emnlp-main.13",
    pages = "213--227"
}

@inproceedings{wang-etal-2018-glue,
    title = "{GLUE}: A Multi-Task Benchmark and Analysis Platform for Natural Language Understanding",
    author = "Wang, Alex  and
      Singh, Amanpreet  and
      Michael, Julian  and
      Hill, Felix  and
      Levy, Omer  and
      Bowman, Samuel",
    editor = "Linzen, Tal  and
      Chrupa\l a, Grzegorz  and
      Alishahi, Afra",
    booktitle = "Proceedings of the 2018 {EMNLP} Workshop {B}lackbox{NLP}: Analyzing and Interpreting Neural Networks for {NLP}",
    month = nov,
    year = "2018",
    address = "Brussels, Belgium",
    publisher = "Association for Computational Linguistics",
    url = "https://aclanthology.org/W18-5446/",
    doi = "10.18653/v1/W18-5446",
    pages = "353--355",
}

@article{Bommasani-2021-FoundationModels,
  author       = {Rishi Bommasani and
                  Drew A. Hudson and
                  Ehsan Adeli and
                  Russ B. Altman and
                  Simran Arora and
                  Sydney von Arx and
                  Michael S. Bernstein and
                  Jeannette Bohg and
                  Antoine Bosselut and
                  Emma Brunskill and
                  Erik Brynjolfsson and
                  Shyamal Buch and
                  Dallas Card and
                  Rodrigo Castellon and
                  Niladri S. Chatterji and
                  Annie S. Chen and
                  Kathleen Creel and
                  Jared Quincy Davis and
                  Dorottya Demszky and
                  Chris Donahue and
                  Moussa Doumbouya and
                  Esin Durmus and
                  Stefano Ermon and
                  John Etchemendy and
                  Kawin Ethayarajh and
                  Li Fei{-}Fei and
                  Chelsea Finn and
                  Trevor Gale and
                  Lauren E. Gillespie and
                  Karan Goel and
                  Noah D. Goodman and
                  Shelby Grossman and
                  Neel Guha and
                  Tatsunori Hashimoto and
                  Peter Henderson and
                  John Hewitt and
                  Daniel E. Ho and
                  Jenny Hong and
                  Kyle Hsu and
                  Jing Huang and
                  Thomas Icard and
                  Saahil Jain and
                  Dan Jurafsky and
                  Pratyusha Kalluri and
                  Siddharth Karamcheti and
                  Geoff Keeling and
                  Fereshte Khani and
                  Omar Khattab and
                  Pang Wei Koh and
                  Mark S. Krass and
                  Ranjay Krishna and
                  Rohith Kuditipudi and
                  et al.},
  title        = {On the Opportunities and Risks of Foundation Models},
  journal      = {CoRR},
  volume       = {abs/2108.07258},
  year         = {2021},
  url          = {https://arxiv.org/abs/2108.07258},
  eprinttype    = {arXiv},
  eprint       = {2108.07258},
  bibsource    = {dblp computer science bibliography, https://dblp.org}
}

@misc{Eval-2023-Harness,
	author       = {Gao, Leo and Tow, Jonathan and Abbasi, Baber and Biderman, Stella and Black, Sid and DiPofi, Anthony and Foster, Charles and Golding, Laurence and Hsu, Jeffrey and Le Noac'h, Alain and Li, Haonan and McDonell, Kyle and Muennighoff, Niklas and Ociepa, Chris and Phang, Jason and Reynolds, Laria and Schoelkopf, Hailey and Skowron, Aviya and Sutawika, Lintang and Tang, Eric and Thite, Anish and Wang, Ben and Wang, Kevin and Zou, Andy},
	title        = {A framework for few-shot language model evaluation},
	month        = 12,
	year         = 2023,
	publisher    = {Zenodo},
	version      = {v0.4.0},
	doi          = {10.5281/zenodo.10256836},
	url          = {https://zenodo.org/records/10256836}
}

@article{gallegos-etal-2024-bias,
    title = "Bias and Fairness in Large Language Models: A Survey",
    author = "Gallegos, Isabel O.  and
      Rossi, Ryan A.  and
      Barrow, Joe  and
      Tanjim, Md Mehrab  and
      Kim, Sungchul  and
      Dernoncourt, Franck  and
      Yu, Tong  and
      Zhang, Ruiyi  and
      Ahmed, Nesreen K.",
    journal = "Computational Linguistics",
    volume = "50",
    number = "3",
    month = sep,
    year = "2024",
    address = "Cambridge, MA",
    publisher = "MIT Press",
    url = "https://aclanthology.org/2024.cl-3.8/",
    doi = "10.1162/coli_a_00524",
    pages = "1097--1179"
}

@inproceedings{
gupta2024calm,
title={{CALM} : A Multi-task Benchmark for Comprehensive Assessment of Language Model Bias},
author={Vipul Gupta and Pranav Narayanan Venkit and Hugo Lauren{\c{c}}on and Shomir Wilson and Rebecca J. Passonneau},
booktitle={First Conference on Language Modeling},
year={2024},
url={https://openreview.net/forum?id=RLFca3arx7}
}

@inproceedings{sheng-etal-2021-societal,
    title = "Societal Biases in Language Generation: Progress and Challenges",
    author = "Sheng, Emily  and
      Chang, Kai-Wei  and
      Natarajan, Prem  and
      Peng, Nanyun",
    editor = "Zong, Chengqing  and
      Xia, Fei  and
      Li, Wenjie  and
      Navigli, Roberto",
    booktitle = "Proceedings of the 59th Annual Meeting of the Association for Computational Linguistics and the 11th International Joint Conference on Natural Language Processing (Volume 1: Long Papers)",
    month = aug,
    year = "2021",
    address = "Online",
    publisher = "Association for Computational Linguistics",
    url = "https://aclanthology.org/2021.acl-long.330/",
    doi = "10.18653/v1/2021.acl-long.330",
    pages = "4275--4293"
}

@article{bolukbasi2016man,
  title={Man is to computer programmer as woman is to homemaker? debiasing word embeddings},
  author={Bolukbasi, Tolga and Chang, Kai-Wei and Zou, James Y and Saligrama, Venkatesh and Kalai, Adam T},
  journal={Advances in neural information processing systems},
  volume={29},
  year={2016}
}

@inproceedings{may2019measuring,
  title={On Measuring Social Biases in Sentence Encoders},
  author={May, Chandler and Wang, Alex and Bordia, Shikha and Bowman, Samuel and Rudinger, Rachel},
  booktitle={Proceedings of the 2019 Conference of the North American Chapter of the Association for Computational Linguistics: Human Language Technologies, Volume 1 (Long and Short Papers)},
  pages={622--628},
  year={2019}
}

@inproceedings{zhao-etal-2018-gender,
    title = "Gender Bias in Coreference Resolution: Evaluation and Debiasing Methods",
    author = "Zhao, Jieyu  and
      Wang, Tianlu  and
      Yatskar, Mark  and
      Ordonez, Vicente  and
      Chang, Kai-Wei",
    booktitle = "Proceedings of the 2018 Conference of the North {A}merican Chapter of the Association for Computational Linguistics: Human Language Technologies, Volume 2 (Short Papers)",
    month = jun,
    year = "2018",
    address = "New Orleans, Louisiana",
    publisher = "Association for Computational Linguistics",
    url = "https://aclanthology.org/N18-2003",
    doi = "10.18653/v1/N18-2003",
    pages = "15--20",
}

@article{fang2024bias,
  title={Bias of AI-generated content: an examination of news produced by large language models},
  author={Fang, Xiao and Che, Shangkun and Mao, Minjia and Zhang, Hongzhe and Zhao, Ming and Zhao, Xiaohang},
  journal={Scientific Reports},
  volume={14},
  number={1},
  pages={5224},
  year={2024},
  publisher={Nature Publishing Group UK London}
}

@inproceedings{wan-chang-2025-white,
    title = "White Men Lead, Black Women Help? Benchmarking and Mitigating Language Agency Social Biases in {LLM}s",
    author = "Wan, Yixin  and
      Chang, Kai-Wei",
    editor = "Che, Wanxiang  and
      Nabende, Joyce  and
      Shutova, Ekaterina  and
      Pilehvar, Mohammad Taher",
    booktitle = "Proceedings of the 63rd Annual Meeting of the Association for Computational Linguistics (Volume 1: Long Papers)",
    month = jul,
    year = "2025",
    address = "Vienna, Austria",
    publisher = "Association for Computational Linguistics",
    url = "https://aclanthology.org/2025.acl-long.445/",
    doi = "10.18653/v1/2025.acl-long.445",
    pages = "9082--9108",
    ISBN = "979-8-89176-251-0"
}

@inproceedings{sheng-etal-2019-woman,
    title = "The Woman Worked as a Babysitter: On Biases in Language Generation",
    author = "Sheng, Emily  and
      Chang, Kai-Wei  and
      Natarajan, Premkumar  and
      Peng, Nanyun",
    editor = "Inui, Kentaro  and
      Jiang, Jing  and
      Ng, Vincent  and
      Wan, Xiaojun",
    booktitle = "Proceedings of the 2019 Conference on Empirical Methods in Natural Language Processing and the 9th International Joint Conference on Natural Language Processing (EMNLP-IJCNLP)",
    month = nov,
    year = "2019",
    address = "Hong Kong, China",
    publisher = "Association for Computational Linguistics",
    url = "https://aclanthology.org/D19-1339/",
    doi = "10.18653/v1/D19-1339",
    pages = "3407--3412"
}

@inproceedings{arras-etal-2019-evaluating,
    title = "Evaluating Recurrent Neural Network Explanations",
    author = {Arras, Leila  and
      Osman, Ahmed  and
      M\"uller, Klaus-Robert  and
      Samek, Wojciech},
    editor = "Linzen, Tal  and
      Chrupa\l a, Grzegorz  and
      Belinkov, Yonatan  and
      Hupkes, Dieuwke",
    booktitle = "Proceedings of the 2019 ACL Workshop BlackboxNLP: Analyzing and Interpreting Neural Networks for NLP",
    month = aug,
    year = "2019",
    address = "Florence, Italy",
    publisher = "Association for Computational Linguistics",
    url = "https://aclanthology.org/W19-4813/",
    doi = "10.18653/v1/W19-4813",
    pages = "113--126"
}

@inproceedings{atanasova2022diagnostics,
  title={Diagnostics-guided explanation generation},
  author={Atanasova, Pepa and Simonsen, Jakob Grue and Lioma, Christina and Augenstein, Isabelle},
  booktitle={Proceedings of the AAAI Conference on Artificial Intelligence},
  volume={36},
  pages={10445--10453},
  year={2022}
}

@article{lyu-etal-2024-towards,
    title = "Towards Faithful Model Explanation in {NLP}: A Survey",
    author = "Lyu, Qing  and
      Apidianaki, Marianna  and
      Callison-Burch, Chris",
    journal = "Computational Linguistics",
    volume = "50",
    number = "2",
    month = jun,
    year = "2024",
    address = "Cambridge, MA",
    publisher = "MIT Press",
    url = "https://aclanthology.org/2024.cl-2.6/",
    doi = "10.1162/coli_a_00511",
    pages = "657--723"
}

@inproceedings{fooling-lime,
author = {Slack, Dylan and Hilgard, Sophie and Jia, Emily and Singh, Sameer and Lakkaraju, Himabindu},
title = {Fooling LIME and SHAP: Adversarial Attacks on Post hoc Explanation Methods},
year = {2020},
isbn = {9781450371100},
publisher = {Association for Computing Machinery},
address = {New York, NY, USA},
url = {https://doi.org/10.1145/3375627.3375830},
doi = {10.1145/3375627.3375830},
booktitle = {Proceedings of the AAAI/ACM Conference on AI, Ethics, and Society},
pages = {180–186},
numpages = {7},
location = {New York, NY, USA},
series = {AIES '20}
}

@article{hatexplain, title={HateXplain: A Benchmark Dataset for Explainable Hate Speech Detection}, volume={35}, url={https://ojs.aaai.org/index.php/AAAI/article/view/17745}, DOI={10.1609/aaai.v35i17.17745}, abstractNote={Hate speech is a challenging issue plaguing the online social media. While better models for hate speech detection are continuously being developed, there is little research on the bias and interpretability aspects of hate speech. In this paper, we introduce HateXplain, the first benchmark hate speech dataset covering multiple aspects of the issue. Each post in our dataset is annotated from three different perspectives: the basic, commonly used 3-class classification (i.e., hate, offensive or normal), the target community (i.e., the community that has been the victim of hate speech/offensive speech in the post), and the rationales, i.e., the portions of the post on which their labelling decision (as hate, offensive or normal) is based. We utilize existing state-of-the-art models and observe that even models that perform very well in classification do not score high on explainability metrics like model plausibility and faithfulness. We also observe that models, which utilize the human rationales for training, perform better in reducing unintended bias towards target communities. We have made our code and dataset public for other researchers.}, number={17}, journal={Proceedings of the AAAI Conference on Artificial Intelligence}, author={Mathew, Binny and Saha, Punyajoy and Yimam, Seid Muhie and Biemann, Chris and Goyal, Pawan and Mukherjee, Animesh}, year={2021}, month={May}, pages={14867-14875} }

@inproceedings{sap-etal-2019-risk,
    title = "The Risk of Racial Bias in Hate Speech Detection",
    author = "Sap, Maarten  and
      Card, Dallas  and
      Gabriel, Saadia  and
      Choi, Yejin  and
      Smith, Noah A.",
    editor = "Korhonen, Anna  and
      Traum, David  and
      M{\`a}rquez, Llu{\'i}s",
    booktitle = "Proceedings of the 57th Annual Meeting of the Association for Computational Linguistics",
    month = jul,
    year = "2019",
    address = "Florence, Italy",
    publisher = "Association for Computational Linguistics",
    url = "https://aclanthology.org/P19-1163/",
    doi = "10.18653/v1/P19-1163",
    pages = "1668--1678"
}

@inproceedings{yin-neubig-2022-interpreting,
    title = "Interpreting Language Models with Contrastive Explanations",
    author = "Yin, Kayo  and
      Neubig, Graham",
    editor = "Goldberg, Yoav  and
      Kozareva, Zornitsa  and
      Zhang, Yue",
    booktitle = "Proceedings of the 2022 Conference on Empirical Methods in Natural Language Processing",
    month = dec,
    year = "2022",
    address = "Abu Dhabi, United Arab Emirates",
    publisher = "Association for Computational Linguistics",
    url = "https://aclanthology.org/2022.emnlp-main.14/",
    doi = "10.18653/v1/2022.emnlp-main.14",
    pages = "184--198"
}

@inproceedings{madsen-etal-2024-self,
    title = "Are self-explanations from Large Language Models faithful?",
    author = "Madsen, Andreas  and
      Chandar, Sarath  and
      Reddy, Siva",
    editor = "Ku, Lun-Wei  and
      Martins, Andre  and
      Srikumar, Vivek",
    booktitle = "Findings of the Association for Computational Linguistics: ACL 2024",
    month = aug,
    year = "2024",
    address = "Bangkok, Thailand",
    publisher = "Association for Computational Linguistics",
    url = "https://aclanthology.org/2024.findings-acl.19/",
    doi = "10.18653/v1/2024.findings-acl.19",
    pages = "295--337"
}

@inproceedings{
ramnath2024tailoring,
title={Tailoring Self-Rationalizers with Multi-Reward Distillation},
author={Sahana Ramnath and Brihi Joshi and Skyler Hallinan and Ximing Lu and Liunian Harold Li and Aaron Chan and Jack Hessel and Yejin Choi and Xiang Ren},
booktitle={The Twelfth International Conference on Learning Representations},
year={2024},
url={https://openreview.net/forum?id=t8eO0CiZJV}
}

@inproceedings{wang-etal-2025-cross,
    title = "Cross-Refine: Improving Natural Language Explanation Generation by Learning in Tandem",
    author = {Wang, Qianli  and
      Anikina, Tatiana  and
      Feldhus, Nils  and
      Ostermann, Simon  and
      M{\"o}ller, Sebastian  and
      Schmitt, Vera},
    editor = "Rambow, Owen  and
      Wanner, Leo  and
      Apidianaki, Marianna  and
      Al-Khalifa, Hend  and
      Eugenio, Barbara Di  and
      Schockaert, Steven",
    booktitle = "Proceedings of the 31st International Conference on Computational Linguistics",
    month = jan,
    year = "2025",
    address = "Abu Dhabi, UAE",
    publisher = "Association for Computational Linguistics",
    url = "https://aclanthology.org/2025.coling-main.77/",
    pages = "1150--1167"
}

@inproceedings{yu-etal-2024-latent,
    title = "Latent Concept-based Explanation of {NLP} Models",
    author = "Yu, Xuemin  and
      Dalvi, Fahim  and
      Durrani, Nadir  and
      Nouri, Marzia  and
      Sajjad, Hassan",
    editor = "Al-Onaizan, Yaser  and
      Bansal, Mohit  and
      Chen, Yun-Nung",
    booktitle = "Proceedings of the 2024 Conference on Empirical Methods in Natural Language Processing",
    month = nov,
    year = "2024",
    address = "Miami, Florida, USA",
    publisher = "Association for Computational Linguistics",
    url = "https://aclanthology.org/2024.emnlp-main.692/",
    doi = "10.18653/v1/2024.emnlp-main.692",
    pages = "12435--12459"
}

@InProceedings{raman-etal-2024-understanding,
  title = 	 {Understanding Inter-Concept Relationships in Concept-Based Models},
  author =       {Raman, Naveen Janaki and Espinosa Zarlenga, Mateo and Jamnik, Mateja},
  booktitle = 	 {Proceedings of the 41st International Conference on Machine Learning},
  pages = 	 {42009--42025},
  year = 	 {2024},
  editor = 	 {Salakhutdinov, Ruslan and Kolter, Zico and Heller, Katherine and Weller, Adrian and Oliver, Nuria and Scarlett, Jonathan and Berkenkamp, Felix},
  volume = 	 {235},
  series = 	 {Proceedings of Machine Learning Research},
  month = 	 {21--27 Jul},
  publisher =    {PMLR},
  url = 	 {https://proceedings.mlr.press/v235/raman24a.html}
}

@inproceedings{simonyan-etal-2014,
  author       = {Karen Simonyan and
                  Andrea Vedaldi and
                  Andrew Zisserman},
  editor       = {Yoshua Bengio and
                  Yann LeCun},
  title        = {Deep Inside Convolutional Networks: Visualising Image Classification
                  Models and Saliency Maps},
  booktitle    = {2nd International Conference on Learning Representations, {ICLR} 2014,
                  Banff, AB, Canada, April 14-16, 2014, Workshop Track Proceedings},
  year         = {2014},
  url          = {http://arxiv.org/abs/1312.6034},
  bibsource    = {dblp computer science bibliography, https://dblp.org}
}

@article{kindermans-etal-2016,
  author       = {Pieter{-}Jan Kindermans and
                  Kristof Sch{\"{u}}tt and
                  Klaus{-}Robert M{\"{u}}ller and
                  Sven D{\"{a}}hne},
  title        = {Investigating the influence of noise and distractors on the interpretation
                  of neural networks},
  journal      = {CoRR},
  volume       = {abs/1611.07270},
  year         = {2016},
  url          = {http://arxiv.org/abs/1611.07270},
  eprinttype    = {arXiv},
  eprint       = {1611.07270},
  bibsource    = {dblp computer science bibliography, https://dblp.org}
}

@InProceedings{Sundararajan-2017-IntegratedGrad,
  title = 	 {Axiomatic Attribution for Deep Networks},
  author =       {Mukund Sundararajan and Ankur Taly and Qiqi Yan},
  booktitle = 	 {Proceedings of the 34th International Conference on Machine Learning},
  pages = 	 {3319--3328},
  year = 	 {2017},
  editor = 	 {Precup, Doina and Teh, Yee Whye},
  volume = 	 {70},
  series = 	 {Proceedings of Machine Learning Research},
  month = 	 {06--11 Aug},
  publisher =    {PMLR},
  url = 	 {https://proceedings.mlr.press/v70/sundararajan17a.html},
}

@inproceedings{enguehard-2023-sequential,
    title = "Sequential Integrated Gradients: a simple but effective method for explaining language models",
    author = "Enguehard, Joseph",
    editor = "Rogers, Anna  and
      Boyd-Graber, Jordan  and
      Okazaki, Naoaki",
    booktitle = "Findings of the Association for Computational Linguistics: ACL 2023",
    month = jul,
    year = "2023",
    address = "Toronto, Canada",
    publisher = "Association for Computational Linguistics",
    url = "https://aclanthology.org/2023.findings-acl.477/",
    doi = "10.18653/v1/2023.findings-acl.477",
    pages = "7555--7565"
}

@article{bach2015pixel,
  title={On pixel-wise explanations for non-linear classifier decisions by layer-wise relevance propagation},
  author={Bach, Sebastian and Binder, Alexander and Montavon, Gr{\'e}goire and Klauschen, Frederick and M{\"u}ller, Klaus-Robert and Samek, Wojciech},
  journal={PloS one},
  volume={10},
  number={7},
  pages={e0130140},
  year={2015},
  publisher={Public Library of Science San Francisco, CA USA}
}

@inproceedings{shrikumar2017learning,
  title={Learning important features through propagating activation differences},
  author={Shrikumar, Avanti and Greenside, Peyton and Kundaje, Anshul},
  booktitle={International conference on machine learning},
  pages={3145--3153},
  year={2017},
  organization={PMlR}
}

@inproceedings{ferrando-etal-2022-measuring,
    title = "Measuring the Mixing of Contextual Information in the Transformer",
    author = "Ferrando, Javier  and
      G\'allego, Gerard I.  and
      Costa-juss\`a, Marta R.",
    editor = "Goldberg, Yoav  and
      Kozareva, Zornitsa  and
      Zhang, Yue",
    booktitle = "Proceedings of the 2022 Conference on Empirical Methods in Natural Language Processing",
    month = dec,
    year = "2022",
    address = "Abu Dhabi, United Arab Emirates",
    publisher = "Association for Computational Linguistics",
    url = "https://aclanthology.org/2022.emnlp-main.595/",
    doi = "10.18653/v1/2022.emnlp-main.595",
    pages = "8698--8714"
}

@inproceedings{modarressi-etal-2022-globenc,
    title = "{G}lob{E}nc: Quantifying Global Token Attribution by Incorporating the Whole Encoder Layer in Transformers",
    author = "Modarressi, Ali  and
      Fayyaz, Mohsen  and
      Yaghoobzadeh, Yadollah  and
      Pilehvar, Mohammad Taher",
    editor = "Carpuat, Marine  and
      de Marneffe, Marie-Catherine  and
      Meza Ruiz, Ivan Vladimir",
    booktitle = "Proceedings of the 2022 Conference of the North American Chapter of the Association for Computational Linguistics: Human Language Technologies",
    month = jul,
    year = "2022",
    address = "Seattle, United States",
    publisher = "Association for Computational Linguistics",
    url = "https://aclanthology.org/2022.naacl-main.19/",
    doi = "10.18653/v1/2022.naacl-main.19",
    pages = "258--271"
}

@inproceedings{Ribeiro-2016-LIME,
  title={" Why should i trust you?" Explaining the predictions of any classifier},
  author={Ribeiro, Marco Tulio and Singh, Sameer and Guestrin, Carlos},
  booktitle={Proceedings of the 22nd ACM SIGKDD international conference on knowledge discovery and data mining},
  pages={1135--1144},
  year={2016}
}

@inproceedings{Lundberg-2017-SHAP,
 author = {Lundberg, Scott M and Lee, Su-In},
 booktitle = {Advances in Neural Information Processing Systems},
 editor = {I. Guyon and U. Von Luxburg and S. Bengio and H. Wallach and R. Fergus and S. Vishwanathan and R. Garnett},
 pages = {1--10},
 publisher = {Curran Associates, Inc.},
 title = {A Unified Approach to Interpreting Model Predictions},
 url = {https://proceedings.neurips.cc/paper_files/paper/2017/file/8a20a8621978632d76c43dfd28b67767-Paper.pdf},
 volume = {30},
 year = {2017}
}

@inproceedings{deiseroth-etal-2023-atman,
 author = {Deiseroth, Bj\"{o}rn and Deb, Mayukh and Weinbach, Samuel and Brack, Manuel and Schramowski, Patrick and Kersting, Kristian},
 booktitle = {Advances in Neural Information Processing Systems},
 editor = {A. Oh and T. Naumann and A. Globerson and K. Saenko and M. Hardt and S. Levine},
 pages = {63437--63460},
 publisher = {Curran Associates, Inc.},
 title = {ATMAN: Understanding Transformer Predictions Through Memory Efficient Attention Manipulation},
 url = {https://proceedings.neurips.cc/paper_files/paper/2023/file/c83bc020a020cdeb966ed10804619664-Paper-Conference.pdf},
 volume = {36},
 year = {2023}
}

@inproceedings{attention,
  author       = {Dzmitry Bahdanau and
                  Kyunghyun Cho and
                  Yoshua Bengio},
  editor       = {Yoshua Bengio and
                  Yann LeCun},
  title        = {Neural Machine Translation by Jointly Learning to Align and Translate},
  booktitle    = {3rd International Conference on Learning Representations, {ICLR} 2015,
                  San Diego, CA, USA, May 7-9, 2015, Conference Track Proceedings},
  year         = {2015},
  url          = {http://arxiv.org/abs/1409.0473},
  bibsource    = {dblp computer science bibliography, https://dblp.org}
}

@inproceedings{modarressi-etal-2023-decompx,
    title = "{D}ecomp{X}: Explaining Transformers Decisions by Propagating Token Decomposition",
    author = "Modarressi, Ali  and
      Fayyaz, Mohsen  and
      Aghazadeh, Ehsan  and
      Yaghoobzadeh, Yadollah  and
      Pilehvar, Mohammad Taher",
    editor = "Rogers, Anna  and
      Boyd-Graber, Jordan  and
      Okazaki, Naoaki",
    booktitle = "Proceedings of the 61st Annual Meeting of the Association for Computational Linguistics (Volume 1: Long Papers)",
    month = jul,
    year = "2023",
    address = "Toronto, Canada",
    publisher = "Association for Computational Linguistics",
    url = "https://aclanthology.org/2023.acl-long.149/",
    doi = "10.18653/v1/2023.acl-long.149",
    pages = "2649--2664"
}

@inproceedings{abnar-zuidema-2020-quantifying,
    title = "Quantifying Attention Flow in Transformers",
    author = "Abnar, Samira  and
      Zuidema, Willem",
    editor = "Jurafsky, Dan  and
      Chai, Joyce  and
      Schluter, Natalie  and
      Tetreault, Joel",
    booktitle = "Proceedings of the 58th Annual Meeting of the Association for Computational Linguistics",
    month = jul,
    year = "2020",
    address = "Online",
    publisher = "Association for Computational Linguistics",
    url = "https://aclanthology.org/2020.acl-main.385/",
    doi = "10.18653/v1/2020.acl-main.385",
    pages = "4190--4197"
}

@article{occlusion,
  author       = {Jiwei Li and
                  Will Monroe and
                  Dan Jurafsky},
  title        = {Understanding Neural Networks through Representation Erasure},
  journal      = {CoRR},
  volume       = {abs/1612.08220},
  year         = {2016},
  url          = {http://arxiv.org/abs/1612.08220},
  eprinttype    = {arXiv},
  eprint       = {1612.08220},
  bibsource    = {dblp computer science bibliography, https://dblp.org}
}

@inproceedings{enouen-etal-2024-textgenshap,
    title = "{T}ext{G}en{SHAP}: Scalable Post-Hoc Explanations in Text Generation with Long Documents",
    author = "Enouen, James  and
      Nakhost, Hootan  and
      Ebrahimi, Sayna  and
      Arik, Sercan  and
      Liu, Yan  and
      Pfister, Tomas",
    editor = "Ku, Lun-Wei  and
      Martins, Andre  and
      Srikumar, Vivek",
    booktitle = "Findings of the Association for Computational Linguistics: ACL 2024",
    month = aug,
    year = "2024",
    address = "Bangkok, Thailand",
    publisher = "Association for Computational Linguistics",
    url = "https://aclanthology.org/2024.findings-acl.832/",
    doi = "10.18653/v1/2024.findings-acl.832",
    pages = "13984--14011"
}

@inproceedings{cohen-etal-2024-contextcite,
 author = {Cohen-Wang, Benjamin and Shah, Harshay and Georgiev, Kristian and Madry, Aleksander}, 
 booktitle = {Advances in Neural Information Processing Systems},
 editor = {A. Globerson and L. Mackey and D. Belgrave and A. Fan and U. Paquet and J. Tomczak and C. Zhang},
 pages = {95764--95807},
 publisher = {Curran Associates, Inc.},
 title = {ContextCite: Attributing Model Generation to Context},
 url = {https://proceedings.neurips.cc/paper_files/paper/2024/file/adbea136219b64db96a9941e4249a857-Paper-Conference.pdf},
 volume = {37},
 year = {2024}
}

@inproceedings{jain-wallace-2019-attention,
    title = "{A}ttention is not {E}xplanation",
    author = "Jain, Sarthak  and
      Wallace, Byron C.",
    editor = "Burstein, Jill  and
      Doran, Christy  and
      Solorio, Thamar",
    booktitle = "Proceedings of the 2019 Conference of the North {A}merican Chapter of the Association for Computational Linguistics: Human Language Technologies, Volume 1 (Long and Short Papers)",
    month = jun,
    year = "2019",
    address = "Minneapolis, Minnesota",
    publisher = "Association for Computational Linguistics",
    url = "https://aclanthology.org/N19-1357/",
    doi = "10.18653/v1/N19-1357",
    pages = "3543--3556"
}

@article{enhancing-santiago-2025,
  author       = {Santiago Gonz{\'{a}}lez{-}Silot and
                  Andr{\'{e}}s Montoro{-}Montarroso and
                  Eugenio Mart{\'{\i}}nez C{\'{a}}mara and
                  Juan G{\'{o}}mez{-}Romero},
  title        = {Enhancing Disinformation Detection with Explainable {AI} and Named
                  Entity Replacement},
  journal      = {CoRR},
  volume       = {abs/2502.04863},
  year         = {2025},
  url          = {https://doi.org/10.48550/arXiv.2502.04863},
  doi          = {10.48550/ARXIV.2502.04863},
  eprinttype    = {arXiv},
  eprint       = {2502.04863},
  bibsource    = {dblp computer science bibliography, https://dblp.org}
}

@inproceedings{borkan2019nuanced,
  title={Nuanced metrics for measuring unintended bias with real data for text classification},
  author={Borkan, Daniel and Dixon, Lucas and Sorensen, Jeffrey and Thain, Nithum and Vasserman, Lucy},
  booktitle={Companion proceedings of the 2019 world wide web conference},
  pages={491--500},
  year={2019}
}

@misc{jigsaw-unintended-bias-in-toxicity-classification,
    author = {cjadams and Daniel Borkan and inversion and Jeffrey Sorensen and Lucas Dixon and Lucy Vasserman and nithum},
    title = {Jigsaw Unintended Bias in Toxicity Classification},
    year = {2019},
    howpublished = {\url{https://kaggle.com/competitions/jigsaw-unintended-bias-in-toxicity-classification}},
    note = {Kaggle}
}

@inproceedings{wang-demberg-2024-parameter,
    title = "A Parameter-Efficient Multi-Objective Approach to Mitigate Stereotypical Bias in Language Models",
    author = "Wang, Yifan  and
      Demberg, Vera",
    editor = "Fale{\'n}ska, Agnieszka  and
      Basta, Christine  and
      Costa-juss{\`a}, Marta  and
      Goldfarb-Tarrant, Seraphina  and
      Nozza, Debora",
    booktitle = "Proceedings of the 5th Workshop on Gender Bias in Natural Language Processing (GeBNLP)",
    month = aug,
    year = "2024",
    address = "Bangkok, Thailand",
    publisher = "Association for Computational Linguistics",
    url = "https://aclanthology.org/2024.gebnlp-1.1/",
    doi = "10.18653/v1/2024.gebnlp-1.1",
    pages = "1--19",
}

@article{caliskan2017semantics,
  title={Semantics derived automatically from language corpora contain human-like biases},
  author={Caliskan, Aylin and Bryson, Joanna J and Narayanan, Arvind},
  journal={Science},
  volume={356},
  number={6334},
  pages={183--186},
  year={2017},
  publisher={American Association for the Advancement of Science}
}

@article{roberta,
  author       = {Yinhan Liu and
                  Myle Ott and
                  Naman Goyal and
                  Jingfei Du and
                  Mandar Joshi and
                  Danqi Chen and
                  Omer Levy and
                  Mike Lewis and
                  Luke Zettlemoyer and
                  Veselin Stoyanov},
  title        = {RoBERTa: {A} Robustly Optimized {BERT} Pretraining Approach},
  journal      = {CoRR},
  volume       = {abs/1907.11692},
  year         = {2019},
  url          = {http://arxiv.org/abs/1907.11692},
  eprinttype    = {arXiv},
  eprint       = {1907.11692},
  bibsource    = {dblp computer science bibliography, https://dblp.org}
}

@inproceedings{devlin-etal-2019-bert,
    title = "{BERT}: Pre-training of Deep Bidirectional Transformers for Language Understanding",
    author = "Devlin, Jacob  and
      Chang, Ming-Wei  and
      Lee, Kenton  and
      Toutanova, Kristina",
    editor = "Burstein, Jill  and
      Doran, Christy  and
      Solorio, Thamar",
    booktitle = "Proceedings of the 2019 Conference of the North {A}merican Chapter of the Association for Computational Linguistics: Human Language Technologies, Volume 1 (Long and Short Papers)",
    month = jun,
    year = "2019",
    address = "Minneapolis, Minnesota",
    publisher = "Association for Computational Linguistics",
    url = "https://aclanthology.org/N19-1423/",
    doi = "10.18653/v1/N19-1423",
    pages = "4171--4186"
}

@inproceedings{zmigrod-etal-2019-counterfactual,
    title = "Counterfactual Data Augmentation for Mitigating Gender Stereotypes in Languages with Rich Morphology",
    author = "Zmigrod, Ran  and
      Mielke, Sabrina J.  and
      Wallach, Hanna  and
      Cotterell, Ryan",
    editor = "Korhonen, Anna  and
      Traum, David  and
      M{\`a}rquez, Llu{\'i}s",
    booktitle = "Proceedings of the 57th Annual Meeting of the Association for Computational Linguistics",
    month = jul,
    year = "2019",
    address = "Florence, Italy",
    publisher = "Association for Computational Linguistics",
    url = "https://aclanthology.org/P19-1161/",
    doi = "10.18653/v1/P19-1161",
    pages = "1651--1661"
}

@article{dropout-debiasing,
  author       = {Kellie Webster and
                  Xuezhi Wang and
                  Ian Tenney and
                  Alex Beutel and
                  Emily Pitler and
                  Ellie Pavlick and
                  Jilin Chen and
                  Slav Petrov},
  title        = {Measuring and Reducing Gendered Correlations in Pre-trained Models},
  journal      = {CoRR},
  volume       = {abs/2010.06032},
  year         = {2020},
  url          = {https://arxiv.org/abs/2010.06032},
  eprinttype    = {arXiv},
  eprint       = {2010.06032},
  bibsource    = {dblp computer science bibliography, https://dblp.org}
}

@inproceedings{zhou-etal-2023-causal,
    title = "Causal-Debias: Unifying Debiasing in Pretrained Language Models and Fine-tuning via Causal Invariant Learning",
    author = "Zhou, Fan  and
      Mao, Yuzhou  and
      Yu, Liu  and
      Yang, Yi  and
      Zhong, Ting",
    editor = "Rogers, Anna  and
      Boyd-Graber, Jordan  and
      Okazaki, Naoaki",
    booktitle = "Proceedings of the 61st Annual Meeting of the Association for Computational Linguistics (Volume 1: Long Papers)",
    month = jul,
    year = "2023",
    address = "Toronto, Canada",
    publisher = "Association for Computational Linguistics",
    url = "https://aclanthology.org/2023.acl-long.232/",
    doi = "10.18653/v1/2023.acl-long.232",
    pages = "4227--4241"
}

@inproceedings{attanasio-etal-2022-entropy,
    title = "Entropy-based Attention Regularization Frees Unintended Bias Mitigation from Lists",
    author = "Attanasio, Giuseppe  and
      Nozza, Debora  and
      Hovy, Dirk  and
      Baralis, Elena",
    editor = "Muresan, Smaranda  and
      Nakov, Preslav  and
      Villavicencio, Aline",
    booktitle = "Findings of the Association for Computational Linguistics: ACL 2022",
    month = may,
    year = "2022",
    address = "Dublin, Ireland",
    publisher = "Association for Computational Linguistics",
    url = "https://aclanthology.org/2022.findings-acl.88/",
    doi = "10.18653/v1/2022.findings-acl.88",
    pages = "1105--1119"
}

@article{kamiran2012data,
  title={Data preprocessing techniques for classification without discrimination},
  author={Kamiran, Faisal and Calders, Toon},
  journal={Knowledge and information systems},
  volume={33},
  number={1},
  pages={1--33},
  year={2012},
  publisher={Springer}
}

@inproceedings{park-etal-2018-reducing,
    title = "Reducing Gender Bias in Abusive Language Detection",
    author = "Park, Ji Ho  and
      Shin, Jamin  and
      Fung, Pascale",
    editor = "Riloff, Ellen  and
      Chiang, David  and
      Hockenmaier, Julia  and
      Tsujii, Jun{'}ichi",
    booktitle = "Proceedings of the 2018 Conference on Empirical Methods in Natural Language Processing",
    month = oct # "-" # nov,
    year = "2018",
    address = "Brussels, Belgium",
    publisher = "Association for Computational Linguistics",
    url = "https://aclanthology.org/D18-1302/",
    doi = "10.18653/v1/D18-1302",
    pages = "2799--2804"
}

@inproceedings{dixon-measuring-and-mitigating-2018,
author = {Dixon, Lucas and Li, John and Sorensen, Jeffrey and Thain, Nithum and Vasserman, Lucy},
title = {Measuring and Mitigating Unintended Bias in Text Classification},
year = {2018},
isbn = {9781450360128},
publisher = {Association for Computing Machinery},
address = {New York, NY, USA},
url = {https://doi.org/10.1145/3278721.3278729},
doi = {10.1145/3278721.3278729},
booktitle = {Proceedings of the 2018 AAAI/ACM Conference on AI, Ethics, and Society},
pages = {67–73},
numpages = {7},
location = {New Orleans, LA, USA},
series = {AIES '18}
}

@inproceedings{pezeshkpour-etal-2022-combining,
    title = "Combining Feature and Instance Attribution to Detect Artifacts",
    author = "Pezeshkpour, Pouya  and
      Jain, Sarthak  and
      Singh, Sameer  and
      Wallace, Byron",
    editor = "Muresan, Smaranda  and
      Nakov, Preslav  and
      Villavicencio, Aline",
    booktitle = "Findings of the Association for Computational Linguistics: ACL 2022",
    month = may,
    year = "2022",
    address = "Dublin, Ireland",
    publisher = "Association for Computational Linguistics",
    url = "https://aclanthology.org/2022.findings-acl.153/",
    doi = "10.18653/v1/2022.findings-acl.153",
    pages = "1934--1946"
}

@article{lertvittayakumjorn-toni-2021-explanation,
    title = "Explanation-Based Human Debugging of {NLP} Models: A Survey",
    author = "Lertvittayakumjorn, Piyawat  and
      Toni, Francesca",
    editor = "Roark, Brian  and
      Nenkova, Ani",
    journal = "Transactions of the Association for Computational Linguistics",
    volume = "9",
    year = "2021",
    address = "Cambridge, MA",
    publisher = "MIT Press",
    url = "https://aclanthology.org/2021.tacl-1.90/",
    doi = "10.1162/tacl_a_00440",
    pages = "1508--1528"
}

@inproceedings{deyoung-etal-2020-eraser,
    title = "{ERASER}: {A} Benchmark to Evaluate Rationalized {NLP} Models",
    author = "DeYoung, Jay  and
      Jain, Sarthak  and
      Rajani, Nazneen Fatema  and
      Lehman, Eric  and
      Xiong, Caiming  and
      Socher, Richard  and
      Wallace, Byron C.",
    editor = "Jurafsky, Dan  and
      Chai, Joyce  and
      Schluter, Natalie  and
      Tetreault, Joel",
    booktitle = "Proceedings of the 58th Annual Meeting of the Association for Computational Linguistics",
    month = jul,
    year = "2020",
    address = "Online",
    publisher = "Association for Computational Linguistics",
    url = "https://aclanthology.org/2020.acl-main.408/",
    doi = "10.18653/v1/2020.acl-main.408",
    pages = "4443--4458"
}

@article{qwen3,
  author       = {An Yang and
                  Anfeng Li and
                  Baosong Yang and
                  Beichen Zhang and
                  Binyuan Hui and
                  Bo Zheng and
                  Bowen Yu and
                  Chang Gao and
                  Chengen Huang and
                  Chenxu Lv and
                  Chujie Zheng and
                  Dayiheng Liu and
                  Fan Zhou and
                  Fei Huang and
                  Feng Hu and
                  Hao Ge and
                  Haoran Wei and
                  Huan Lin and
                  Jialong Tang and
                  Jian Yang and
                  Jianhong Tu and
                  Jianwei Zhang and
                  Jian Yang and
                  Jiaxi Yang and
                  Jingren Zhou and
                  Junyang Lin and
                  Kai Dang and
                  Keqin Bao and
                  Kexin Yang and
                  Le Yu and
                  Lianghao Deng and
                  Mei Li and
                  Mingfeng Xue and
                  Mingze Li and
                  Pei Zhang and
                  Peng Wang and
                  Qin Zhu and
                  Rui Men and
                  Ruize Gao and
                  Shixuan Liu and
                  Shuang Luo and
                  Tianhao Li and
                  Tianyi Tang and
                  Wenbiao Yin and
                  Xingzhang Ren and
                  Xinyu Wang and
                  Xinyu Zhang and
                  Xuancheng Ren and
                  Yang Fan and
                  Yang Su and
                  Yichang Zhang and
                  Yinger Zhang and
                  Yu Wan and
                  Yuqiong Liu and
                  Zekun Wang and
                  Zeyu Cui and
                  Zhenru Zhang and
                  Zhipeng Zhou and
                  Zihan Qiu},
  title        = {Qwen3 Technical Report},
  journal      = {CoRR},
  volume       = {abs/2505.09388},
  year         = {2025},
  url          = {https://doi.org/10.48550/arXiv.2505.09388},
  doi          = {10.48550/ARXIV.2505.09388},
  eprinttype    = {arXiv},
  eprint       = {2505.09388},
  bibsource    = {dblp computer science bibliography, https://dblp.org}
}

@article{llama3,
  author       = {Abhimanyu Dubey and
                  Abhinav Jauhri and
                  Abhinav Pandey and
                  Abhishek Kadian and
                  Ahmad Al{-}Dahle and
                  Aiesha Letman and
                  Akhil Mathur and
                  Alan Schelten and
                  Amy Yang and
                  Angela Fan and
                  Anirudh Goyal and
                  Anthony Hartshorn and
                  Aobo Yang and
                  Archi Mitra and
                  Archie Sravankumar and
                  Artem Korenev and
                  Arthur Hinsvark and
                  Arun Rao and
                  Aston Zhang and
                  Aur{\'{e}}lien Rodriguez and
                  Austen Gregerson and
                  Ava Spataru and
                  Baptiste Rozi{\`{e}}re and
                  Bethany Biron and
                  Binh Tang and
                  Bobbie Chern and
                  Charlotte Caucheteux and
                  Chaya Nayak and
                  Chloe Bi and
                  Chris Marra and
                  Chris McConnell and
                  Christian Keller and
                  Christophe Touret and
                  Chunyang Wu and
                  Corinne Wong and
                  Cristian Canton Ferrer and
                  Cyrus Nikolaidis and
                  Damien Allonsius and
                  Daniel Song and
                  Danielle Pintz and
                  Danny Livshits and
                  David Esiobu and
                  Dhruv Choudhary and
                  Dhruv Mahajan and
                  Diego Garcia{-}Olano and
                  Diego Perino and
                  Dieuwke Hupkes and
                  Egor Lakomkin and
                  Ehab AlBadawy and
                  Elina Lobanova and
                  Emily Dinan and
                  Eric Michael Smith and
                  Filip Radenovic and
                  Frank Zhang and
                  Gabriel Synnaeve and
                  Gabrielle Lee and
                  Georgia Lewis Anderson and
                  Graeme Nail and
                  Gr{\'{e}}goire Mialon and
                  Guan Pang and
                  Guillem Cucurell and
                  Hailey Nguyen and
                  Hannah Korevaar and
                  Hu Xu and
                  Hugo Touvron and
                  Iliyan Zarov and
                  Imanol Arrieta Ibarra and
                  Isabel M. Kloumann and
                  Ishan Misra and
                  Ivan Evtimov and
                  Jade Copet and
                  Jaewon Lee and
                  Jan Geffert and
                  Jana Vranes and
                  Jason Park and
                  Jay Mahadeokar and
                  Jeet Shah and
                  Jelmer van der Linde and
                  Jennifer Billock and
                  Jenny Hong and
                  Jenya Lee and
                  Jeremy Fu and
                  Jianfeng Chi and
                  Jianyu Huang and
                  Jiawen Liu and
                  Jie Wang and
                  Jiecao Yu and
                  Joanna Bitton and
                  Joe Spisak and
                  Jongsoo Park and
                  Joseph Rocca and
                  Joshua Johnstun and
                  Joshua Saxe and
                  Junteng Jia and
                  Kalyan Vasuden Alwala and
                  Kartikeya Upasani and
                  Kate Plawiak and
                  Ke Li and
                  Kenneth Heafield and
                  Kevin Stone and
                  et al.},
  title        = {The Llama 3 Herd of Models},
  journal      = {CoRR},
  volume       = {abs/2407.21783},
  year         = {2024},
  url          = {https://doi.org/10.48550/arXiv.2407.21783},
  doi          = {10.48550/ARXIV.2407.21783},
  eprinttype    = {arXiv},
  eprint       = {2407.21783},
  bibsource    = {dblp computer science bibliography, https://dblp.org}
}

@article{constitutional_ai,
  author       = {Yuntao Bai and
                  Saurav Kadavath and
                  Sandipan Kundu and
                  Amanda Askell and
                  Jackson Kernion and
                  Andy Jones and
                  Anna Chen and
                  Anna Goldie and
                  Azalia Mirhoseini and
                  Cameron McKinnon and
                  Carol Chen and
                  Catherine Olsson and
                  Christopher Olah and
                  Danny Hernandez and
                  Dawn Drain and
                  Deep Ganguli and
                  Dustin Li and
                  Eli Tran{-}Johnson and
                  Ethan Perez and
                  Jamie Kerr and
                  Jared Mueller and
                  Jeffrey Ladish and
                  Joshua Landau and
                  Kamal Ndousse and
                  Kamile Lukosiute and
                  Liane Lovitt and
                  Michael Sellitto and
                  Nelson Elhage and
                  Nicholas Schiefer and
                  Noem{\'{\i}} Mercado and
                  Nova DasSarma and
                  Robert Lasenby and
                  Robin Larson and
                  Sam Ringer and
                  Scott Johnston and
                  Shauna Kravec and
                  Sheer El Showk and
                  Stanislav Fort and
                  Tamera Lanham and
                  Timothy Telleen{-}Lawton and
                  Tom Conerly and
                  Tom Henighan and
                  Tristan Hume and
                  Samuel R. Bowman and
                  Zac Hatfield{-}Dodds and
                  Ben Mann and
                  Dario Amodei and
                  Nicholas Joseph and
                  Sam McCandlish and
                  Tom Brown and
                  Jared Kaplan},
  title        = {Constitutional {AI:} Harmlessness from {AI} Feedback},
  journal      = {CoRR},
  volume       = {abs/2212.08073},
  year         = {2022},
  url          = {https://doi.org/10.48550/arXiv.2212.08073},
  doi          = {10.48550/ARXIV.2212.08073},
  eprinttype    = {arXiv},
  eprint       = {2212.08073},
  bibsource    = {dblp computer science bibliography, https://dblp.org}
}

@InProceedings{pmlr-v235-kariyappa24a,
  title = 	 {Progressive Inference: Explaining Decoder-Only Sequence Classification Models Using Intermediate Predictions},
  author =       {Kariyappa, Sanjay and Lecue, Freddy and Mishra, Saumitra and Pond, Christopher and Magazzeni, Daniele and Veloso, Manuela},
  booktitle = 	 {Proceedings of the 41st International Conference on Machine Learning},
  pages = 	 {23238--23255},
  year = 	 {2024},
  editor = 	 {Salakhutdinov, Ruslan and Kolter, Zico and Heller, Katherine and Weller, Adrian and Oliver, Nuria and Scarlett, Jonathan and Berkenkamp, Felix},
  volume = 	 {235},
  series = 	 {Proceedings of Machine Learning Research},
  month = 	 {21--27 Jul},
  publisher =    {PMLR},
  url = 	 {https://proceedings.mlr.press/v235/kariyappa24a.html}
}

@article{rethinking-yang-2025,
  author       = {Xinyi Yang and
                  Runzhe Zhan and
                  Derek F. Wong and
                  Shu Yang and
                  Junchao Wu and
                  Lidia S. Chao},
  title        = {Rethinking Prompt-based Debiasing in Large Language Models},
  journal      = {CoRR},
  volume       = {abs/2503.09219},
  year         = {2025},
  url          = {https://doi.org/10.48550/arXiv.2503.09219},
  doi          = {10.48550/ARXIV.2503.09219},
  eprinttype    = {arXiv},
  eprint       = {2503.09219},
  bibsource    = {dblp computer science bibliography, https://dblp.org}
}

@article{devil-liu-2024,
  author       = {Yan Liu and
                  Yu Liu and
                  Xiaokang Chen and
                  Pin{-}Yu Chen and
                  Daoguang Zan and
                  Min{-}Yen Kan and
                  Tsung{-}Yi Ho},
  title        = {The Devil is in the Neurons: Interpreting and Mitigating Social Biases
                  in Pre-trained Language Models},
  journal      = {CoRR},
  volume       = {abs/2406.10130},
  year         = {2024},
  url          = {https://doi.org/10.48550/arXiv.2406.10130},
  doi          = {10.48550/ARXIV.2406.10130},
  eprinttype    = {arXiv},
  eprint       = {2406.10130},
  bibsource    = {dblp computer science bibliography, https://dblp.org}
}

@article{survey_on_automatic_detection,
author = {Fortuna, Paula and Nunes, S\'{e}rgio},
title = {A Survey on Automatic Detection of Hate Speech in Text},
year = {2018},
issue_date = {July 2019},
publisher = {Association for Computing Machinery},
address = {New York, NY, USA},
volume = {51},
number = {4},
issn = {0360-0300},
url = {https://doi.org/10.1145/3232676},
doi = {10.1145/3232676},
journal = {ACM Comput. Surv.},
month = jul,
articleno = {85},
numpages = {30}
}

@article{Davidson_Warmsley_Macy_Weber_2017, title={Automated Hate Speech Detection and the Problem of Offensive Language}, volume={11}, url={https://ojs.aaai.org/index.php/ICWSM/article/view/14955}, DOI={10.1609/icwsm.v11i1.14955}, abstractNote={ &lt;p&gt; A key challenge for automatic hate-speech detection on social media is the separation of hate speech from other instances of offensive language. Lexical detection methods tend to have low precision because they classify all messages containing particular terms as hate speech and previous work using supervised learning has failed to distinguish between the two categories. We used a crowd-sourced hate speech lexicon to collect tweets containing hate speech keywords. We use crowd-sourcing to label a sample of these tweets into three categories: those containing hate speech, only offensive language, and those with neither. We train a multi-class classifier to distinguish between these different categories. Close analysis of the predictions and the errors shows when we can reliably separate hate speech from other offensive language and when this differentiation is more difficult. We find that racist and homophobic tweets are more likely to be classified as hate speech but that sexist tweets are generally classified as offensive. Tweets without explicit hate keywords are also more difficult to classify. &lt;/p&gt; }, number={1}, journal={Proceedings of the International AAAI Conference on Web and Social Media}, author={Davidson, Thomas and Warmsley, Dana and Macy, Michael and Weber, Ingmar}, year={2017}, month={May}, pages={512-515} }

@inproceedings{abusive_language_detection,
author = {Nobata, Chikashi and Tetreault, Joel and Thomas, Achint and Mehdad, Yashar and Chang, Yi},
title = {Abusive Language Detection in Online User Content},
year = {2016},
isbn = {9781450341431},
publisher = {International World Wide Web Conferences Steering Committee},
address = {Republic and Canton of Geneva, CHE},
url = {https://doi.org/10.1145/2872427.2883062},
doi = {10.1145/2872427.2883062},
booktitle = {Proceedings of the 25th International Conference on World Wide Web},
pages = {145–153},
numpages = {9},
location = {Montr\'{e}al, Qu\'{e}bec, Canada},
series = {WWW '16}
}

@inproceedings{blodgett-etal-2020-language,
    title = "Language (Technology) is Power: A Critical Survey of ``Bias'' in {NLP}",
    author = "Blodgett, Su Lin  and
      Barocas, Solon  and
      Daum{\'e} III, Hal  and
      Wallach, Hanna",
    editor = "Jurafsky, Dan  and
      Chai, Joyce  and
      Schluter, Natalie  and
      Tetreault, Joel",
    booktitle = "Proceedings of the 58th Annual Meeting of the Association for Computational Linguistics",
    month = jul,
    year = "2020",
    address = "Online",
    publisher = "Association for Computational Linguistics",
    url = "https://aclanthology.org/2020.acl-main.485/",
    doi = "10.18653/v1/2020.acl-main.485",
    pages = "5454--5476"
}

@inproceedings{are_explanations_helpful,
author = {Wang, Xinru and Yin, Ming},
title = {Are Explanations Helpful? A Comparative Study of the Effects of Explanations in AI-Assisted Decision-Making},
year = {2021},
isbn = {9781450380171},
publisher = {Association for Computing Machinery},
address = {New York, NY, USA},
url = {https://doi.org/10.1145/3397481.3450650},
doi = {10.1145/3397481.3450650},
booktitle = {Proceedings of the 26th International Conference on Intelligent User Interfaces},
pages = {318–328},
numpages = {11},
location = {College Station, TX, USA},
series = {IUI '21}
}

@inproceedings{jacovi-goldberg-2020-towards,
    title = "Towards Faithfully Interpretable {NLP} Systems: How Should We Define and Evaluate Faithfulness?",
    author = "Jacovi, Alon  and
      Goldberg, Yoav",
    editor = "Jurafsky, Dan  and
      Chai, Joyce  and
      Schluter, Natalie  and
      Tetreault, Joel",
    booktitle = "Proceedings of the 58th Annual Meeting of the Association for Computational Linguistics",
    month = jul,
    year = "2020",
    address = "Online",
    publisher = "Association for Computational Linguistics",
    url = "https://aclanthology.org/2020.acl-main.386/",
    doi = "10.18653/v1/2020.acl-main.386",
    pages = "4198--4205"
}

@article{human_interpretability_evaluation,
  author       = {Isaac Lage and
                  Emily Chen and
                  Jeffrey He and
                  Menaka Narayanan and
                  Been Kim and
                  Sam Gershman and
                  Finale Doshi{-}Velez},
  title        = {An Evaluation of the Human-Interpretability of Explanation},
  journal      = {CoRR},
  volume       = {abs/1902.00006},
  year         = {2019},
  url          = {http://arxiv.org/abs/1902.00006},
  eprinttype    = {arXiv},
  eprint       = {1902.00006},
  bibsource    = {dblp computer science bibliography, https://dblp.org}
}

@inproceedings{sahoo-etal-2022-detecting,
    title = "Detecting Unintended Social Bias in Toxic Language Datasets",
    author = "Sahoo, Nihar  and
      Gupta, Himanshu  and
      Bhattacharyya, Pushpak",
    editor = "Fokkens, Antske  and
      Srikumar, Vivek",
    booktitle = "Proceedings of the 26th Conference on Computational Natural Language Learning (CoNLL)",
    month = dec,
    year = "2022",
    address = "Abu Dhabi, United Arab Emirates (Hybrid)",
    publisher = "Association for Computational Linguistics",
    url = "https://aclanthology.org/2022.conll-1.10/",
    doi = "10.18653/v1/2022.conll-1.10",
    pages = "132--143"
}

@article{fair_hate_speech_detection,
  author       = {Aida Mostafazadeh Davani and
                  Ali Omrani and
                  Brendan Kennedy and
                  Mohammad Atari and
                  Xiang Ren and
                  Morteza Dehghani},
  title        = {Fair Hate Speech Detection through Evaluation of Social Group Counterfactuals},
  journal      = {CoRR},
  volume       = {abs/2010.12779},
  year         = {2020},
  url          = {https://arxiv.org/abs/2010.12779},
  eprinttype    = {arXiv},
  eprint       = {2010.12779},
  bibsource    = {dblp computer science bibliography, https://dblp.org}
}

@inproceedings{schafer-etal-2024-hierarchical,
    title = "Hierarchical Adversarial Correction to Mitigate Identity Term Bias in Toxicity Detection",
    author = {Sch{\"a}fer, Johannes  and
      Heid, Ulrich  and
      Klinger, Roman},
    editor = "De Clercq, Orph{\'e}e  and
      Barriere, Valentin  and
      Barnes, Jeremy  and
      Klinger, Roman  and
      Sedoc, Jo{\~a}o  and
      Tafreshi, Shabnam",
    booktitle = "Proceedings of the 14th Workshop on Computational Approaches to Subjectivity, Sentiment, {\&} Social Media Analysis",
    month = aug,
    year = "2024",
    address = "Bangkok, Thailand",
    publisher = "Association for Computational Linguistics",
    url = "https://aclanthology.org/2024.wassa-1.4/",
    doi = "10.18653/v1/2024.wassa-1.4",
    pages = "35--51"
}

@inproceedings{mamta-etal-2024-biaswipe,
    title = "{B}ias{W}ipe: Mitigating Unintended Bias in Text Classifiers through Model Interpretability",
    author = "Mamta, Mamta  and
      Chigrupaatii, Rishikant  and
      Ekbal, Asif",
    editor = "Al-Onaizan, Yaser  and
      Bansal, Mohit  and
      Chen, Yun-Nung",
    booktitle = "Proceedings of the 2024 Conference on Empirical Methods in Natural Language Processing",
    month = nov,
    year = "2024",
    address = "Miami, Florida, USA",
    publisher = "Association for Computational Linguistics",
    url = "https://aclanthology.org/2024.emnlp-main.1172/",
    doi = "10.18653/v1/2024.emnlp-main.1172",
    pages = "21059--21070"
}

@inproceedings{kim-etal-2022-hate,
    title = "Why Is It Hate Speech? Masked Rationale Prediction for Explainable Hate Speech Detection",
    author = "Kim, Jiyun  and
      Lee, Byounghan  and
      Sohn, Kyung-Ah",
    editor = "Calzolari, Nicoletta  and
      Huang, Chu-Ren  and
      Kim, Hansaem  and
      Pustejovsky, James  and
      Wanner, Leo  and
      Choi, Key-Sun  and
      Ryu, Pum-Mo  and
      Chen, Hsin-Hsi  and
      Donatelli, Lucia  and
      Ji, Heng  and
      Kurohashi, Sadao  and
      Paggio, Patrizia  and
      Xue, Nianwen  and
      Kim, Seokhwan  and
      Hahm, Younggyun  and
      He, Zhong  and
      Lee, Tony Kyungil  and
      Santus, Enrico  and
      Bond, Francis  and
      Na, Seung-Hoon",
    booktitle = "Proceedings of the 29th International Conference on Computational Linguistics",
    month = oct,
    year = "2022",
    address = "Gyeongju, Republic of Korea",
    publisher = "International Committee on Computational Linguistics",
    url = "https://aclanthology.org/2022.coling-1.577/",
    pages = "6644--6655"
}

@inproceedings{ecai_rationale_guided,
  author={Punyajoy Saha and Divyanshu Sheth and Kushal Kedia and Binny Mathew and Animesh Mukherjee},
  title={Rationale-Guided Few-Shot Classification to Detect Abusive Language},
  year={2023},
  cdate={1672531200000},
  pages={2041-2048},
  url={https://doi.org/10.3233/FAIA230497},
  booktitle={ECAI},
}

@inproceedings{nirmal-etal-2024-towards,
    title = "Towards Interpretable Hate Speech Detection using Large Language Model-extracted Rationales",
    author = "Nirmal, Ayushi  and
      Bhattacharjee, Amrita  and
      Sheth, Paras  and
      Liu, Huan",
    editor = {Chung, Yi-Ling  and
      Talat, Zeerak  and
      Nozza, Debora  and
      Plaza-del-Arco, Flor Miriam  and
      R{\"o}ttger, Paul  and
      Mostafazadeh Davani, Aida  and
      Calabrese, Agostina},
    booktitle = "Proceedings of the 8th Workshop on Online Abuse and Harms (WOAH 2024)",
    month = jun,
    year = "2024",
    address = "Mexico City, Mexico",
    publisher = "Association for Computational Linguistics",
    url = "https://aclanthology.org/2024.woah-1.17/",
    doi = "10.18653/v1/2024.woah-1.17",
    pages = "223--233"
}

@inproceedings{roy-etal-2023-probing,
    title = "Probing {LLM}s for hate speech detection: strengths and vulnerabilities",
    author = "Roy, Sarthak  and
      Harshvardhan, Ashish  and
      Mukherjee, Animesh  and
      Saha, Punyajoy",
    editor = "Bouamor, Houda  and
      Pino, Juan  and
      Bali, Kalika",
    booktitle = "Findings of the Association for Computational Linguistics: EMNLP 2023",
    month = dec,
    year = "2023",
    address = "Singapore",
    publisher = "Association for Computational Linguistics",
    url = "https://aclanthology.org/2023.findings-emnlp.407/",
    doi = "10.18653/v1/2023.findings-emnlp.407",
    pages = "6116--6128"
}

@inproceedings{wang-demberg-2024-rsa,
    title = "{RSA}-Control: A Pragmatics-Grounded Lightweight Controllable Text Generation Framework",
    author = "Wang, Yifan  and
      Demberg, Vera",
    editor = "Al-Onaizan, Yaser  and
      Bansal, Mohit  and
      Chen, Yun-Nung",
    booktitle = "Proceedings of the 2024 Conference on Empirical Methods in Natural Language Processing",
    month = nov,
    year = "2024",
    address = "Miami, Florida, USA",
    publisher = "Association for Computational Linguistics",
    url = "https://aclanthology.org/2024.emnlp-main.318/",
    doi = "10.18653/v1/2024.emnlp-main.318",
    pages = "5561--5582"
}

@article{
bcos-wang-2025,
title={B-cos {LM}: Efficiently Transforming Pre-trained Language Models for Improved Explainability},
author={Yifan Wang and Sukrut Rao and Ji-Ung Lee and Mayank Jobanputra and Vera Demberg},
journal={Transactions on Machine Learning Research},
issn={2835-8856},
year={2025},
url={https://openreview.net/forum?id=c180UH8Dg8},
note={}
}
\bibliographystyle{iclr2026_conference}

\appendix

\section{Fairness and Explainability in Hate Speech Detection}
\label{appendix:definition_and_motivation}

To better motivate our focus on fairness and explainability in the hate speech detection task, we provide additional background in this section. We begin by clarifying our definitions of hate speech and social bias, then review relevant work on fairness and explainability in hate speech detection. Finally, we explain why our study specifically focuses on input-based explanations.

\paragraph{Hate Speech} We follow~\citep{survey_on_automatic_detection} in defining hate speech as a specific form of abusive or toxic language that targets and attacks protected or identifiable social groups. Under this view, hate speech is a subset of abusive language. This definition is consistent with widely adopted formulations in prior work on hate speech detection (e.g.,~\citealp{abusive_language_detection, Davidson_Warmsley_Macy_Weber_2017}). Because our study focuses on fairness and, in particular, analyzes model behavior on examples involving specific social groups (race, gender, religion), this standard definition of hate speech aligns well with the scope and goals of our work. We still use the toxic vs. non-toxic labels following the terminology used in the Civil Comments~\citep{borkan2019nuanced} and Jigsaw~\citep{jigsaw-unintended-bias-in-toxicity-classification} datasets. Although these datasets include multiple subtypes of abusive content, they group them under the broader notion of toxicity.

\paragraph{Social Bias} 

Following the conceptualization of~\citep{blodgett-etal-2020-language}, we define social bias as the presence of stereotypical associations between social groups and certain attributes, as well as disparities in how these groups are treated as a result. Such biases can lead to both representational harms (e.g., demeaning or misrepresenting targeted groups) and allocational harms (e.g., unfair distribution of opportunities or resources). Given the potential for NLP systems to reproduce or amplify these harms, and their growing influence in everyday life, it is essential to detect and mitigate social bias in these models.

\paragraph{Social Bias in Hate Speech Detection}

Social bias has been widely documented in hate speech detection systems. \citet{dixon-measuring-and-mitigating-2018} showed that training data often contain uneven distributions of identity terms and stereotypical associations, which in turn propagates bias into downstream models. Subsequent studies revealed multiple dimensions of such disparities: \citet{sap-etal-2019-risk} demonstrated systematic dialectal prejudice against African-American English (AAE), while \citet{park-etal-2018-reducing} reported significant performance gaps across gendered identities. \citet{garg-etal-2019-counterfactual-fairness} further found that models frequently assign different toxicity labels to otherwise identical content when only the referenced social group is varied. \citet{sahoo-etal-2022-detecting} expanded the scope of this line of study by curating the ToxicBias dataset and examining bias across a broader set of social categories. More recently, \citet{roy-etal-2023-probing} found that LLMs exhibit similar bias in hate speech detection. Together, these studies underscore the persistence and multifaceted nature of social bias in hate speech detection.

To address such biases, a rich line of work has proposed mitigation techniques at different stages of the modeling pipeline. Pre-processing methods include debiasing word embeddings to reduce spurious associations between identity terms and toxicity~\citep{park-etal-2018-reducing}, re-sampling or re-weighting examples to obtain more balanced label distributions across identity groups~\citep{dixon-measuring-and-mitigating-2018}, and counterfactual data augmentation~\citep{park-etal-2018-reducing, garg-etal-2019-counterfactual-fairness}. In-processing approaches mostly modify the training objective, for instance by adding fairness-aware regularizers that penalize correlations between identity terms and toxic predictions~\citep{garg-etal-2019-counterfactual-fairness, fair_hate_speech_detection, attanasio-etal-2022-entropy, schafer-etal-2024-hierarchical}. Post-processing methods adjust model outputs without retraining~\citep{wang-demberg-2024-rsa}: threshold adjustment per group has been used to trade off subgroup false positive and false negative rates and reduce disparities~\citep{dixon-measuring-and-mitigating-2018}, while~\citet{mamta-etal-2024-biaswipe} identify neurons associated with biased behavior and prune or edit them to improve fairness. 

Despite substantial progress on identifying and mitigating social bias in hate speech detection, relatively little work has systematically explored whether model explanations can be leveraged to detect or reduce such biases.

\paragraph{Explainability in Hate Speech Detection} In parallel, there is a growing line of work on input-based explanations for hate speech detection. HateXplain~\citep{hatexplain} introduces a benchmark with human-annotated rationales and shows that models trained with rationale supervision improve both interpretability and reduce unintended bias towards target communities. Building on this, \citet{kim-etal-2022-hate} and \citet{ecai_rationale_guided} train models to jointly predict human rationales and toxicity labels, leading to more robust and explainable hate speech detection systems. More recent work further leverages LLM-generated rationales to supervise hate speech classifiers, achieving improved performance and interpretability~\citep{nirmal-etal-2024-towards}. 

However, while existing efforts focus primarily on improving hate speech detection performance, relatively little work examines whether and how input-based explanations can be systematically leveraged to improve fairness in hate speech detection models. Since fairness and explainability have both been extensively studied in this task, hate speech detection serves as an ideal setting for a thorough empirical examination of how these two dimensions interact in NLP models.

\paragraph{Input-Based Explanations} We focus on input-based explanations because they offer the most direct view into which parts of the input influence a model's prediction~\citep{are_explanations_helpful}, and they have long been regarded as central tools for fairness auditing in ML~\citep{balkir-etal-2022-challenges, deck-etal-2024-critical-survey}. Their methodological diversity also makes them an ideal testbed for our study, enabling a comprehensive examination of whether and how explanations can improve fairness~\citep{lyu-etal-2024-towards}. In addition, both automated and human-centric metrics for evaluating explanation properties (e.g., faithfulness, interpretability) are well established~\citep{deyoung-etal-2020-eraser, jacovi-goldberg-2020-towards, human_interpretability_evaluation}. This allows us to analyze how these properties relate to an explanation method's (in)effectiveness in fairness-related tasks. Finally, input-based explanations are often mandated by laws, such as the EU Artificial Intelligence Act, making it practically important to understand how their use interacts with fairness considerations.

\section{Details on Experimental Setups}
\label{appendix: experimental details}
\paragraph{Datasets and Models} 

\begin{table}[!ht]
    \caption{Splits for the Civil Comments and Jigsaw datasets used in this work. The instances are sampled randomly from the original dataset.\\}
    \centering
    \begin{tabular}{lrrrrrr}
    \toprule
    Split & \multicolumn{3}{c}{Civil Comments} & \multicolumn{3}{c}{Jigsaw} \\
    & race & gender & religion & race & gender & religion \\
    \midrule
    Test & 2,000 & 2,000 & 1,000 & 400 & 800 & 200 \\
    Training & 8,000 & 8,000 & 6,300 & 8,000 & 8,000 & 6,300 \\
    \bottomrule
    \end{tabular}
    \label{tab:appendix dataset statistics}
\end{table}

Table~\ref{tab:appendix dataset statistics} shows the dataset statistics for the Civil Comments\footnote{\href{https://huggingface.co/datasets/lighteval/civil_comments_helm}{https://huggingface.co/datasets/lighteval/civil\_comments\_helm}} and Jigsaw\footnote{\href{https://huggingface.co/datasets/google/jigsaw_unintended_bias}{https://huggingface.co/datasets/google/jigsaw\_unintended\_bias}} datasets, respectively. 
When constructing subsets for each social group, we include only examples that contain words from the corresponding group’s vocabulary and exclude those containing words from other groups. Additionally, we remove examples that include insulting references to social groups.

For encoder-only models, we either train separate models for each bias type or a single model on the combined set of all bias types. The latter are marked as (all) in the results.
Models are trained for 5 epochs with a batch size of 8, a learning rate of $2\times10^{-5}$, and 10\% warm-up steps. 
We implement debiasing methods of encoder-only models ourselves.
Debiased models are trained on the same number of examples for comparability, therefore we do not sample more examples from external datasets for causal debias. 
All experiments are carried out using a maximum of 4x NVIDIA H100 GPUs.

We use the base version of BERT and RoBERTa models for our encoder-only experiments, and for decoder-only experiments, we adopt the instruct versions of the models from HuggingFace.

\paragraph{Prompts}

We employ the following prompt for zero-shot and few-shot prompting with decoder-only models.

\begin{verbatim}
Task: Determine whether the following sentence contains any toxic 
content. Toxicity includes insults, threats, hate speech, or any 
language that is likely to offend or harm others.

([EXAMPLES]) 
(Now, please analyze the following sentence and provide your 
answer. Answer only Yes or No.)

Sentence: [TEST EXAMPLE]

Question: Does this sentence contain toxicity?

Answer (Yes or No):
\end{verbatim}
We use one toxic and one non-toxic example for each social group associated with a bias type to construct the corresponding few-shot prompt.
For fairness imagination, we prepend the task instruction with:
\begin{verbatim}
Imagine a world with no bias regarding [BIAS TYPE].
\end{verbatim}
For fairness instruction, we add:
\begin{verbatim}
In the following, do not rely on stereotypes or biases regarding 
[BIAS TYPE] to make predictions.
\end{verbatim}

\paragraph{Input-Based Explanation Methods}

We implement Attention, Attention Rollout, Attention Flow, and Progressive Inference ourselves, and use the official codebase from the DecompX paper for its implementation. 
For other explanation methods, we generate them using the Captum API\footnote{\href{https://captum.ai/api/}{https://captum.ai/api/}}
Single-pass Progressive Inference is used for simplicity.

\paragraph{Explanation-Based Bias Mitigation}

We reimplement all explanation methods used in RQ3 to enable gradient tracking during training and ensure consistency with the explanations provided by the Captum API. 
For Occlusion, we simplify computation by penalizing the probability difference between the original input and the input with all sensitive tokens masked.

\paragraph{LLM decision and LLM attribution}

To elicit bias judgments from LLMs, we append the following instructions to their predictions.

\textbf{LLM decision:}
\begin{verbatim}
Do not re-answer the task. Instead, reflect only on your previous 
answer.

Question: Was your previous answer influenced by stereotypes or 
biases related to [BIAS TYPE]?

Answer (Yes or No):
\end{verbatim}

\textbf{LLM attribution:}
\begin{verbatim}
Now select exactly [NUM_TOKENS] words from the input sentence that 
most influenced your decision. List the words separated by commas. 
Please begin directly with the selected word.

Selected words:
\end{verbatim}

\section{Explanation Efficiency}\label{appendix:explanation-efficiency}

Table~\ref{tab: expenses} reports the time and GPU memory costs for each explanation method. Most post-hoc explanation methods are lightweight when applied to BERT, whereas IntGrad, Occlusion, and KernelSHAP require substantially more time and computational resources when generating explanations for LLMs.

\begin{table}[h]
    \centering
    \caption{Computational costs per example for generating explanations across 200 instances on BERT and Qwen3-4B.
Results are computed on the race subset of the Civil Comments dataset using a batch size of 1 and are averaged over three runs.
All methods are run on a single 80-GB H100 GPU, except Integrated Gradients, which uses two H100 GPUs with gradient checkpointing to reduce memory usage.
Because explanation methods within the same family incur similar computational costs, we report each family only once.\\}
    \resizebox{0.75\linewidth}{!}{
    \begin{tabular}{lcclcc}
    \toprule
    \multicolumn{3}{c}{\textbf{BERT}} & \multicolumn{3}{c}{\textbf{Qwen3-4B}} \\
    \textbf{Method} & \textbf{Time (s/example)} & \textbf{Memory (GB)} & \textbf{Method} & \textbf{Time (s/example)} & \textbf{Memory (GB)} \\
    \midrule
    Attention & 0.027 & 0.529 & Attention & 0.112 & 16.598 \\
    Grad & 0.026 & 0.603 & Grad & 0.237 & 19.631 \\
    IxG & 0.025 & 0.603 & IxG & 0.236 & 19.631 \\
    IntGrad & 0.064 & 7.010 & IntGrad & 1.784 & 101.694 \\
    DeepLift & 0.027 & 0.748 & DeepLift & 0.323 & 23.530 \\
    Occlusion & 0.330 & 0.508 & Occlusion & 4.204 & 15.639 \\
    KernelSHAP & 0.138 & 0.508 & KernelSHAP & 1.374 & 20.013 \\
    DecompX & 0.074 & 7.222 & ProgInfer & 0.068 & 15.810 \\
    \bottomrule
         
    \end{tabular}
    }
    \label{tab: expenses}
\end{table}

\section{Model Performance and Fairness Results}
\label{appendix: performance and fairness}

\begin{table*}[!h]
    \centering
         \caption{Task performance and fairness of default and debiased models on the Civil Comments dataset.
     Results are provided for race/gender/religion biases.
     \textcolor{forestgreen}{Green} (\textcolor{red}{red}) indicates the results are \textcolor{forestgreen}{better} (\textcolor{red}{worse}) than the default/zero-shot models.
     \textit{All} indicates the model is trained on data containing all bias types.\\}
     \resizebox{\textwidth}{!}{
     \begin{tabular}{llccccc}
     \toprule
        Model  & Method & Accuracy ($\uparrow$) & $\text{Disp}_\text{acc} (\downarrow)$ & $\text{Disp}_\text{fpr} (\downarrow)$ & $\text{Disp}_\text{fnr} (\downarrow)$ & $\text{Avg}_\text{iu} (\downarrow)$ \\
    \midrule
    \multirow{7}{*}{\shortstack{BERT}}& Default & 78.38/88.05/85.93 & 2.05/3.30/18.07 & 0.50/0.03/5.77 & 10.04/11.98/30.90 & 3.17/0.66/1.27 \\
 & Group balance & \textcolor{forestgreen}{79.25}/\textcolor{red}{87.25}/\textcolor{forestgreen}{86.83} & \textcolor{red}{3.10}/\textcolor{forestgreen}{2.80}/\textcolor{forestgreen}{13.53} & \textcolor{forestgreen}{0.25}/\textcolor{red}{1.73}/\textcolor{red}{11.53} & \textcolor{red}{10.46}/\textcolor{forestgreen}{5.38}/\textcolor{forestgreen}{30.31} & \textcolor{red}{3.79}/\textcolor{forestgreen}{0.42}/\textcolor{red}{2.01} \\
 & Group-class balance & \textcolor{red}{78.00}/\textcolor{red}{87.02}/\textcolor{red}{85.77} & \textcolor{forestgreen}{1.80}/\textcolor{forestgreen}{2.75}/\textcolor{forestgreen}{14.73} & \textcolor{red}{2.42}/\textcolor{red}{0.99}/\textcolor{forestgreen}{3.09} & \textcolor{red}{10.63}/\textcolor{forestgreen}{7.26}/\textcolor{red}{33.14} & \textcolor{red}{4.43}/\textcolor{red}{0.98}/\textcolor{forestgreen}{0.71} \\
 & CDA & \textcolor{red}{76.83}/\textcolor{red}{86.70}/\textcolor{red}{84.83} & \textcolor{red}{2.35}/\textcolor{red}{3.60}/\textcolor{forestgreen}{14.13} & \textcolor{red}{5.88}/\textcolor{red}{2.00}/\textcolor{forestgreen}{5.67} & \textcolor{red}{18.45}/\textcolor{forestgreen}{7.57}/\textcolor{forestgreen}{24.12} & \textcolor{forestgreen}{0.50}/\textcolor{forestgreen}{0.50}/\textcolor{forestgreen}{0.90} \\
 & Dropout & \textcolor{forestgreen}{78.53}/\textcolor{forestgreen}{88.20}/\textcolor{red}{85.03} & \textcolor{red}{2.25}/\textcolor{forestgreen}{2.10}/\textcolor{forestgreen}{15.67} & \textcolor{red}{0.78}/\textcolor{red}{1.46}/\textcolor{red}{5.93} & \textcolor{red}{10.82}/\textcolor{forestgreen}{3.50}/\textcolor{forestgreen}{27.16} & \textcolor{red}{3.43}/\textcolor{forestgreen}{0.52}/\textcolor{red}{1.51} \\
 & Attention entropy & \textcolor{forestgreen}{79.15}/\textcolor{red}{87.67}/\textcolor{red}{84.93} & \textcolor{red}{2.60}/\textcolor{forestgreen}{2.05}/\textcolor{forestgreen}{15.07} & \textcolor{red}{0.99}/\textcolor{red}{0.10}/\textcolor{forestgreen}{4.99} & \textcolor{red}{11.71}/\textcolor{forestgreen}{7.11}/\textcolor{forestgreen}{26.52} & \textcolor{forestgreen}{2.95}/\textcolor{red}{0.67}/\textcolor{red}{1.58} \\
 & Causal debias & \textcolor{forestgreen}{78.80}/\textcolor{red}{86.17}/\textcolor{forestgreen}{86.40} & \textcolor{forestgreen}{0.00}/\textcolor{forestgreen}{2.65}/\textcolor{forestgreen}{16.40} & \textcolor{red}{3.90}/\textcolor{red}{0.46}/\textcolor{red}{8.82} & \textcolor{forestgreen}{7.98}/\textcolor{forestgreen}{10.67}/\textcolor{forestgreen}{30.46} & \textcolor{red}{3.83}/\textcolor{forestgreen}{0.48}/\textcolor{red}{2.10} \\
 \midrule

    \multirow{7}{*}{\shortstack{BERT\\ (all)}} & Default & 78.30/88.20/87.43 & 2.00/3.20/13.47 & 0.02/1.11/6.24 & 8.44/8.58/23.53 & 3.99/0.96/1.76 \\
     & Group balance & \textcolor{forestgreen}{79.05}/\textcolor{forestgreen}{88.85}/\textcolor{forestgreen}{87.47} & \textcolor{red}{3.50}/\textcolor{forestgreen}{2.80}/\textcolor{red}{13.67} & \textcolor{red}{1.72}/\textcolor{forestgreen}{0.31}/\textcolor{red}{6.92} & \textcolor{red}{8.83}/\textcolor{red}{11.08}/\textcolor{red}{23.91} & \textcolor{red}{4.13}/\textcolor{red}{1.17}/\textcolor{red}{2.15}   \\
     & Group-class balance & \textcolor{red}{78.17}/\textcolor{forestgreen}{88.25}/\textcolor{red}{86.90} & \textcolor{forestgreen}{1.95}/\textcolor{forestgreen}{1.70}/\textcolor{red}{14.60} & \textcolor{red}{1.35}/\textcolor{forestgreen}{0.51}/\textcolor{red}{8.52} & \textcolor{red}{9.33}/\textcolor{forestgreen}{4.66}/\textcolor{red}{33.13} & \textcolor{red}{4.83}/\textcolor{forestgreen}{0.93}/\textcolor{forestgreen}{1.37}   \\
     & CDA & \textcolor{red}{78.08}/\textcolor{red}{87.70}/\textcolor{red}{86.83} & \textcolor{red}{2.65}/\textcolor{forestgreen}{2.70}/\textcolor{red}{14.33} & \textcolor{red}{6.38}/\textcolor{forestgreen}{1.05}/\textcolor{forestgreen}{4.70} & \textcolor{red}{20.35}/\textcolor{forestgreen}{6.92}/\textcolor{red}{30.23} & \textcolor{forestgreen}{0.60}/\textcolor{forestgreen}{0.46}/\textcolor{forestgreen}{0.71}   \\
     & Dropout & \textcolor{red}{78.08}/\textcolor{red}{87.60}/\textcolor{forestgreen}{87.67} & \textcolor{red}{2.45}/\textcolor{forestgreen}{3.10}/13.47 & \textcolor{red}{0.30}/\textcolor{forestgreen}{1.05}/\textcolor{forestgreen}{5.53} & \textcolor{red}{9.99}/\textcolor{forestgreen}{8.39}/\textcolor{red}{33.12} & \textcolor{forestgreen}{3.60}/\textcolor{forestgreen}{0.89}/\textcolor{forestgreen}{1.59}   \\
     & Attention entropy & \textcolor{forestgreen}{78.35}/\textcolor{red}{87.90}/\textcolor{forestgreen}{87.77} & \textcolor{red}{2.10}/\textcolor{forestgreen}{2.30}/\textcolor{forestgreen}{11.67} & \textcolor{red}{1.28}/\textcolor{forestgreen}{0.10}/\textcolor{red}{6.55} & \textcolor{forestgreen}{5.92}/\textcolor{forestgreen}{8.01}/\textcolor{red}{36.15} & \textcolor{red}{4.98}/0.96/\textcolor{red}{2.10} 
   \\
     & Causal debias & \textcolor{forestgreen}{79.40}/\textcolor{forestgreen}{88.75}/\textcolor{forestgreen}{87.70} & \textcolor{red}{2.20}/\textcolor{forestgreen}{2.60}/\textcolor{forestgreen}{12.60} & \textcolor{red}{2.51}/\textcolor{forestgreen}{0.70}/\textcolor{red}{6.70} & \textcolor{red}{13.13}/\textcolor{forestgreen}{7.44}/\textcolor{red}{31.28} & \textcolor{forestgreen}{3.54}/\textcolor{forestgreen}{0.80}/\textcolor{red}{2.12} 
  \\
    \midrule
    \multirow{7}{*}{\shortstack{RoBERTa}}  & Default & 78.50/88.33/85.23 & 2.80/2.05/17.07 & 2.84/1.66/6.59 & 15.46/2.78/31.64 & 2.56/0.60/1.55 \\
 & Group balance & \textcolor{red}{78.25}/\textcolor{forestgreen}{88.50}/\textcolor{forestgreen}{87.03} & \textcolor{forestgreen}{2.00}/\textcolor{red}{2.20}/\textcolor{forestgreen}{16.93} & \textcolor{forestgreen}{2.10}/\textcolor{forestgreen}{1.27}/\textcolor{red}{11.36} & \textcolor{forestgreen}{9.85}/\textcolor{red}{4.57}/\textcolor{forestgreen}{29.48} & \textcolor{red}{3.95}/\textcolor{red}{0.68}/\textcolor{forestgreen}{1.19} \\
 & Group-class balance & \textcolor{forestgreen}{78.57}/\textcolor{red}{84.50}/\textcolor{red}{83.60} & \textcolor{forestgreen}{1.65}/\textcolor{red}{2.30}/\textcolor{red}{18.80} & \textcolor{red}{3.31}/\textcolor{forestgreen}{0.76}/\textcolor{forestgreen}{3.89} & \textcolor{forestgreen}{12.91}/\textcolor{red}{5.82}/\textcolor{red}{38.88} & \textcolor{red}{3.28}/\textcolor{forestgreen}{0.42}/\textcolor{forestgreen}{0.87} \\
 & CDA & \textcolor{red}{76.75}/\textcolor{red}{87.58}/\textcolor{red}{85.20} & \textcolor{forestgreen}{1.60}/\textcolor{forestgreen}{1.75}/\textcolor{forestgreen}{14.20} & \textcolor{red}{6.37}/\textcolor{forestgreen}{0.31}/\textcolor{forestgreen}{4.10} & \textcolor{red}{15.91}/\textcolor{red}{5.41}/\textcolor{red}{35.70} & \textcolor{forestgreen}{0.82}/\textcolor{forestgreen}{0.42}/\textcolor{forestgreen}{1.19} \\
 & Dropout & \textcolor{red}{78.33}/\textcolor{forestgreen}{88.92}/\textcolor{forestgreen}{86.73} & \textcolor{forestgreen}{2.15}/\textcolor{forestgreen}{1.55}/\textcolor{forestgreen}{14.53} & \textcolor{forestgreen}{2.42}/\textcolor{forestgreen}{0.58}/\textcolor{red}{8.86} & \textcolor{forestgreen}{11.11}/\textcolor{red}{3.96}/\textcolor{forestgreen}{27.05} & \textcolor{red}{4.08}/\textcolor{forestgreen}{0.56}/\textcolor{red}{2.10} \\
 & Attention entropy & \textcolor{red}{78.33}/\textcolor{forestgreen}{88.42}/\textcolor{forestgreen}{86.67} & \textcolor{forestgreen}{1.75}/\textcolor{forestgreen}{1.75}/\textcolor{forestgreen}{15.73} & \textcolor{red}{2.89}/\textcolor{forestgreen}{0.23}/\textcolor{red}{9.23} & \textcolor{forestgreen}{10.91}/\textcolor{red}{5.60}/\textcolor{forestgreen}{24.68} & \textcolor{red}{3.82}/\textcolor{red}{0.69}/\textcolor{red}{1.75} \\
 & Causal debias & \textcolor{forestgreen}{78.83}/\textcolor{red}{87.52}/\textcolor{forestgreen}{86.00} & \textcolor{forestgreen}{2.65}/\textcolor{red}{2.45}/\textcolor{forestgreen}{15.60} & \textcolor{forestgreen}{1.48}/\textcolor{forestgreen}{0.85}/\textcolor{red}{10.56} & \textcolor{forestgreen}{11.34}/\textcolor{red}{6.51}/\textcolor{forestgreen}{30.14} & \textcolor{red}{4.06}/\textcolor{forestgreen}{0.56}/\textcolor{forestgreen}{1.34} \\
    \midrule
    \multirow{7}{*}{\shortstack{RoBERTa\\ (all)}}  & Default & 78.88/88.70/87.90 & 2.95/2.40/13.80 & 2.24/0.58/9.50 & 13.55/7.19/33.47 & 4.14/0.95/2.35 \\
 & Group balance & \textcolor{forestgreen}{79.30}/\textcolor{red}{88.65}/\textcolor{forestgreen}{87.93} & \textcolor{forestgreen}{2.90}/\textcolor{forestgreen}{2.00}/\textcolor{red}{14.73} & \textcolor{forestgreen}{1.27}/\textcolor{forestgreen}{0.17}/\textcolor{red}{12.30} & \textcolor{forestgreen}{11.03}/\textcolor{red}{7.74}/\textcolor{forestgreen}{31.69} & \textcolor{red}{5.02}/\textcolor{red}{1.06}/\textcolor{red}{2.80} \\
 & Group-class balance & \textcolor{forestgreen}{79.40}/\textcolor{forestgreen}{89.15}/\textcolor{forestgreen}{87.93} & \textcolor{forestgreen}{1.70}/\textcolor{forestgreen}{1.10}/\textcolor{forestgreen}{12.73} & \textcolor{red}{4.43}/\textcolor{forestgreen}{0.24}/\textcolor{forestgreen}{5.08} & \textcolor{red}{13.65}/\textcolor{forestgreen}{3.24}/\textcolor{forestgreen}{25.90} & \textcolor{red}{4.17}/\textcolor{forestgreen}{0.75}/\textcolor{forestgreen}{1.58} \\
 & CDA & \textcolor{red}{77.75}/\textcolor{red}{88.25}/\textcolor{red}{86.90} & \textcolor{forestgreen}{2.50}/\textcolor{forestgreen}{2.00}/13.80 & \textcolor{red}{5.93}/\textcolor{red}{1.25}/\textcolor{forestgreen}{6.33} & \textcolor{red}{18.80}/\textcolor{forestgreen}{3.71}/\textcolor{forestgreen}{22.62} & \textcolor{forestgreen}{1.13}/\textcolor{forestgreen}{0.55}/\textcolor{forestgreen}{1.18} \\
 & Dropout & 78.88/\textcolor{red}{88.40}/\textcolor{red}{87.70} & \textcolor{forestgreen}{2.75}/\textcolor{red}{3.00}/\textcolor{red}{14.80} & \textcolor{forestgreen}{1.80}/\textcolor{red}{1.33}/\textcolor{forestgreen}{6.66} & \textcolor{forestgreen}{12.46}/\textcolor{red}{7.34}/\textcolor{forestgreen}{33.39} & \textcolor{red}{4.26}/\textcolor{red}{0.99}/\textcolor{forestgreen}{2.13} \\
 & Attention entropy & \textcolor{red}{78.80}/\textcolor{forestgreen}{88.72}/\textcolor{red}{87.83} & \textcolor{forestgreen}{2.10}/\textcolor{forestgreen}{2.15}/\textcolor{forestgreen}{13.53} & \textcolor{red}{2.64}/\textcolor{red}{1.33}/\textcolor{forestgreen}{7.55} & \textcolor{forestgreen}{11.31}/\textcolor{forestgreen}{4.18}/\textcolor{forestgreen}{28.68} & \textcolor{red}{4.46}/\textcolor{red}{1.09}/\textcolor{red}{2.57} \\
 & Causal debias & \textcolor{forestgreen}{79.27}/\textcolor{forestgreen}{89.78}/\textcolor{red}{87.80} & \textcolor{red}{3.35}/\textcolor{forestgreen}{1.25}/\textcolor{red}{15.00} & \textcolor{red}{3.24}/\textcolor{forestgreen}{0.51}/\textcolor{red}{11.86} & \textcolor{red}{16.00}/\textcolor{forestgreen}{3.05}/\textcolor{red}{37.57} & \textcolor{forestgreen}{3.56}/\textcolor{forestgreen}{0.74}/\textcolor{red}{2.70} \\
 \midrule

     \multirow{4}{*}{\shortstack{Llama3.2-3B-Instruct}} & Zero-shot & 63.78/74.62/71.27 & 1.45/2.35/24.67 & 11.03/3.52/36.81 & 10.54/1.03/2.95 & 2.13/2.94/3.83  \\
 & Few-shot & \textcolor{forestgreen}{67.80}/\textcolor{forestgreen}{79.80}/\textcolor{forestgreen}{80.10} & \textcolor{red}{1.60}/\textcolor{forestgreen}{1.70}/\textcolor{forestgreen}{18.20} & \textcolor{forestgreen}{2.49}/\textcolor{forestgreen}{0.08}/\textcolor{forestgreen}{6.73} & \textcolor{forestgreen}{6.73}/\textcolor{red}{6.05}/\textcolor{red}{10.77} & \textcolor{forestgreen}{1.39}/\textcolor{forestgreen}{2.05}/\textcolor{forestgreen}{1.90}  \\
 & Fairness imagination & \textcolor{forestgreen}{64.95}/\textcolor{forestgreen}{75.92}/\textcolor{forestgreen}{73.37} & \textcolor{forestgreen}{0.80}/\textcolor{forestgreen}{0.85}/\textcolor{forestgreen}{21.87} & \textcolor{forestgreen}{8.70}/\textcolor{red}{3.61}/\textcolor{forestgreen}{32.54} & \textcolor{forestgreen}{9.44}/\textcolor{red}{6.79}/\textcolor{red}{5.98} & \textcolor{red}{2.65}/\textcolor{red}{3.58}/\textcolor{forestgreen}{3.50}  \\
 & Fairness instruction & \textcolor{forestgreen}{65.90}/\textcolor{forestgreen}{76.95}/\textcolor{forestgreen}{78.07} & \textcolor{red}{2.60}/\textcolor{forestgreen}{1.70}/\textcolor{forestgreen}{21.53} & \textcolor{forestgreen}{1.89}/\textcolor{forestgreen}{0.39}/\textcolor{forestgreen}{7.00} & \textcolor{forestgreen}{3.79}/\textcolor{red}{6.35}/\textcolor{red}{4.24} & \textcolor{forestgreen}{1.35}/\textcolor{forestgreen}{1.13}/\textcolor{forestgreen}{1.71}  \\
 \midrule
 \multirow{4}{*}{\shortstack{Qwen3-4B}}& Zero-shot & 69.55/79.75/77.50 & 0.60/0.00/17.40 & 7.13/1.40/21.07 & 13.25/3.71/5.17 & 2.55/2.41/3.32  \\
 & Few-shot & \textcolor{forestgreen}{70.15}/\textcolor{forestgreen}{80.73}/\textcolor{forestgreen}{79.53} & \textcolor{red}{1.80}/\textcolor{red}{0.65}/\textcolor{red}{18.93} & \textcolor{red}{10.02}/\textcolor{red}{2.50}/\textcolor{forestgreen}{19.31} & \textcolor{forestgreen}{11.89}/\textcolor{red}{9.15}/\textcolor{red}{5.57} & \textcolor{red}{3.18}/\textcolor{red}{3.34}/\textcolor{red}{3.76}  \\
 & Fairness imagination & \textcolor{forestgreen}{71.23}/\textcolor{forestgreen}{80.40}/\textcolor{forestgreen}{80.83} & \textcolor{red}{0.85}/\textcolor{red}{1.00}/\textcolor{red}{18.27} & \textcolor{forestgreen}{4.03}/\textcolor{red}{2.11}/\textcolor{forestgreen}{10.51} & \textcolor{forestgreen}{11.62}/\textcolor{red}{9.21}/\textcolor{forestgreen}{4.28} & \textcolor{red}{2.98}/\textcolor{red}{3.16}/\textcolor{forestgreen}{2.20}  \\
 & Fairness instruction & \textcolor{forestgreen}{70.40}/\textcolor{forestgreen}{79.77}/\textcolor{forestgreen}{80.47} & 0.60/\textcolor{red}{1.35}/\textcolor{red}{19.33} & \textcolor{forestgreen}{4.30}/\textcolor{forestgreen}{0.39}/\textcolor{forestgreen}{4.67} & \textcolor{forestgreen}{11.11}/\textcolor{red}{5.24}/\textcolor{forestgreen}{5.08} & \textcolor{forestgreen}{2.02}/\textcolor{forestgreen}{1.83}/\textcolor{forestgreen}{1.71}  \\
 \midrule
 \multirow{4}{*}{\shortstack{Qwen3-8B}}& Zero-shot & 59.27/69.23/66.30 & 1.25/0.15/26.80 & 8.18/0.07/42.05 & 4.65/0.80/3.02 & 3.27/3.40/4.74  \\
 & Few-shot & \textcolor{forestgreen}{66.97}/\textcolor{forestgreen}{77.30}/\textcolor{forestgreen}{77.47} & \textcolor{forestgreen}{0.05}/\textcolor{forestgreen}{0.00}/\textcolor{forestgreen}{23.27} & \textcolor{forestgreen}{6.14}/\textcolor{red}{2.73}/\textcolor{forestgreen}{29.51} & \textcolor{red}{7.95}/\textcolor{red}{7.64}/\textcolor{forestgreen}{2.34} & \textcolor{red}{4.23}/\textcolor{red}{4.58}/\textcolor{red}{5.96}  \\
 & Fairness imagination & \textcolor{forestgreen}{62.10}/\textcolor{forestgreen}{72.92}/\textcolor{forestgreen}{69.97} & \textcolor{red}{1.60}/\textcolor{red}{0.55}/\textcolor{forestgreen}{21.87} & \textcolor{forestgreen}{7.80}/\textcolor{red}{2.62}/\textcolor{forestgreen}{32.27} & \textcolor{forestgreen}{4.28}/\textcolor{red}{5.42}/\textcolor{red}{9.43} & \textcolor{forestgreen}{2.54}/\textcolor{forestgreen}{2.08}/\textcolor{forestgreen}{2.58}  \\
 & Fairness instruction & \textcolor{forestgreen}{66.50}/\textcolor{forestgreen}{75.15}/\textcolor{forestgreen}{73.90} & \textcolor{forestgreen}{0.90}/\textcolor{forestgreen}{0.10}/\textcolor{forestgreen}{21.20} & \textcolor{forestgreen}{8.03}/\textcolor{red}{1.76}/\textcolor{forestgreen}{28.60} & \textcolor{red}{8.79}/\textcolor{red}{4.59}/\textcolor{red}{7.45} & \textcolor{forestgreen}{2.45}/\textcolor{forestgreen}{2.95}/\textcolor{forestgreen}{3.15}  \\

    \bottomrule
     \end{tabular}
     }

     \label{tab:appendix civil performance fairness}
\end{table*}

\begin{table*}[!h]
    \centering
         \caption{Task performance and fairness results of default and debiased models on the Jigsaw dataset.
     Results are provided for race/gender/religion biases.
     \textcolor{forestgreen}{Green} (\textcolor{red}{red}) indicates the results are \textcolor{forestgreen}{better} (\textcolor{red}{worse}) than the default/zero-shot models.
     \textit{All} indicates the model is trained on data containing all bias types.\\}
     \resizebox{\textwidth}{!}{
     \begin{tabular}{llccccc}
     \toprule
        Model  & Method & Accuracy ($\uparrow$) & $\text{Disp}_\text{acc} (\downarrow)$ & $\text{Disp}_\text{fpr} (\downarrow)$ & $\text{Disp}_\text{fnr} (\downarrow)$ & $\text{Avg}_\text{iu} (\downarrow)$ \\
    \midrule
    \multirow{7}{*}{\shortstack{BERT}}  & Default & 85.50/93.00/90.50 & 0.50/2.25/6.00 & 0.64/2.34/5.22 & 0.70/3.28/21.54 & 2.02/0.36/1.33 \\
 & Group balance & \textcolor{red}{84.88}/\textcolor{red}{92.75}/\textcolor{red}{89.67} & \textcolor{red}{2.75}/\textcolor{forestgreen}{1.00}/\textcolor{red}{10.67} & \textcolor{red}{1.28}/\textcolor{forestgreen}{0.82}/\textcolor{forestgreen}{3.90} & \textcolor{red}{7.77}/\textcolor{red}{4.56}/\textcolor{red}{38.29} & \textcolor{forestgreen}{1.90}/0.36/\textcolor{forestgreen}{0.67} \\
 & Group-class balance & \textcolor{red}{84.38}/\textcolor{red}{92.81}/\textcolor{forestgreen}{90.83} & \textcolor{forestgreen}{0.25}/\textcolor{forestgreen}{0.62}/\textcolor{red}{6.33} & \textcolor{red}{1.58}/\textcolor{forestgreen}{0.15}/\textcolor{forestgreen}{1.98} & \textcolor{red}{8.03}/\textcolor{red}{9.64}/\textcolor{red}{43.57} & \textcolor{forestgreen}{0.97}/\textcolor{red}{0.65}/\textcolor{forestgreen}{0.34} \\
 & CDA & \textcolor{red}{85.25}/\textcolor{red}{91.81}/90.50 & \textcolor{red}{4.00}/\textcolor{red}{3.63}/\textcolor{red}{10.00} & \textcolor{red}{4.12}/\textcolor{red}{3.44}/\textcolor{forestgreen}{5.10} & \textcolor{red}{2.97}/\textcolor{red}{7.38}/\textcolor{red}{37.39} & \textcolor{forestgreen}{0.39}/\textcolor{forestgreen}{0.28}/\textcolor{forestgreen}{0.45} \\
 & Dropout & \textcolor{forestgreen}{85.62}/\textcolor{red}{92.69}/\textcolor{red}{89.83} & \textcolor{red}{1.25}/\textcolor{red}{3.37}/\textcolor{red}{9.67} & \textcolor{forestgreen}{0.31}/\textcolor{red}{3.03}/\textcolor{red}{5.46} & \textcolor{red}{6.51}/\textcolor{red}{8.41}/\textcolor{red}{27.37} & \textcolor{red}{2.75}/0.36/\textcolor{forestgreen}{1.00} \\
 & Attention entropy & \textcolor{red}{85.00}/\textcolor{red}{92.06}/\textcolor{red}{89.83} & \textcolor{forestgreen}{0.00}/\textcolor{red}{3.12}/\textcolor{red}{9.33} & \textcolor{forestgreen}{0.62}/\textcolor{red}{3.03}/\textcolor{forestgreen}{4.29} & \textcolor{red}{1.72}/\textcolor{red}{6.00}/\textcolor{red}{28.06} & \textcolor{red}{2.93}/\textcolor{red}{0.50}/\textcolor{forestgreen}{0.98} \\
 & Causal debias & 85.50/\textcolor{forestgreen}{93.38}/\textcolor{red}{89.83} & \textcolor{red}{4.00}/\textcolor{forestgreen}{0.75}/\textcolor{red}{7.33} & \textcolor{red}{1.28}/\textcolor{forestgreen}{0.28}/\textcolor{forestgreen}{3.55} & \textcolor{red}{13.73}/\textcolor{red}{12.77}/\textcolor{forestgreen}{17.12} & \textcolor{red}{3.16}/\textcolor{red}{0.43}/\textcolor{forestgreen}{1.10} \\
    
    \midrule
    \multirow{7}{*}{\shortstack{BERT\\ (all)}}  & Default & 85.62/93.19/90.33 & 1.25/1.12/9.33 & 1.59/1.51/4.65 & 12.69/0.36/21.76 & 1.30/0.33/1.18 \\
 & Group balance & \textcolor{red}{83.38}/93.19/\textcolor{red}{90.17} & \textcolor{red}{1.75}/1.12/\textcolor{red}{9.67} & \textcolor{forestgreen}{1.56}/\textcolor{forestgreen}{1.10}/\textcolor{red}{4.66} & \textcolor{forestgreen}{3.10}/\textcolor{red}{3.23}/\textcolor{red}{26.79} & \textcolor{red}{2.81}/\textcolor{red}{0.40}/\textcolor{forestgreen}{0.76} \\
 & Group-class balance & \textcolor{red}{84.88}/\textcolor{red}{92.94}/\textcolor{red}{90.00} & 1.25/\textcolor{forestgreen}{0.87}/\textcolor{red}{10.00} & \textcolor{forestgreen}{1.27}/\textcolor{forestgreen}{0.41}/\textcolor{forestgreen}{2.09} & \textcolor{forestgreen}{0.37}/\textcolor{red}{7.49}/\textcolor{red}{58.07} & \textcolor{forestgreen}{1.29}/\textcolor{forestgreen}{0.28}/\textcolor{forestgreen}{0.47} \\
 & CDA & 85.62/\textcolor{red}{92.19}/\textcolor{red}{90.00} & \textcolor{red}{3.25}/\textcolor{red}{1.88}/\textcolor{forestgreen}{7.00} & \textcolor{red}{2.86}/\textcolor{red}{1.78}/\textcolor{forestgreen}{4.02} & \textcolor{forestgreen}{4.17}/\textcolor{red}{4.41}/\textcolor{red}{38.24} & \textcolor{forestgreen}{0.69}/\textcolor{forestgreen}{0.29}/\textcolor{forestgreen}{0.46} \\
 & Dropout & \textcolor{forestgreen}{86.50}/\textcolor{forestgreen}{93.44}/\textcolor{forestgreen}{91.00} & \textcolor{red}{3.00}/\textcolor{red}{1.38}/\textcolor{forestgreen}{7.00} & \textcolor{forestgreen}{1.26}/\textcolor{forestgreen}{1.10}/\textcolor{red}{5.60} & \textcolor{forestgreen}{10.24}/\textcolor{red}{6.10}/\textcolor{forestgreen}{13.16} & \textcolor{red}{1.91}/0.33/\textcolor{red}{1.27} \\
 & Attention entropy & \textcolor{red}{85.25}/\textcolor{forestgreen}{93.75}/\textcolor{forestgreen}{91.50} & \textcolor{forestgreen}{0.50}/\textcolor{red}{2.75}/\textcolor{forestgreen}{8.00} & \textcolor{forestgreen}{0.65}/\textcolor{red}{2.62}/\textcolor{red}{5.19} & \textcolor{forestgreen}{0.57}/\textcolor{red}{5.54}/\textcolor{red}{34.85} & \textcolor{red}{2.57}/\textcolor{red}{0.41}/\textcolor{forestgreen}{1.07} \\
 & Causal debias & \textcolor{red}{84.50}/\textcolor{forestgreen}{93.44}/\textcolor{forestgreen}{90.50} & \textcolor{forestgreen}{1.00}/\textcolor{red}{1.38}/\textcolor{forestgreen}{9.00} & \textcolor{red}{2.22}/\textcolor{forestgreen}{1.38}/\textcolor{forestgreen}{4.27} & \textcolor{forestgreen}{4.35}/\textcolor{red}{3.38}/\textcolor{red}{24.10} & \textcolor{red}{1.40}/\textcolor{red}{0.40}/\textcolor{forestgreen}{1.00} \\

 \midrule
 \multirow{7}{*}{\shortstack{RoBERTa}}  & Default & 84.50/93.00/90.33 & 1.00/3.75/10.33 & 2.87/3.44/1.82 & 6.54/8.31/47.47 & 2.55/0.30/0.89 \\
 & Group balance & \textcolor{forestgreen}{85.50}/\textcolor{red}{92.31}/\textcolor{red}{89.83} & \textcolor{red}{2.50}/\textcolor{forestgreen}{0.62}/\textcolor{red}{11.33} & \textcolor{forestgreen}{0.94}/\textcolor{forestgreen}{0.27}/\textcolor{forestgreen}{1.55} & \textcolor{red}{9.11}/\textcolor{forestgreen}{6.41}/\textcolor{forestgreen}{38.00} & \textcolor{forestgreen}{2.44}/\textcolor{forestgreen}{0.26}/\textcolor{forestgreen}{0.46} \\
 & Group-class balance & \textcolor{forestgreen}{85.00}/\textcolor{red}{92.50}/\textcolor{forestgreen}{90.67} & 1.00/\textcolor{forestgreen}{1.50}/\textcolor{forestgreen}{5.33} & \textcolor{forestgreen}{1.59}/\textcolor{forestgreen}{0.26}/\textcolor{red}{2.01} & \textcolor{red}{11.53}/\textcolor{red}{14.87}/\textcolor{forestgreen}{24.59} & \textcolor{forestgreen}{1.55}/\textcolor{red}{0.53}/\textcolor{forestgreen}{0.62} \\
 & CDA & \textcolor{forestgreen}{85.12}/\textcolor{forestgreen}{93.19}/\textcolor{red}{89.33} & \textcolor{forestgreen}{0.75}/\textcolor{forestgreen}{1.88}/\textcolor{forestgreen}{8.67} & \textcolor{red}{4.12}/\textcolor{forestgreen}{1.10}/\textcolor{red}{3.90} & \textcolor{red}{12.64}/\textcolor{red}{11.13}/\textcolor{forestgreen}{25.89} & \textcolor{forestgreen}{0.36}/\textcolor{forestgreen}{0.23}/\textcolor{forestgreen}{0.40} \\
 & Dropout & \textcolor{red}{83.88}/\textcolor{forestgreen}{93.69}/\textcolor{red}{90.17} & \textcolor{red}{1.75}/\textcolor{forestgreen}{0.88}/\textcolor{forestgreen}{7.67} & \textcolor{forestgreen}{1.29}/\textcolor{forestgreen}{0.82}/\textcolor{red}{2.97} & \textcolor{forestgreen}{3.10}/\textcolor{forestgreen}{3.28}/\textcolor{forestgreen}{26.86} & \textcolor{red}{2.71}/\textcolor{forestgreen}{0.23}/\textcolor{forestgreen}{0.87} \\
 & Attention entropy & \textcolor{forestgreen}{85.00}/\textcolor{forestgreen}{93.50}/90.33 & \textcolor{forestgreen}{0.50}/\textcolor{forestgreen}{1.75}/\textcolor{forestgreen}{6.67} & \textcolor{forestgreen}{2.23}/\textcolor{forestgreen}{2.06}/\textcolor{forestgreen}{1.01} & \textcolor{red}{6.55}/\textcolor{forestgreen}{0.62}/\textcolor{forestgreen}{22.78} & \textcolor{forestgreen}{2.39}/\textcolor{forestgreen}{0.24}/\textcolor{forestgreen}{0.81} \\
 & Causal debias & \textcolor{forestgreen}{86.25}/\textcolor{red}{92.19}/\textcolor{red}{89.50} & \textcolor{red}{2.00}/\textcolor{forestgreen}{3.37}/\textcolor{forestgreen}{10.00} & \textcolor{forestgreen}{2.23}/\textcolor{forestgreen}{2.33}/\textcolor{red}{1.84} & \textcolor{forestgreen}{0.60}/\textcolor{red}{14.77}/\textcolor{forestgreen}{43.47} & \textcolor{forestgreen}{2.09}/\textcolor{red}{0.39}/\textcolor{forestgreen}{0.66} \\

 \midrule
 \multirow{7}{*}{\shortstack{RoBERTa\\ (all)}}  & Default & 85.50/93.75/91.50 & 0.50/1.75/7.00 & 0.01/1.51/5.56 & 3.06/5.74/31.14 & 2.52/0.35/1.55 \\
 & Group balance & \textcolor{red}{85.38}/\textcolor{red}{93.62}/\textcolor{forestgreen}{91.67} & \textcolor{red}{1.75}/\textcolor{red}{3.25}/\textcolor{red}{9.33} & 0.01/\textcolor{red}{2.47}/\textcolor{forestgreen}{4.12} & \textcolor{red}{9.01}/\textcolor{red}{11.90}/\textcolor{red}{40.29} & \textcolor{red}{2.76}/\textcolor{forestgreen}{0.30}/\textcolor{forestgreen}{0.96} \\
 & Group-class balance & \textcolor{forestgreen}{86.38}/\textcolor{red}{92.56}/\textcolor{red}{90.17} & \textcolor{red}{2.25}/\textcolor{red}{1.88}/\textcolor{red}{10.67} & \textcolor{red}{0.62}/\textcolor{forestgreen}{1.37}/\textcolor{forestgreen}{2.58} & \textcolor{red}{9.05}/\textcolor{red}{8.62}/\textcolor{red}{64.35} & \textcolor{red}{4.75}/\textcolor{forestgreen}{0.23}/\textcolor{forestgreen}{0.34} \\
 & CDA & \textcolor{red}{85.25}/\textcolor{red}{92.56}/\textcolor{red}{90.67} & \textcolor{red}{1.00}/\textcolor{forestgreen}{0.62}/\textcolor{red}{7.67} & \textcolor{red}{1.59}/\textcolor{forestgreen}{0.13}/\textcolor{forestgreen}{1.80} & \textcolor{red}{11.53}/\textcolor{red}{7.49}/\textcolor{red}{31.28} & \textcolor{forestgreen}{0.52}/\textcolor{forestgreen}{0.23}/\textcolor{forestgreen}{0.74} \\
 & Dropout & \textcolor{forestgreen}{86.00}/\textcolor{red}{93.00}/\textcolor{red}{90.17} & \textcolor{red}{2.50}/1.75/\textcolor{forestgreen}{4.67} & \textcolor{red}{1.27}/1.51/\textcolor{forestgreen}{4.19} & \textcolor{red}{17.51}/\textcolor{red}{6.21}/\textcolor{forestgreen}{28.72} & \textcolor{forestgreen}{1.02}/\textcolor{forestgreen}{0.33}/\textcolor{forestgreen}{0.79} \\
 & Attention entropy & \textcolor{forestgreen}{86.75}/\textcolor{red}{93.50}/91.50 & 0.50/\textcolor{red}{2.50}/7.00 & \textcolor{red}{0.96}/\textcolor{red}{2.06}/\textcolor{forestgreen}{3.16} & \textcolor{red}{6.54}/\textcolor{red}{8.05}/\textcolor{forestgreen}{24.59} & \textcolor{red}{3.40}/\textcolor{red}{0.38}/\textcolor{forestgreen}{1.19} \\
 & Causal debias & \textcolor{red}{85.38}/\textcolor{red}{93.25}/\textcolor{red}{91.00} & \textcolor{forestgreen}{0.25}/\textcolor{red}{3.50}/\textcolor{red}{10.00} & 0.01/\textcolor{red}{2.62}/\textcolor{forestgreen}{5.41} & \textcolor{forestgreen}{1.88}/\textcolor{red}{13.69}/\textcolor{red}{34.14} & \textcolor{red}{2.55}/\textcolor{red}{0.40}/\textcolor{forestgreen}{0.80} \\ 
  \midrule

 \multirow{4}{*}{\shortstack{Llama3.2-3B-Instruct}} & Zero-shot & 54.00/70.50/65.17 & 8.50/1.00/25.67 & 10.91/1.53/31.87 & 0.20/4.56/8.33 & 2.39/3.00/4.28  \\
 & Few-shot & \textcolor{forestgreen}{73.12}/\textcolor{forestgreen}{88.62}/\textcolor{forestgreen}{86.83} & \textcolor{forestgreen}{7.25}/\textcolor{forestgreen}{0.50}/\textcolor{forestgreen}{7.67} & \textcolor{red}{13.01}/\textcolor{forestgreen}{0.16}/\textcolor{forestgreen}{4.83} & \textcolor{red}{15.05}/\textcolor{red}{9.08}/\textcolor{red}{23.17} & \textcolor{forestgreen}{1.63}/\textcolor{forestgreen}{1.68}/\textcolor{forestgreen}{2.00}  \\
 & Fairness imagination & \textcolor{forestgreen}{57.75}/\textcolor{forestgreen}{73.56}/\textcolor{forestgreen}{66.83} & \textcolor{forestgreen}{5.00}/\textcolor{red}{1.62}/\textcolor{red}{26.33} & \textcolor{forestgreen}{6.47}/\textcolor{red}{1.63}/\textcolor{forestgreen}{30.03} & \textcolor{red}{0.26}/\textcolor{forestgreen}{1.74}/\textcolor{red}{17.48} & \textcolor{red}{2.86}/\textcolor{red}{3.73}/\textcolor{forestgreen}{3.92}  \\
 & Fairness imagination & \textcolor{forestgreen}{57.75}/\textcolor{forestgreen}{73.56}/\textcolor{forestgreen}{66.83} & \textcolor{forestgreen}{5.00}/\textcolor{red}{1.62}/\textcolor{red}{26.33} & \textcolor{forestgreen}{6.47}/\textcolor{red}{1.63}/\textcolor{forestgreen}{30.03} & \textcolor{red}{0.26}/\textcolor{forestgreen}{1.74}/\textcolor{red}{17.48} & \textcolor{red}{2.86}/\textcolor{red}{3.73}/\textcolor{forestgreen}{3.92}  \\
 & Fairness instruction & \textcolor{forestgreen}{77.00}/\textcolor{forestgreen}{89.00}/\textcolor{forestgreen}{87.17} & \textcolor{forestgreen}{2.00}/\textcolor{forestgreen}{0.75}/\textcolor{forestgreen}{10.67} & \textcolor{forestgreen}{2.87}/\textcolor{forestgreen}{0.84}/\textcolor{forestgreen}{3.97} & \textcolor{red}{2.19}/\textcolor{forestgreen}{3.33}/\textcolor{red}{36.36} & \textcolor{forestgreen}{1.39}/\textcolor{forestgreen}{0.97}/\textcolor{forestgreen}{1.87}  \\
 \midrule

 \multirow{4}{*}{\shortstack{Qwen3-4B}} & Zero-shot & 66.75/77.25/77.33 & 3.50/3.75/16.33 & 4.21/3.78/17.40 & 0.80/4.05/5.89 & 3.05/2.31/3.67  \\
 & Few-shot & \textcolor{red}{57.88}/\textcolor{red}{68.06}/\textcolor{forestgreen}{77.83} & \textcolor{red}{8.75}/\textcolor{forestgreen}{2.12}/\textcolor{forestgreen}{9.33} & \textcolor{red}{11.52}/\textcolor{forestgreen}{1.80}/\textcolor{forestgreen}{10.86} & \textcolor{red}{1.49}/\textcolor{red}{5.79}/\textcolor{red}{9.45} & \textcolor{red}{3.60}/\textcolor{red}{4.31}/\textcolor{red}{4.18}  \\
 & Fairness imagination & \textcolor{forestgreen}{73.75}/\textcolor{forestgreen}{82.88}/\textcolor{forestgreen}{86.33} & \textcolor{forestgreen}{3.00}/\textcolor{forestgreen}{1.00}/\textcolor{forestgreen}{10.33} & \textcolor{red}{5.12}/\textcolor{forestgreen}{0.89}/\textcolor{forestgreen}{5.79} & \textcolor{red}{5.39}/\textcolor{forestgreen}{0.82}/\textcolor{red}{24.13} & \textcolor{red}{3.14}/\textcolor{red}{2.97}/\textcolor{forestgreen}{2.51}  \\
 & Fairness instruction & \textcolor{forestgreen}{78.00}/\textcolor{forestgreen}{89.50}/\textcolor{forestgreen}{89.33} & \textcolor{forestgreen}{3.00}/\textcolor{forestgreen}{0.50}/\textcolor{forestgreen}{9.33} & \textcolor{forestgreen}{4.14}/\textcolor{forestgreen}{0.26}/\textcolor{forestgreen}{3.13} & \textcolor{red}{2.04}/\textcolor{red}{5.23}/\textcolor{red}{26.74} & \textcolor{forestgreen}{1.95}/\textcolor{forestgreen}{1.43}/\textcolor{forestgreen}{1.61}  \\

 \midrule 
 \multirow{4}{*}{\shortstack{Qwen3-8B}}& Zero-shot & 48.12/59.50/56.50 & 7.25/0.00/13.00 & 9.68/0.37/18.59 & 1.31/4.92/6.30 & 3.31/3.47/5.52  \\
 & Few-shot & \textcolor{forestgreen}{53.75}/\textcolor{forestgreen}{67.19}/\textcolor{forestgreen}{77.17} & \textcolor{forestgreen}{5.50}/\textcolor{red}{1.12}/\textcolor{forestgreen}{8.67} & \textcolor{forestgreen}{6.18}/\textcolor{red}{1.53}/\textcolor{forestgreen}{10.51} & \textcolor{red}{3.41}/\textcolor{forestgreen}{1.95}/\textcolor{forestgreen}{4.88} & \textcolor{red}{4.50}/\textcolor{red}{5.04}/\textcolor{red}{5.99}  \\
 & Fairness imagination & \textcolor{forestgreen}{51.62}/\textcolor{forestgreen}{67.50}/\textcolor{forestgreen}{61.83} & \textcolor{forestgreen}{4.25}/\textcolor{red}{0.75}/\textcolor{forestgreen}{9.67} & \textcolor{forestgreen}{5.24}/\textcolor{red}{0.56}/\textcolor{forestgreen}{11.23} & \textcolor{forestgreen}{1.05}/\textcolor{forestgreen}{3.49}/\textcolor{red}{6.45} & \textcolor{forestgreen}{2.51}/\textcolor{forestgreen}{2.02}/\textcolor{forestgreen}{2.55}  \\
 & Fairness instruction & \textcolor{forestgreen}{60.25}/\textcolor{forestgreen}{71.19}/\textcolor{forestgreen}{67.50} & \textcolor{red}{8.50}/\textcolor{red}{1.87}/\textcolor{forestgreen}{12.00} & \textcolor{red}{10.57}/\textcolor{red}{2.23}/\textcolor{forestgreen}{14.22} & \textcolor{forestgreen}{0.93}/\textcolor{forestgreen}{1.90}/\textcolor{forestgreen}{2.10} & \textcolor{forestgreen}{2.46}/\textcolor{forestgreen}{3.13}/\textcolor{forestgreen}{3.60}  \\

    \bottomrule
     \end{tabular}
     }

     \label{tab:appendix jigsaw performance fairness}
\end{table*}

Tables~\ref{tab:appendix civil performance fairness} and~\ref{tab:appendix jigsaw performance fairness} show the task performance and fairness scores for the default/zero-shot and debiased models on the Civil Comments and Jigsaw datasets respectively. 
To better identify the differences between different debiasing methods, we conduct an analysis based on how often a debiasing method successfully reduces the average individual unfairness ($\text{Avg}_\text{iu}$) and maintains the task performance (Accuracy) of the default/zero-shot model.

\paragraph{Encoder-only models}
Analyzing the results with respect to the dataset, we find that the models are able to better preserve their original accuracy on the Civil Comments dataset (48.61\% of the cases) compared to the Jigsaw dataset (40.28\% of the cases). 
In contrast, mitigating bias seems substantially easier on the Jigsaw dataset (in 63.88\% of the cases) than on the Civil Comments (only 50\% of the cases). 
On closer inspection, we find that this skew comes from religion bias in the Jigsaw dataset which is improved in 95.83\% of the cases after debiasing, followed by race bias (50\%) and gender bias (45.83\%).
In the Civil Comments dataset, we find that gender bias is mitigated best (improvement in 62.5\% of the cases), followed by religion bias (54.17\%) and race bias (33.33\%).

With respect to the debiasing method, we find that CDA performs best in terms of debiasing, as it reduces $\text{Avg}_\text{iu}$ across all bias types, datasets, and models. 
The second best performing method is group-class balance which manages to reduce $\text{Avg}_\text{iu}$ in 58.33\% of the cases on the Civil Comments dataset and in 75\% cases on the Jigsaw dataset.
For the other methods, the results are mixed as we again observe dataset-specific differences.
For example, we find that Attention entropy performs well on the Jigsaw dataset (50\%) but performs worst on the Civil Comments dataset (16.67\%).
These differences become even more pronounced when looking at different bias types.
For instance, causal debiasing improves $\text{Avg}_\text{iu}$ for religion bias across all models on the Jigsaw dataset but at the same time, does not improve a single model in terms of $\text{Avg}_\text{iu}$ for gender bias in the same dataset.
Interestingly, we find an inverse trend on the Civil Comments dataset; i.e., causal debiasing succeeds on all models for gender bias, but only for one model for religion bias.
These findings highlight the importance of considering a diverse set of datasets for evaluating debiasing methods, as results on a single dataset can be misleading.

\paragraph{Decoder-only models}
We find that the debiasing methods (fairness imagination and fairness instruction) for the decoder-only models consistently improve the task performance across all bias types and datasets.
Contrary to this, we see increases in average individual unfairness of the fairness imagination approach for race and gender bias on Llama3.2-3B-Instruct and Qwen3-4B across both datasets.
Only for religion, fairness imagination leads to a consistent decrease of the individual unfairness across models. 
For fairness instruction, we observe a consistent improvement across all three bias types and both datasets, showing the clear superiority of the approach.
The consistency of the results is especially surprising when considering that both decoder-only models are instruction-tuned and aligned with human values, and that \citet{chen-etal-2025-causally} identify a bias amplification effect from instruction tuning. 
We conclude that fairness instruction is a good baseline to evaluate other debiasing methods for decoder-only models.

\section{Bias Detection Results}
\label{appendix: rq1}

\begin{figure*}[h!]  
    \centering
    \includegraphics[width=\linewidth]{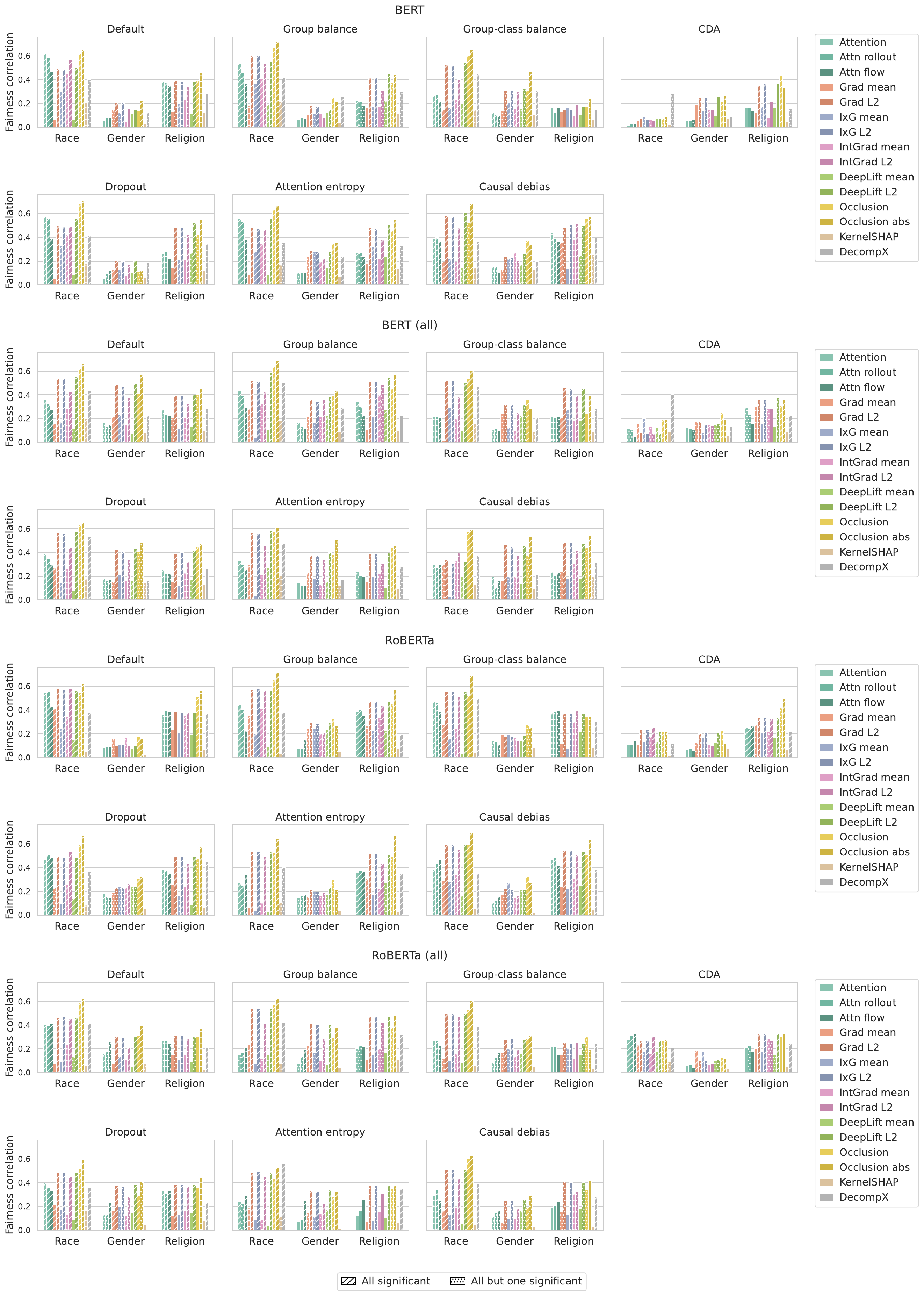}
    \caption{Fairness correlation results on Civil Comments for each explanation method across encoder-only models and bias types. Higher values indicate that the method is more effective and reliable in detecting biased predictions at inference time. \textit{All} indicates the model is trained on data containing all bias types.}
    \label{fig:appendix fairness correlation civil encoder}
    
\end{figure*}

\begin{figure*}[h!]  
    \centering
    \includegraphics[width=\linewidth]{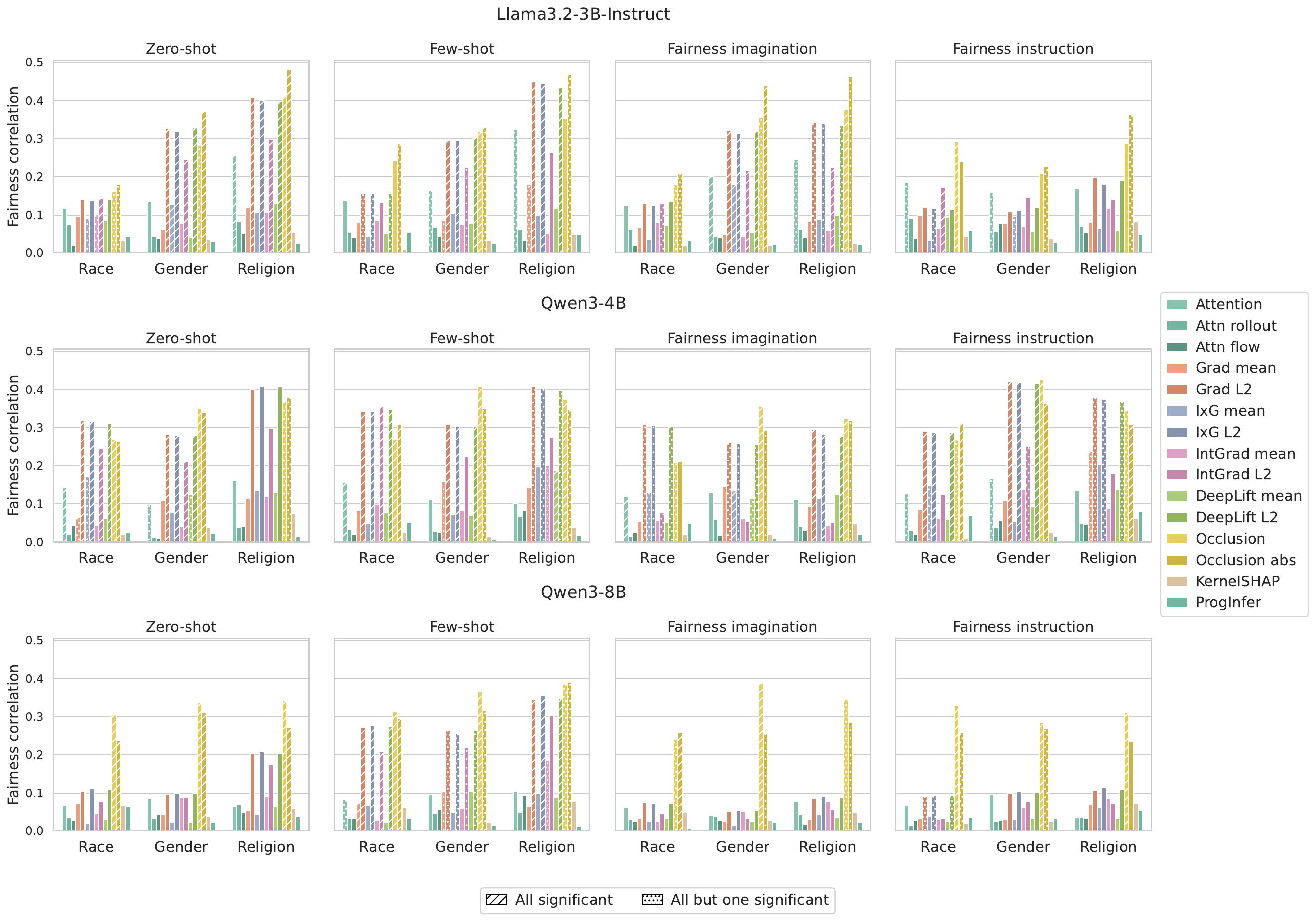}
    \caption{Fairness correlation results on Civil Comments for each explanation method across decoder-only models and bias types. Higher values indicate that the method is more effective and reliable in detecting biased predictions at inference time.}
    \label{fig:appendix fairness correlation civil decoder}
    
\end{figure*}

\begin{figure*}[h!]  
    \centering
    \includegraphics[width=\linewidth]{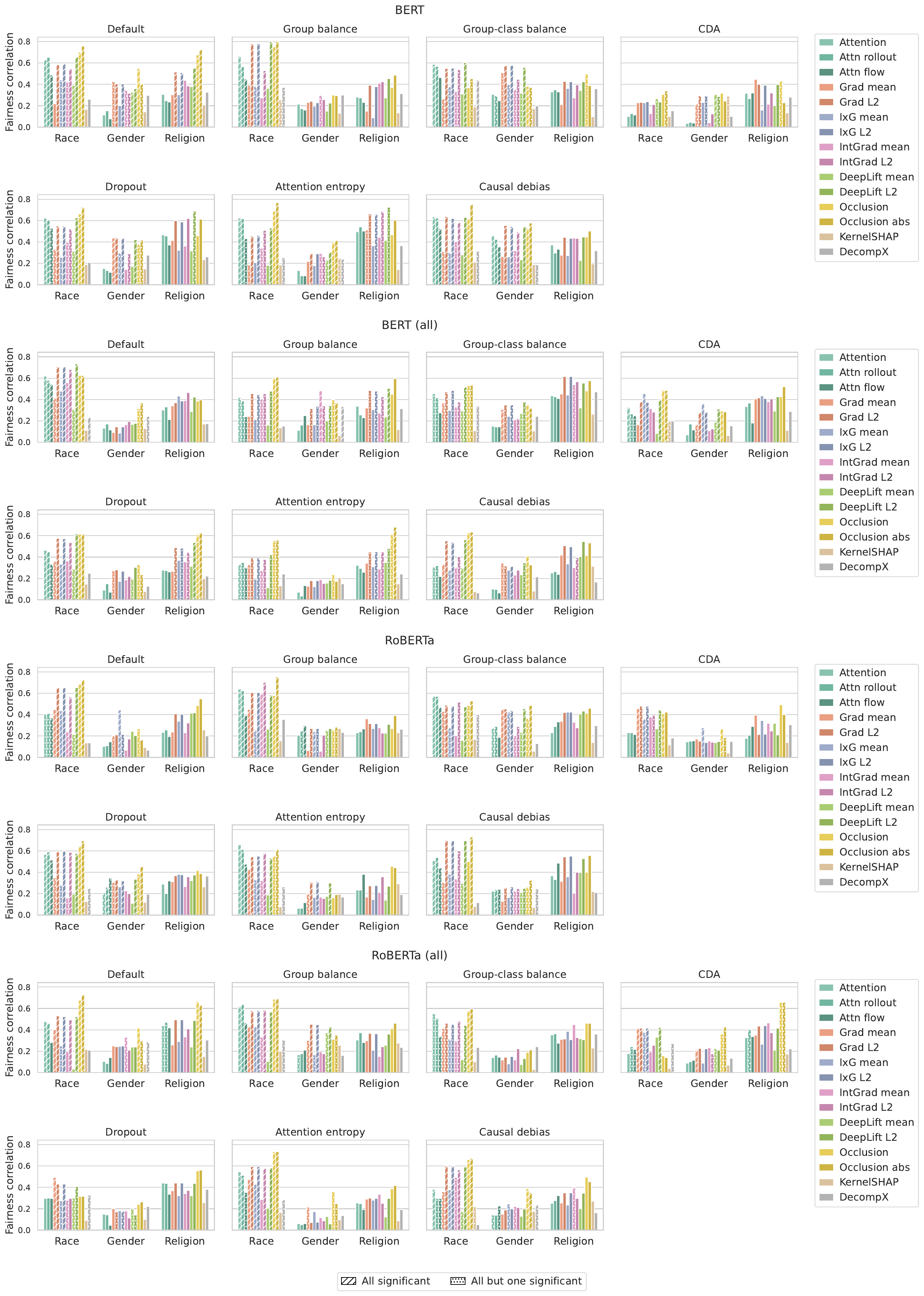}
    \caption{Fairness correlation results on Jigsaw for each explanation method across encoder-only models and bias types. Higher values indicate that the method is more effective and reliable in detecting biased predictions at inference time.
    \textit{All} indicates the model is trained on data containing all bias types.}
    \label{fig:appendix fairness correlation jigsaw encoder}
    
\end{figure*}

\begin{figure*}[h!]  
    \centering
    \includegraphics[width=\linewidth]{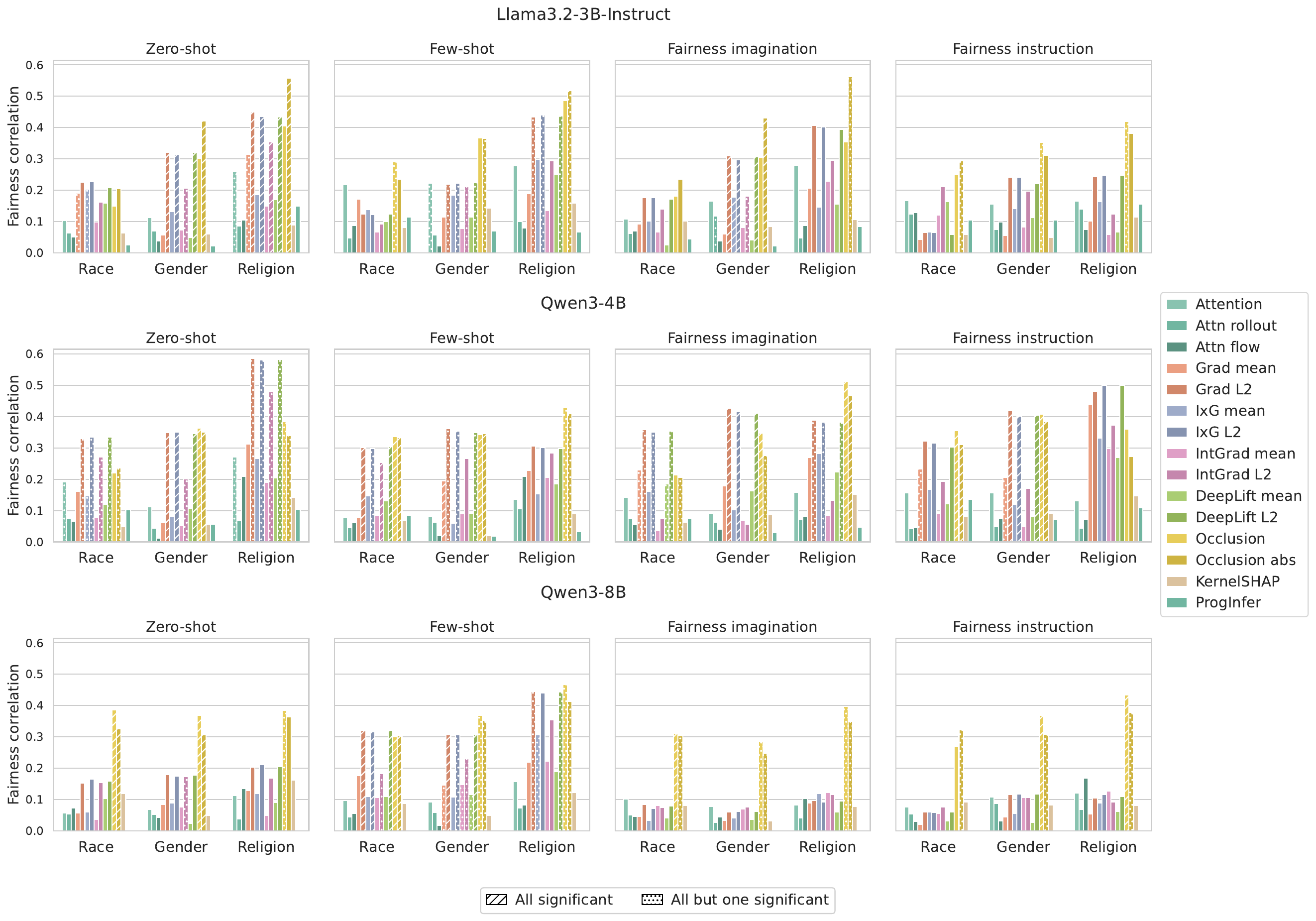}
    \caption{Fairness correlation results on Jigsaw for each explanation method across decoder-only models and bias types. Higher values indicate that the method is more effective and reliable in detecting biased predictions at inference time.}
    \label{fig:appendix fairness correlation jigsaw decoder}
    
\end{figure*}

\paragraph{Fairness correlation}
We present the full fairness correlation results of encoder- and decoder-only models with different debiasing methods on Civil Comments and Jigsaw in Figures\ref{fig:appendix fairness correlation civil encoder}, \ref{fig:appendix fairness correlation civil decoder}, \ref{fig:appendix fairness correlation jigsaw encoder}, \ref{fig:appendix fairness correlation jigsaw decoder}. Consistent with findings presented in the main text, Occlusion- and L2-based explanation methods achieve strong fairness correlations across different setups.

Comparing different debiasing methods, we find that low correlation scores primarily occur when individual unfairness is less pronounced, such as in CDA models.
In these cases, the models themselves produce fewer biased predictions, making the detection of bias through explanations less critical.
The lower correlations therefore do not substantially undermine the role of explanations in bias identification.

\section{Model Selection Results}
\label{appendix: rq2}

\paragraph{Explanation-Based Metrics and Fair Model Selection Results}
We evaluate several explanation-based metrics for selecting fair models with respect to different fairness criteria:

\begin{itemize}
    \item \textbf{Average absolute sensitive token reliance:} used to predict average individual unfairness, under the assumption that higher reliance on sensitive tokens implies greater sensitivity to group substitutions.
    \item \textbf{Group differences in average absolute sensitive token reliance:} used to predict disparities in accuracy, assuming that stronger reliance on sensitive features increases the risk of incorrect predictions.
    \item \textbf{Group differences in average absolute sensitive token reliance for positive/negative predictions:} used to predict disparities in false positive and false negative rates, respectively.
\end{itemize}

\begin{figure*}[h!]  
    \centering
    \includegraphics[width=\linewidth]{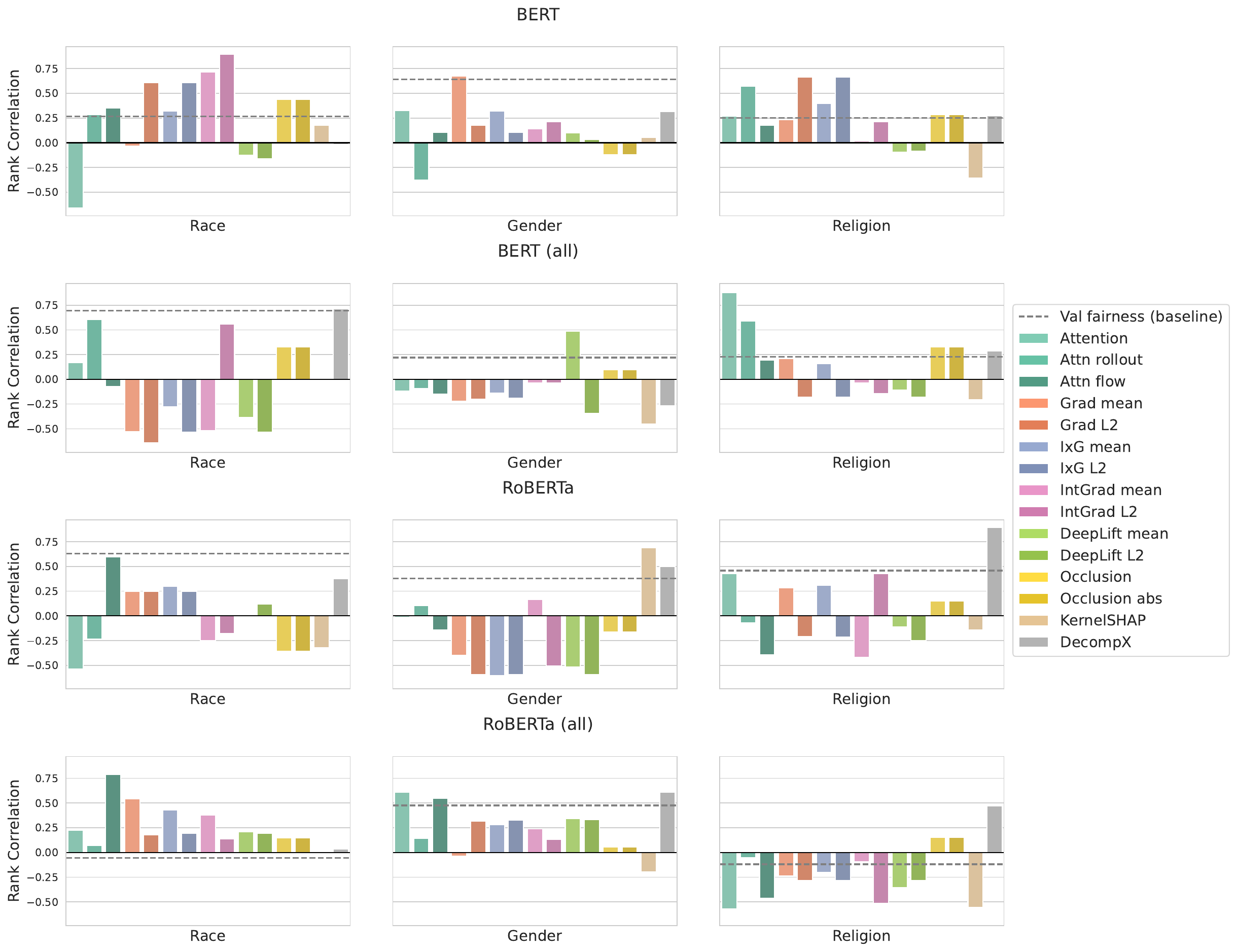}
    \caption{Rank correlations between validation set average absolute sensitive token reliance and individual unfairness on the test set for encoder-only models on Civil Comments.  
    The validation set sizes are 500 for race, 500 for gender, and 200 for religion. 
    Higher correlation values indicate greater effectiveness in ranking models.
    \textit{All} indicates the model is trained on all bias types.}
    \label{fig:appendix model selection civil 500 bert individual fairness rank}
    
\end{figure*}

\begin{figure*}[h!]  
    \centering
    \includegraphics[width=\linewidth]{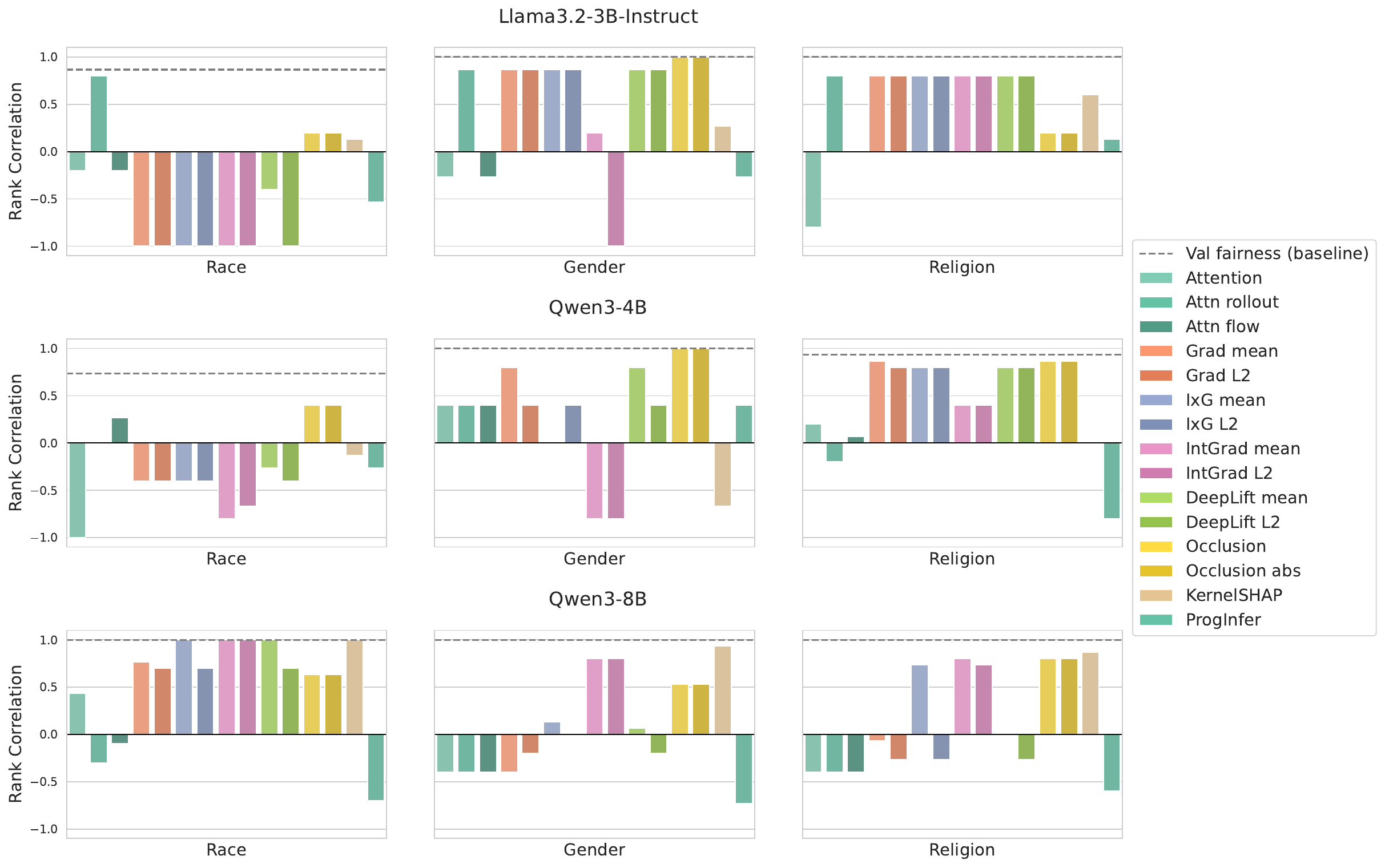}
    \caption{Rank correlations between validation set average absolute sensitive token reliance and individual unfairness on the test set for decoder-only models on Civil Comments.  
    The validation set sizes are 500 for race, 500 for gender, and 200 for religion. 
    Higher correlation values indicate greater effectiveness in ranking models.}
    \label{fig:appendix model selection civil 500 roberta individual fairness rank}
    
\end{figure*}

\begin{figure*}[h!]  
    \centering
    \includegraphics[width=\linewidth]{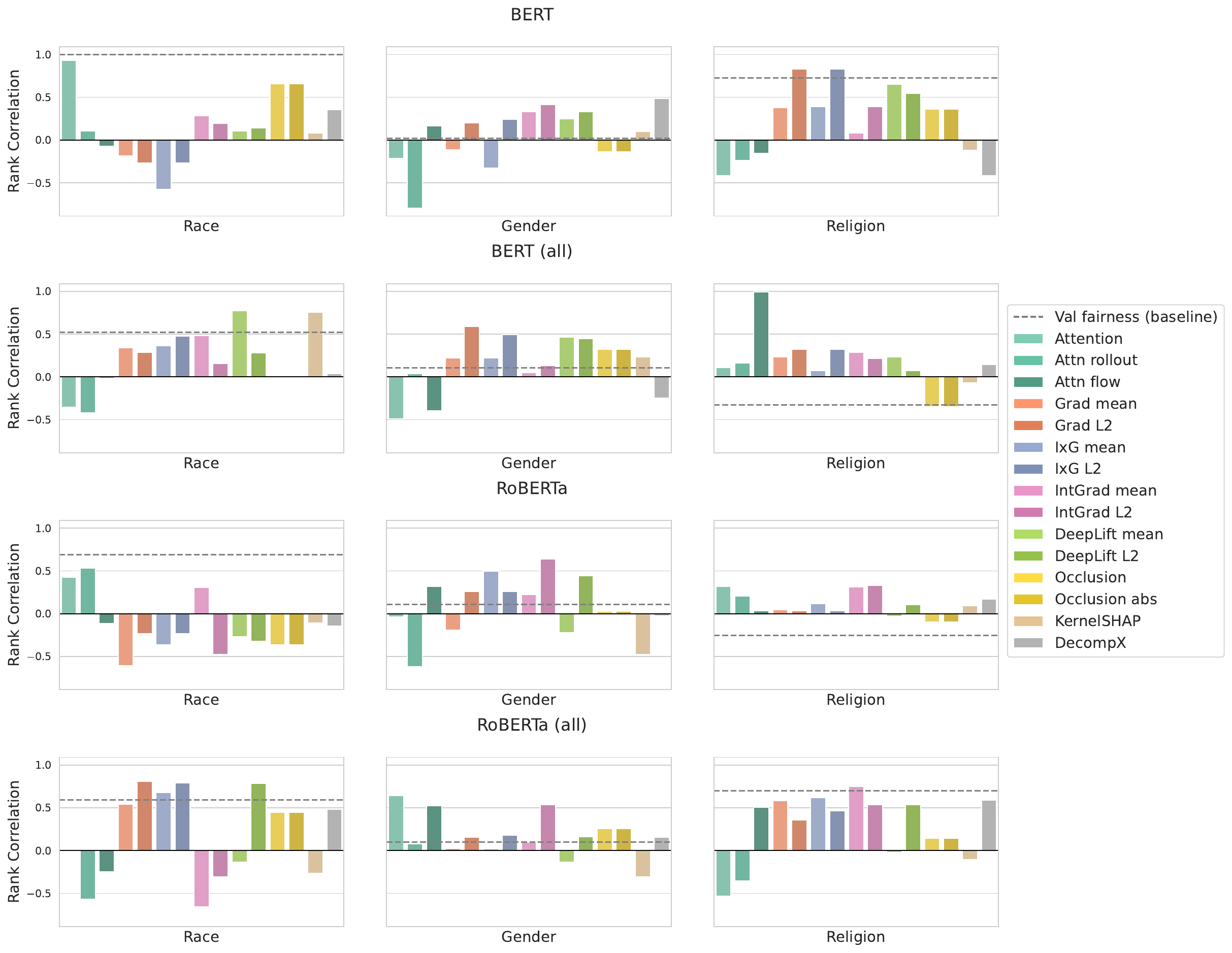}
    \caption{Rank correlations between validation set average absolute sensitive token reliance and individual unfairness on the test set for encoder-only models on Jigsaw.  
    The validation set size is 200. 
    Higher correlation values indicate greater effectiveness in ranking models.
    \textit{All} indicates the model is trained on all bias types.}
    \label{fig:appendix model selection jigsaw 500 bert individual fairness rank}
    
\end{figure*}

\begin{figure*}[h!]  
    \centering
    \includegraphics[width=\linewidth]{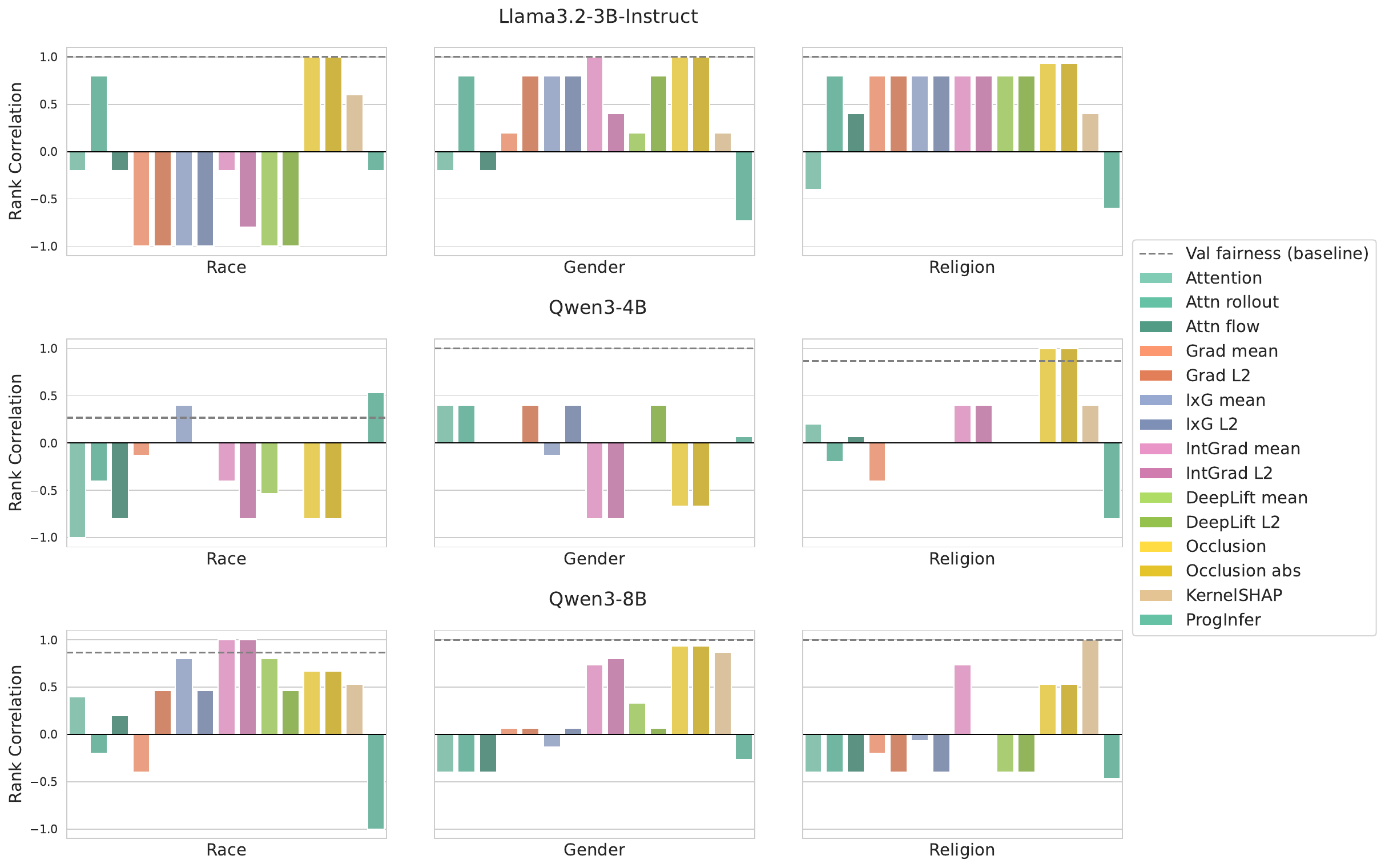}
    \caption{Rank correlations between validation set average absolute sensitive token reliance and individual unfairness on the test set for decoder-only models on Jigsaw.  
    The validation set size is 200. 
    Higher correlation values indicate greater effectiveness in ranking models.}
    \label{fig:appendix model selection jigsaw 500 roberta individual fairness rank}
    
\end{figure*}

Among these, only average absolute sensitive token reliance exhibits rank correlations above random chance with its target fairness metric (individual unfairness).
The correlations for other metrics remain at chance level.
Figures~\ref{fig:appendix model selection civil 500 bert individual fairness rank}, \ref{fig:appendix model selection civil 500 roberta individual fairness rank}, \ref{fig:appendix model selection jigsaw 500 bert individual fairness rank}, \ref{fig:appendix model selection jigsaw 500 roberta individual fairness rank} demonstrate that no explanation methods can consistently match baseline rank correlation results.

\begin{figure*}[h!]  
    \centering
    \includegraphics[width=\linewidth]{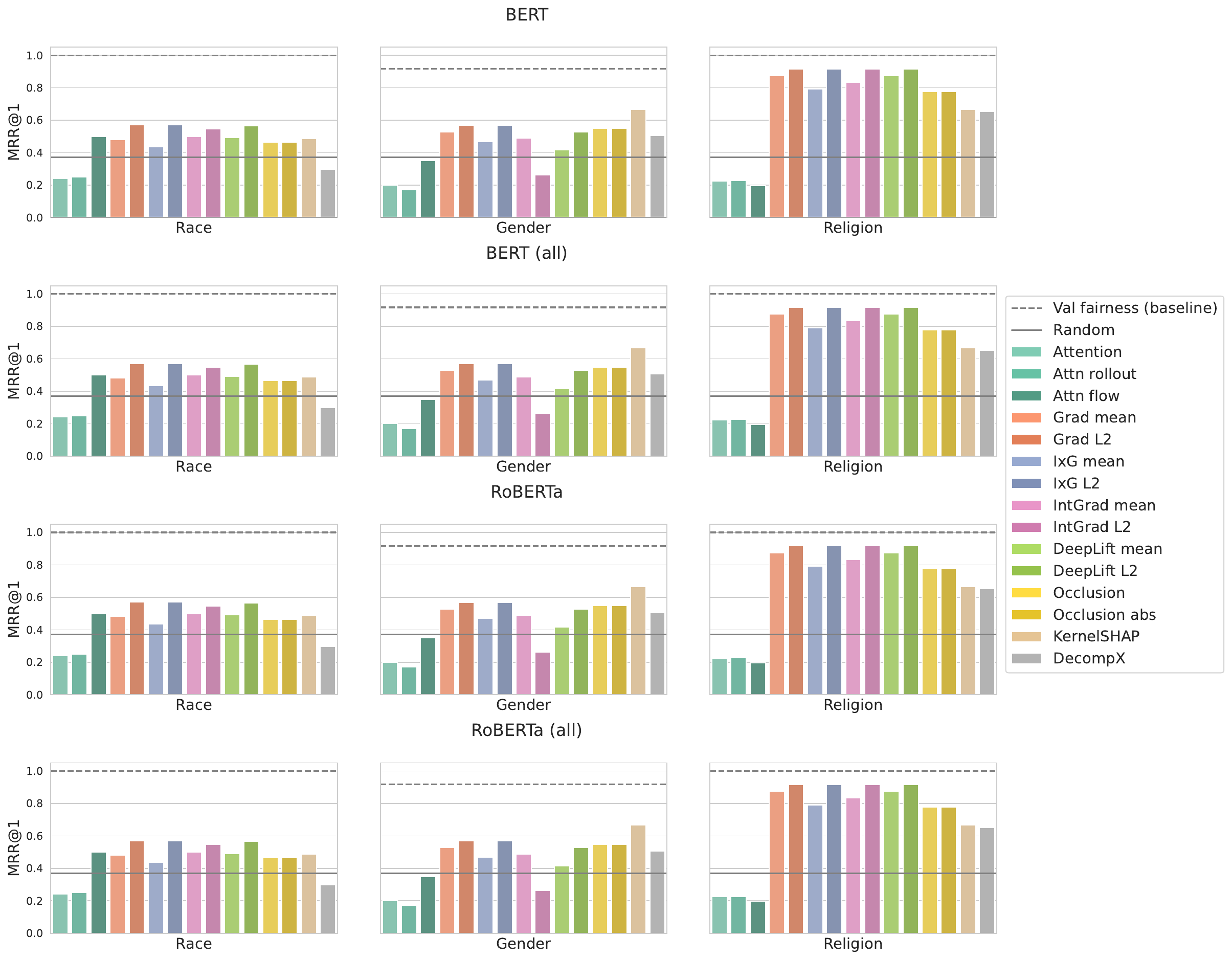}
    \caption{MRR@1 results for encoder-only models on Civil Comments.  
    The validation set sizes are 500 for race, 500 for gender, and 200 for religion.
    Higher MRR@1 scores indicate explanations are more effective in selecting the fairest models.
    \textit{All} indicates the model is trained on all bias types.}
    \label{fig:appendix mrr civil bert}
    
\end{figure*}

\begin{figure*}[h!]  
    \centering
    \includegraphics[width=\linewidth]{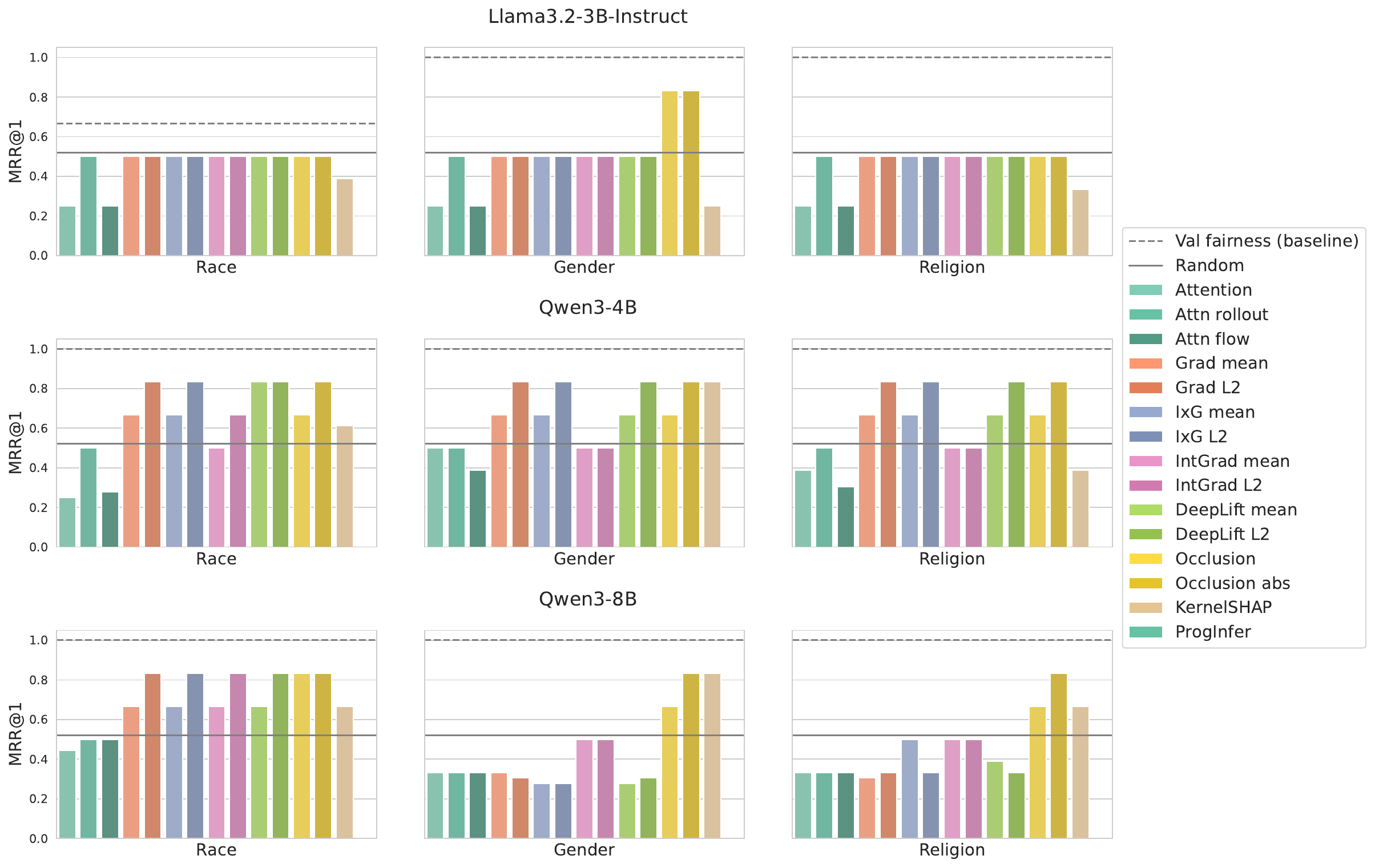}
    \caption{MRR@1 results for decoder-only models on Civil Comments.  
    The validation set sizes are 500 for race, 500 for gender, and 200 for religion.
    Higher MRR@1 scores indicate explanations are more effective in selecting the fairest models.}
    \label{fig:appendix mrr civil roberta}
    
\end{figure*}

\begin{figure*}[h!]  
    \centering
    \includegraphics[width=\linewidth]{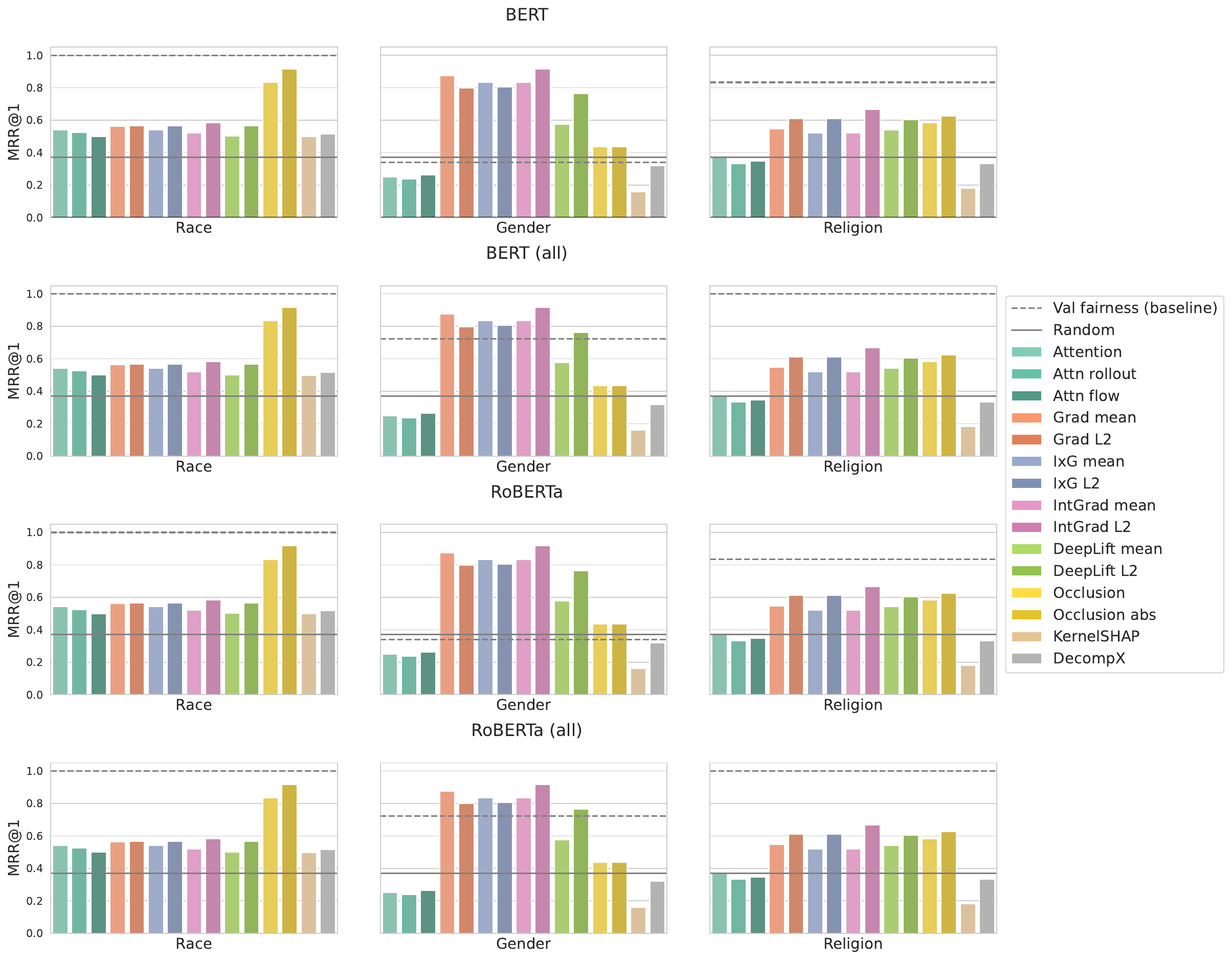}
    \caption{MRR@1 results for encoder-only models on Jigsaw.  
    The validation set size is 200.
    Higher MRR@1 scores indicate explanations are more effective in selecting the fairest models.
    \textit{All} indicates the model is trained on all bias types.}
    \label{fig:appendix mrr jigsaw bert}
    
\end{figure*}

\begin{figure*}[h!]  
    \centering
    \includegraphics[width=\linewidth]{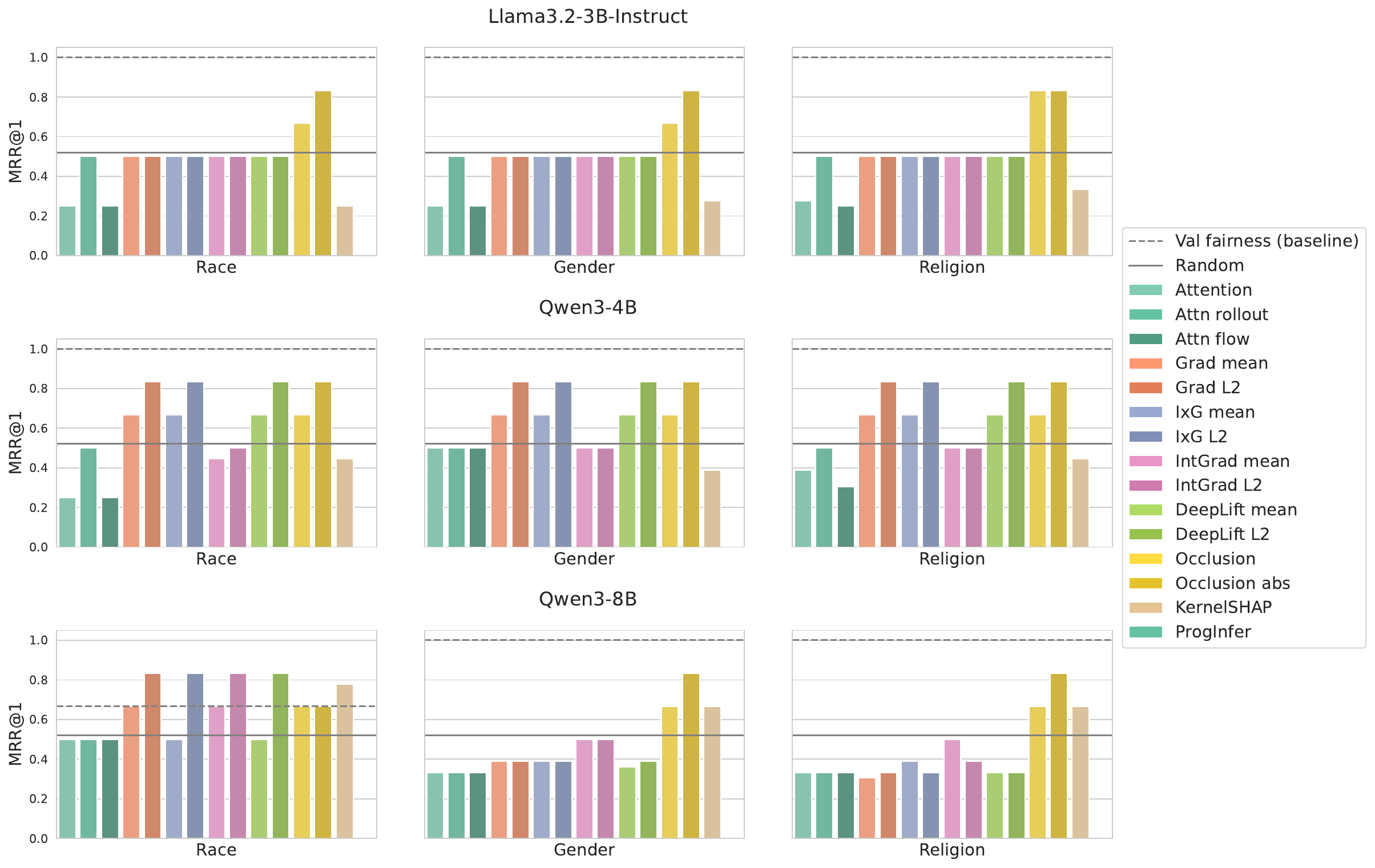}
    \caption{MRR@1 results for decoder-only models on Jigsaw.  
    The validation set size is 200.
    Higher MRR@1 scores indicate explanations are more effective in selecting the fairest models.}
    \label{fig:appendix mrr jigsaw roberta}
    
\end{figure*}

Figures~\ref{fig:appendix mrr civil bert}, \ref{fig:appendix mrr civil roberta}, \ref{fig:appendix mrr jigsaw bert}, \ref{fig:appendix mrr jigsaw roberta} further reveal that explanation methods are not able to robustly select the fairest models.
These findings underline the unreliability of explanation-based model selection.

\section{Bias Mitigation Results}
\label{appendix: rq3}

\begin{figure*}[]
\centering
\includegraphics[width=\linewidth]{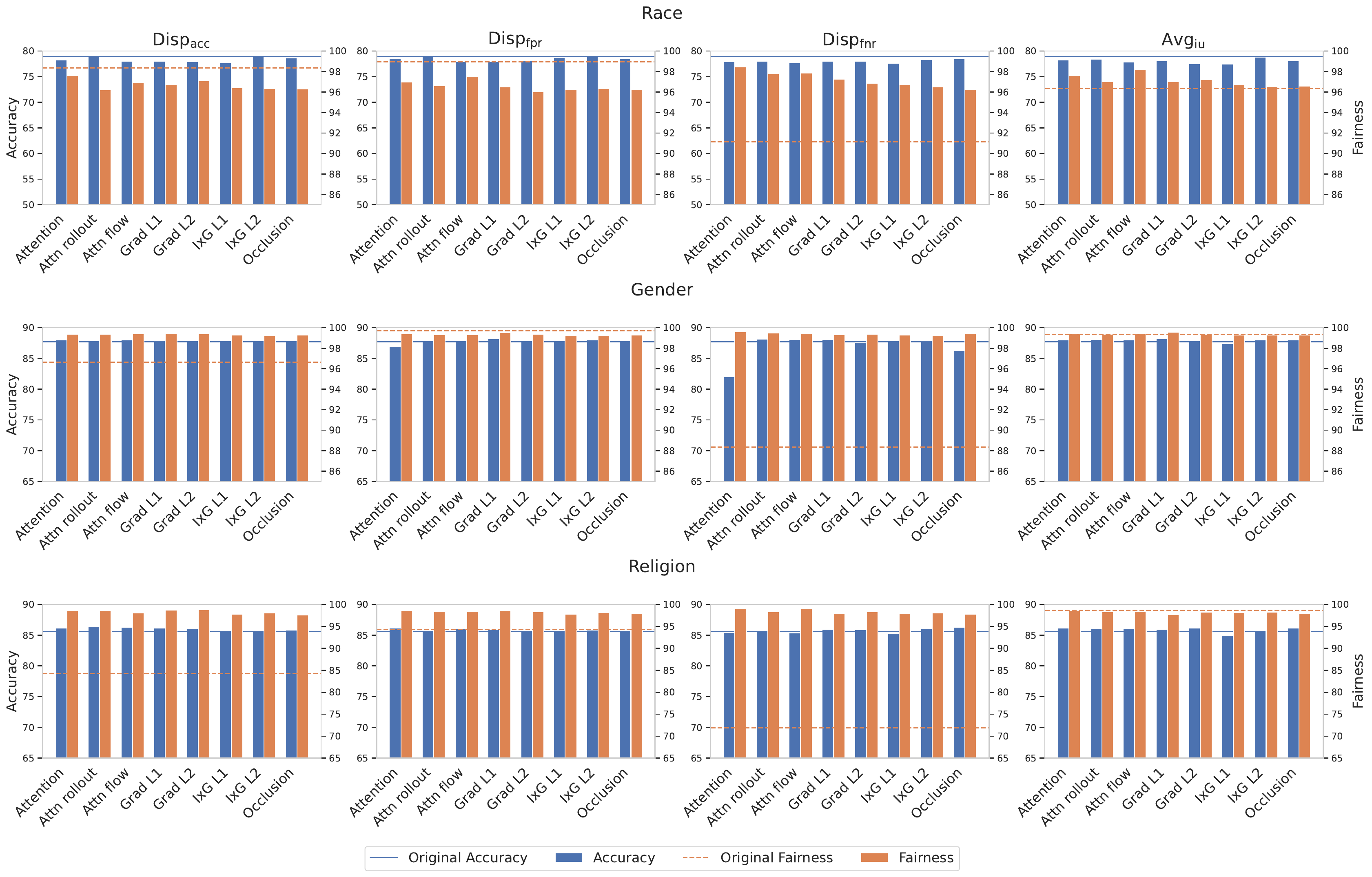}
\caption{Accuracy and fairness results for bias mitigation in BERT on the Civil Comments dataset, using different explanation methods during training. 
For consistency with accuracy, fairness results are reported as $100-\{\text{Disp}_\text{acc}, \text{Disp}_\text{fpr}, \text{Disp}_\text{fnr}, \text{Avg}_\text{iu}\}$, so that higher values indicate better debiasing performance.
Each column corresponds to models selected by maximizing the fairness-balanced metric with respect to the indicated bias metric.}
\label{fig:appendix bias mitigation civil bert}
\end{figure*}

\begin{figure*}[]
\centering
\includegraphics[width=\linewidth]{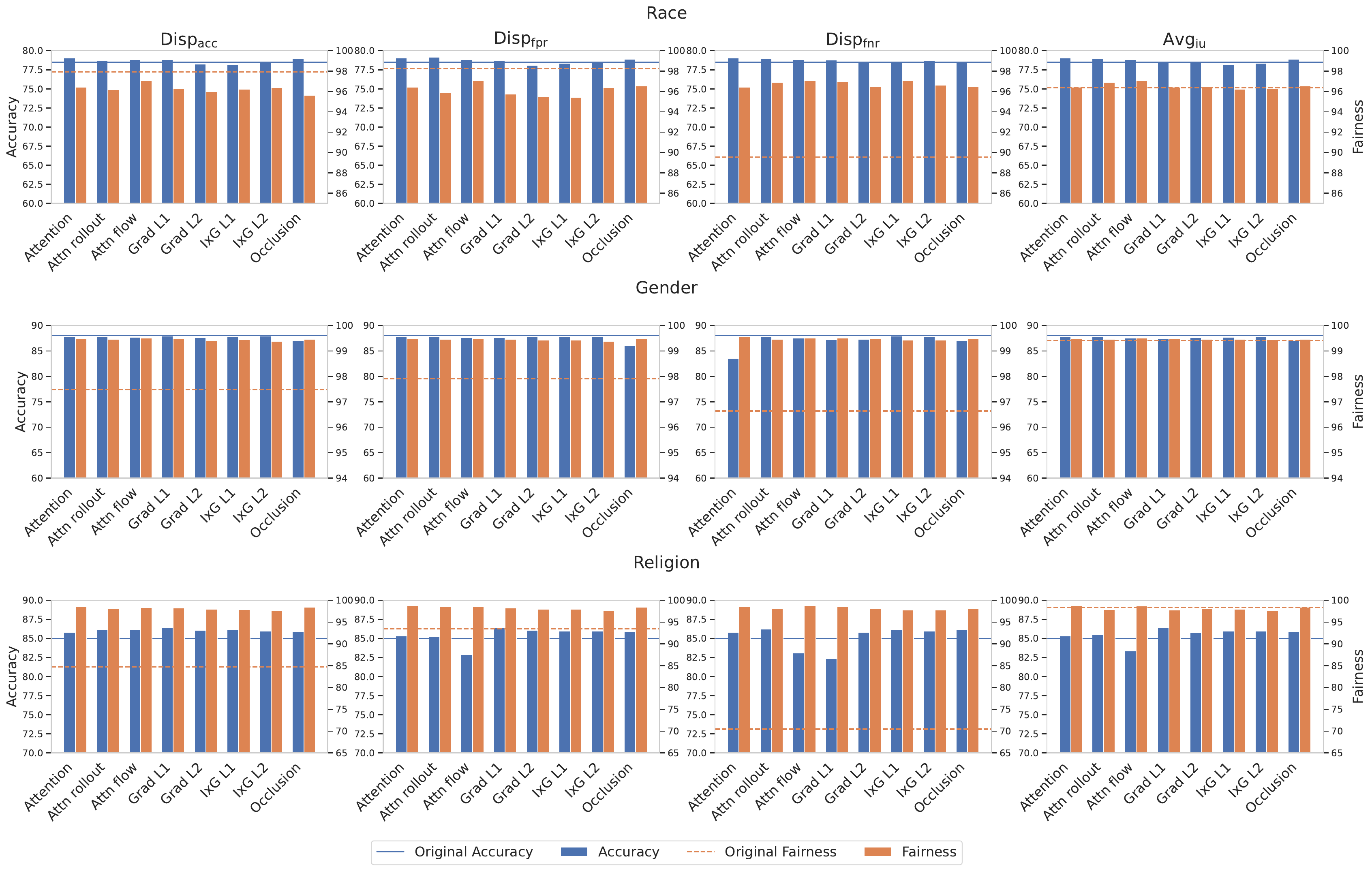}
\caption{Accuracy and fairness results for bias mitigation in RoBERTa on the Civil Comments dataset, using different explanation methods during training. 
For consistency with accuracy, fairness results are reported as $100-\{\text{Disp}_\text{acc}, \text{Disp}_\text{fpr}, \text{Disp}_\text{fnr}, \text{Avg}_\text{iu}\}$, so that higher values indicate better debiasing performance.
Each column corresponds to models selected by maximizing the fairness-balanced metric with respect to the indicated bias metric.}
\label{fig:appendix bias mitigation civil roberta}
\end{figure*}

\begin{figure*}[]
\centering
\includegraphics[width=\linewidth]{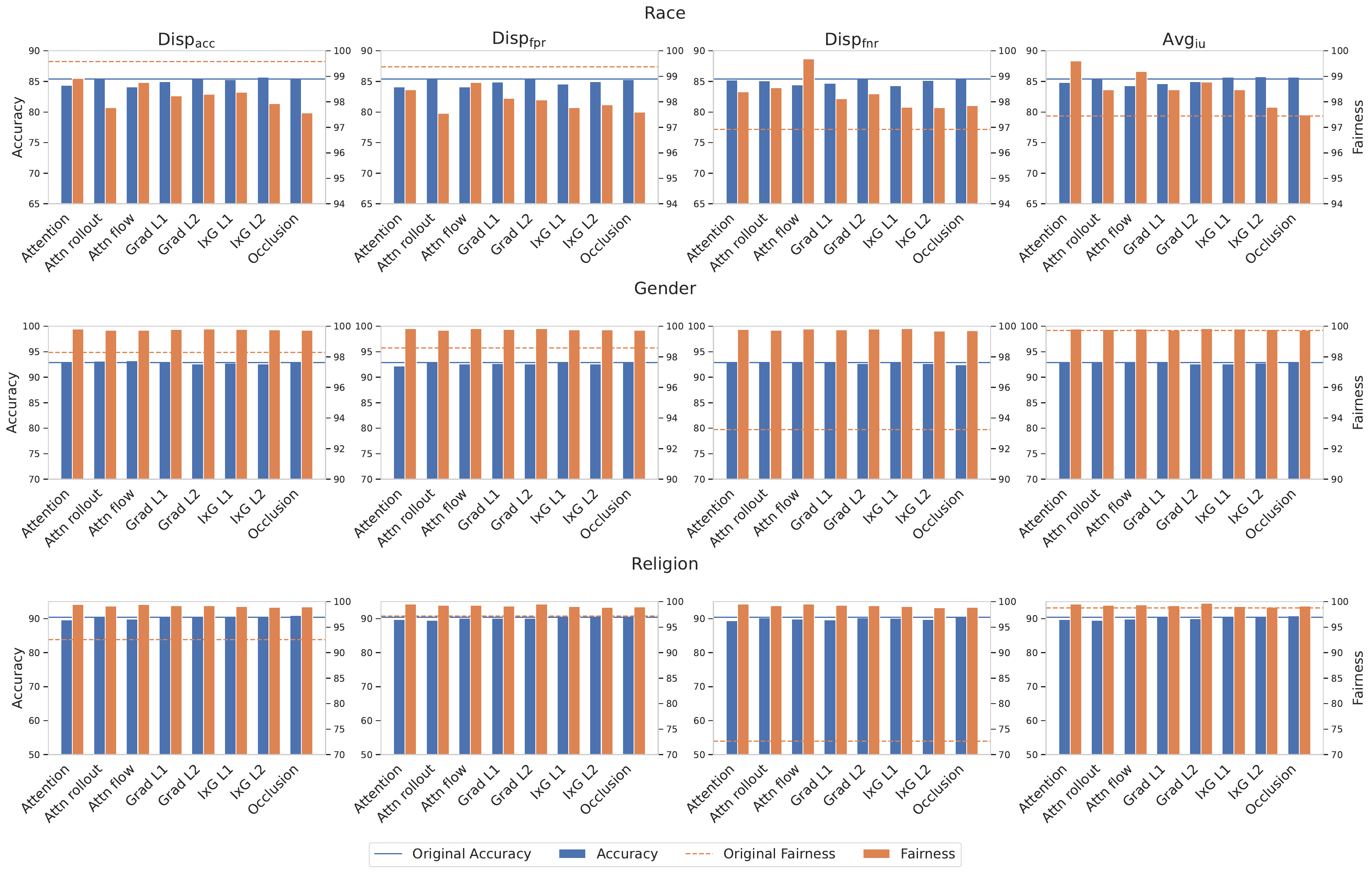}
\caption{Accuracy and fairness results for bias mitigation in BERT on the Jigsaw, using different explanation methods during training. 
For consistency with accuracy, fairness results are reported as $100-\{\text{Disp}_\text{acc}, \text{Disp}_\text{fpr}, \text{Disp}_\text{fnr}, \text{Avg}_\text{iu}\}$, so that higher values indicate better debiasing performance.
Each column corresponds to models selected by maximizing the fairness-balanced metric with respect to the indicated bias metric.}
\label{fig:appendix bias mitigation jigsaw bert}
\end{figure*}

\begin{figure*}[]
\centering
\includegraphics[width=\linewidth]{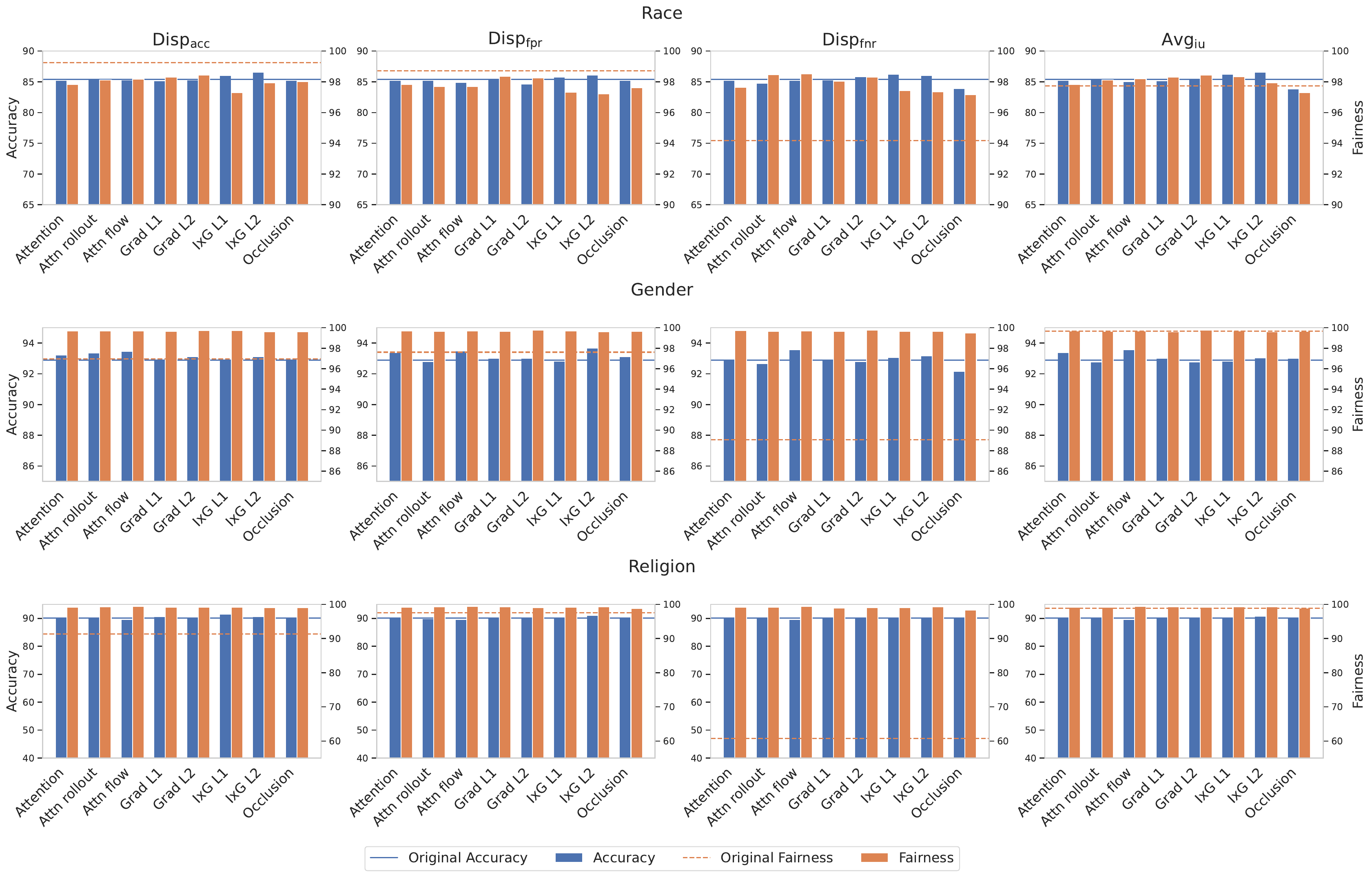}
\caption{Accuracy and fairness results for bias mitigation in RoBERTa on the Jigsaw dataset, using different explanation methods during training. 
For consistency with accuracy, fairness results are reported as $100-\{\text{Disp}_\text{acc}, \text{Disp}_\text{fpr}, \text{Disp}_\text{fnr}, \text{Avg}_\text{iu}\}$, so that higher values indicate better debiasing performance.
Each column corresponds to models selected by maximizing the fairness-balanced metric with respect to the indicated bias metric.}
\label{fig:appendix bias mitigation jigsaw roberta}
\end{figure*}

\begin{figure}
    \centering
    \includegraphics[width=\linewidth]{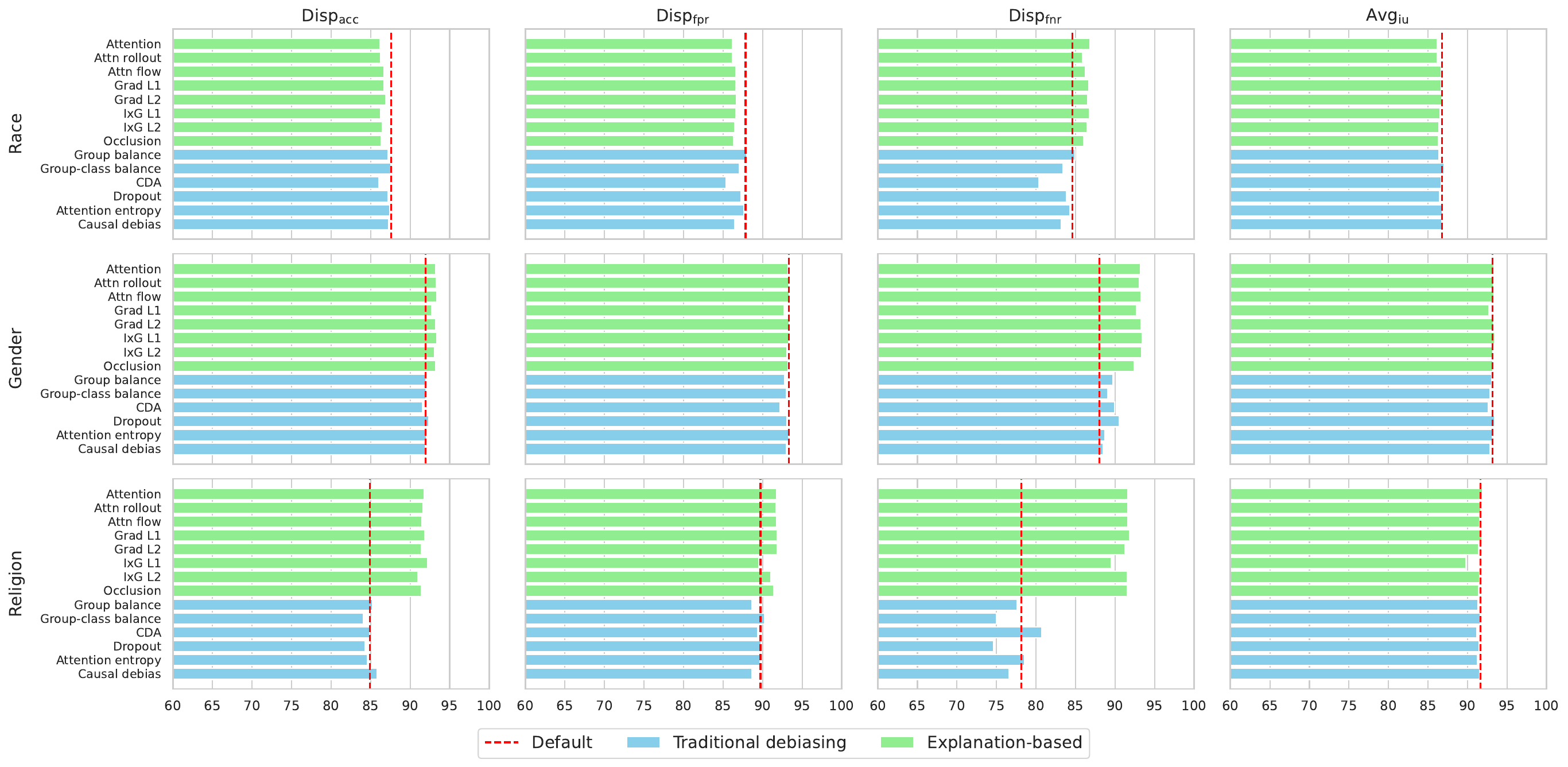}
    \caption{Harmonic mean between accuracy and fairness for established debiasing methods and explanation-based methods for BERT on Civil Comments. A higher score indicates better balance between model performance and fairness.}
    \label{fig:appendix harmonic mean bert civil}
\end{figure}

\begin{figure}
    \centering
    \includegraphics[width=\linewidth]{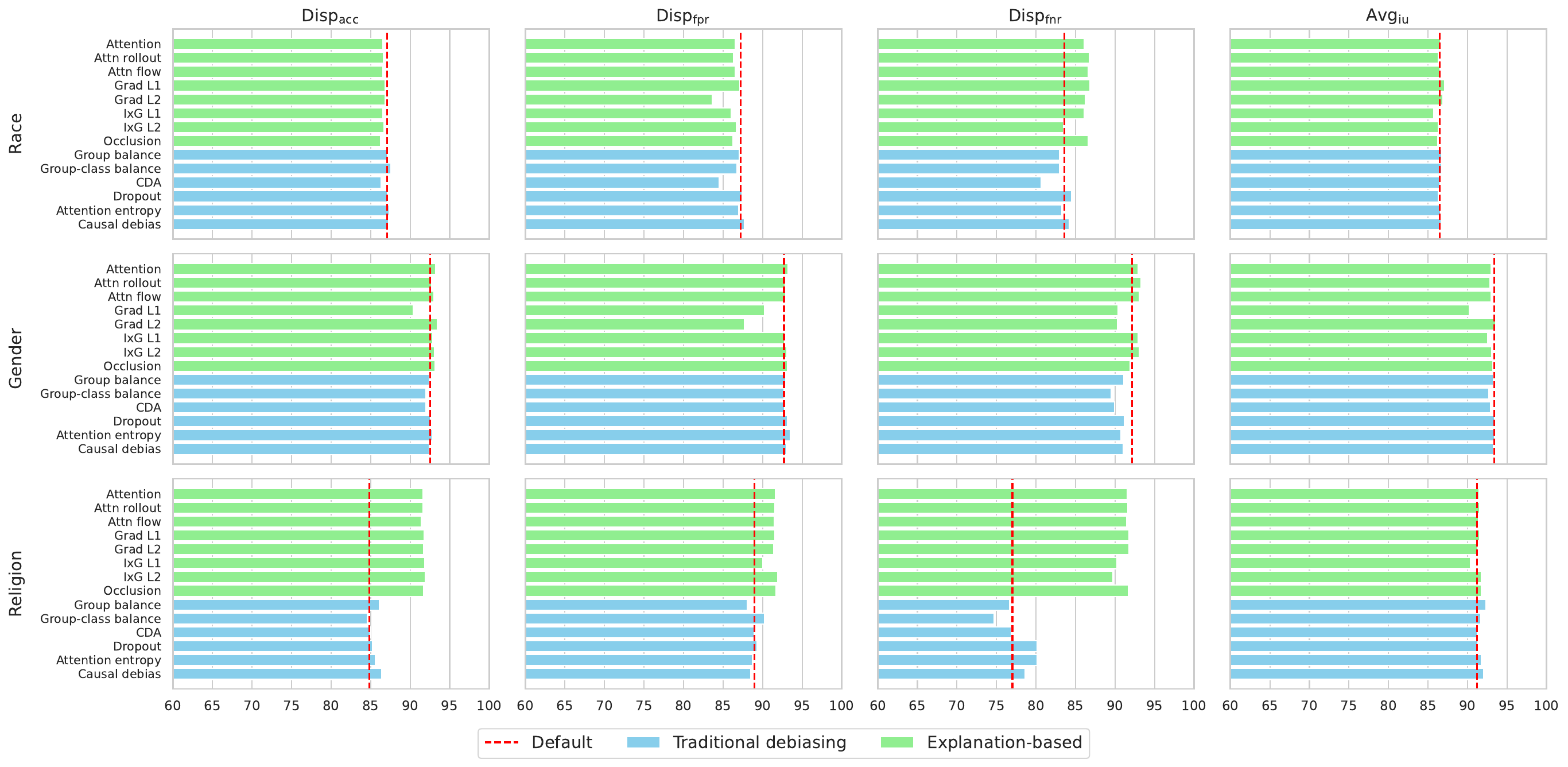}
    \caption{Harmonic mean between accuracy and fairness for established debiasing methods and explanation-based methods for RoBERTa on Civil Comments. A higher score indicates better balance between model performance and fairness.}
    \label{fig:appendix harmonic mean roberta civil}
\end{figure}

\begin{figure}
    \centering
    \includegraphics[width=\linewidth]{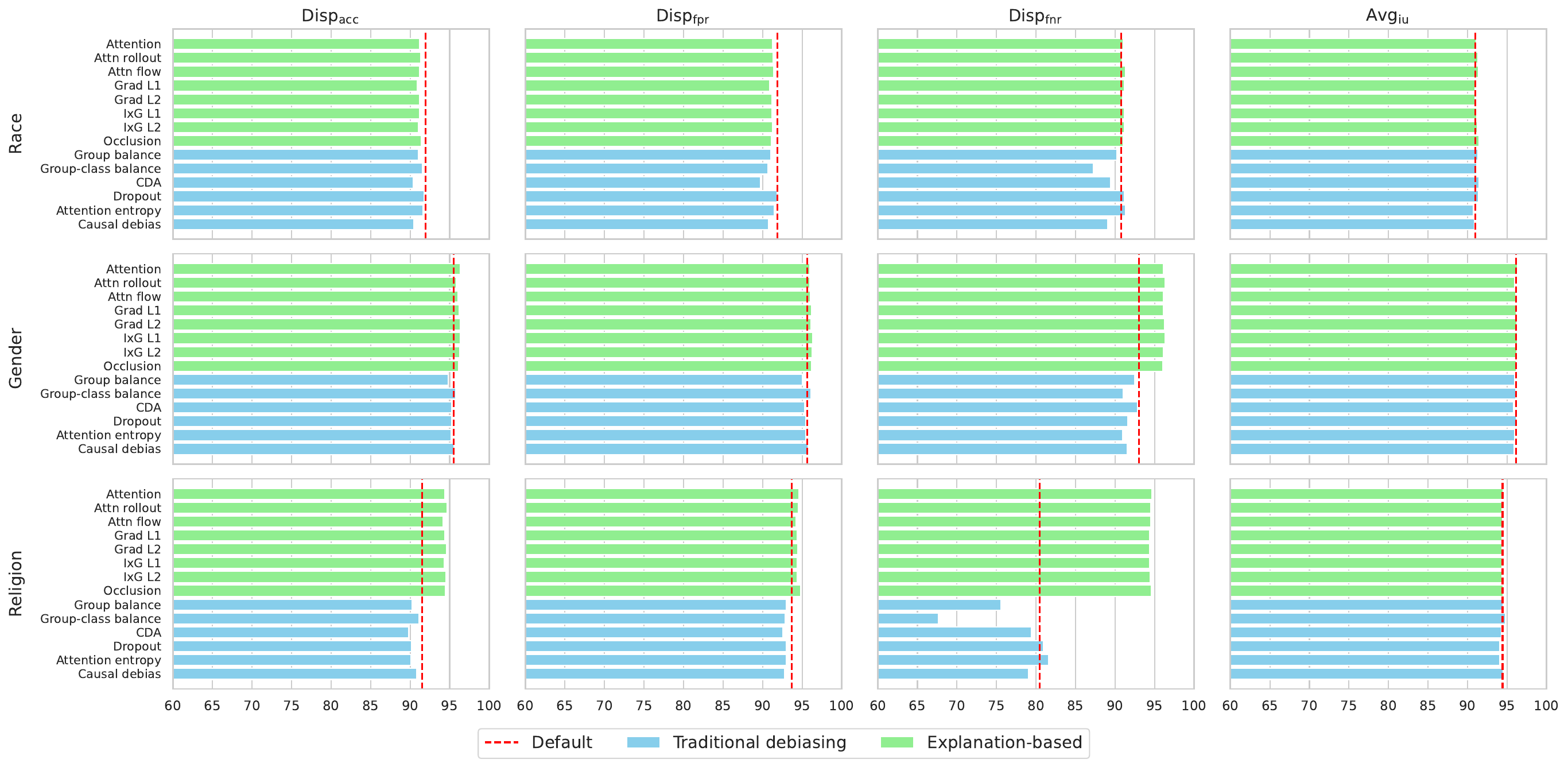}
    \caption{Harmonic mean between accuracy and fairness for established debiasing methods and explanation-based methods for BERT on Jigsaw. A higher score indicates better balance between model performance and fairness.}
    \label{fig:appendix harmonic mean bert jigsaw}
\end{figure}

\begin{figure}
    \centering
    \includegraphics[width=\linewidth]{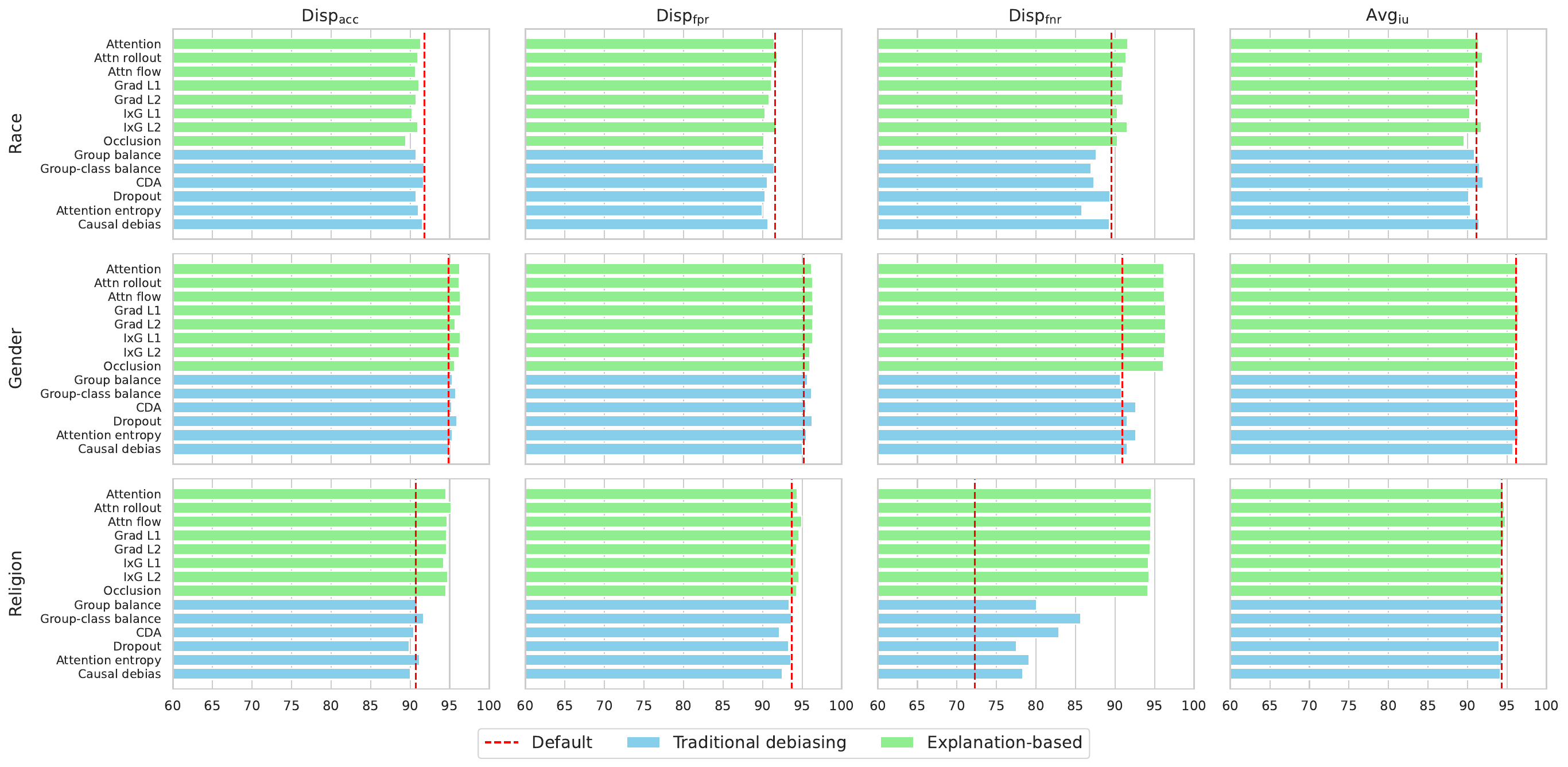}
    \caption{Harmonic mean between accuracy and fairness for established debiasing methods and explanation-based methods for RoBERTa on Jigsaw. A higher score indicates better balance between model performance and fairness.}
    \label{fig:appendix harmonic mean roberta jigsaw}
\end{figure}

The complete bias mitigation results are presented in Figures~\ref{fig:appendix bias mitigation civil bert}, \ref{fig:appendix bias mitigation civil roberta}, \ref{fig:appendix bias mitigation jigsaw bert}, \ref{fig:appendix bias mitigation jigsaw roberta}.
The findings are in line with conclusions from the main paper, that explanation-based debiasing can effectively reduce model biases across different fairness metrics, bias types, models, and datasets. 
In addition, the accuracy-fairness harmonic mean results shown in Figures~\ref{fig:appendix harmonic mean bert civil},~\ref{fig:appendix harmonic mean roberta civil},~\ref{fig:appendix harmonic mean bert jigsaw}, \ref{fig:appendix harmonic mean roberta jigsaw} demonstrate that explanation-based debiasing achieves comparable or superior balance between fairness and task performance than default models and traditional debiasing approaches.

We additionally report the results of Integrated Gradients for bias mitigation in Table~\ref{tab: intgrad bias mitigation}. Similar to other explanation methods, IntGrad-based debiasing achieves substantial bias reduction and maintains a good balance between fairness and task performance in $\text{Disp}_\text{fnr}$ and $\text{Avg}_\text{iu}$.

\begin{table*}[h]
    \centering
    \caption{Results of mitigating race bias in BERT models using Intgrad explanations on Civil Comments.
    For consistency with accuracy, fairness results are reported as $100-\{\text{Disp}_\text{acc}, \text{Disp}_\text{fpr}, \text{Disp}_\text{fnr}, \text{Avg}_\text{iu}\}$, so that higher values indicate better debiasing performance.
Each column corresponds to models selected by maximizing the fairness-balanced metric with respect to the indicated bias metric.
H-Mean indicates harmonic mean between fairness and accuracy.
     \textcolor{forestgreen}{Green} (\textcolor{red}{red}) indicates the results are \textcolor{forestgreen}{better} (\textcolor{red}{worse}) than the default models.
     \\}
    \resizebox{\textwidth}{!}{
    \begin{tabular}{lcccccccccccc}
    \toprule
      \multirow{2}{*}{Method}   &  \multicolumn{3}{c}{$\text{Disp}_\text{{acc}}$} & \multicolumn{3}{c}{$\text{Disp}_\text{{fpr}}$} & \multicolumn{3}{c}{$\text{Disp}_\text{{fnr}}$} & \multicolumn{3}{c}{$\text{Avg}_\text{{iu}}$}\\
      
         & Acc & Fairness & H-Mean & Acc & Fairness & H-Mean & Acc & Fairness & H-Mean & Acc & Fairness & H-Mean \\
         \midrule
         IntGrad L1 & \textcolor{red}{78.55} & \textcolor{red}{96.66} & \textcolor{red}{86.67} & \textcolor{red}{78.55} & \textcolor{red}{96.66} & \textcolor{red}{86.67} & \textcolor{red}{77.98} & \textcolor{forestgreen}{97.86} & \textcolor{forestgreen}{86.8} & \textcolor{red}{78.37} & \textcolor{forestgreen}{97.02} & \textcolor{red}{86.7} \\
        IntGrad L2 & \textcolor{red}{77.85} & \textcolor{red}{96.58} & \textcolor{red}{86.21} & \textcolor{red}{77.71} & \textcolor{red}{96.33} & \textcolor{red}{86.02} & \textcolor{red}{77.85} & \textcolor{forestgreen}{96.58} & \textcolor{forestgreen}{86.21} & \textcolor{red}{78.09} & \textcolor{forestgreen}{97.1} & \textcolor{red}{86.56} \\
         \midrule
         Default & 78.97 & 98.37 & 87.61 & 78.97 & 98.96 & 87.84 & 78.97 & 91.15 & 84.62 & 78.97 & 96.36 & 86.8 \\
         \bottomrule
    \end{tabular}
    }
    \label{tab: intgrad bias mitigation}
\end{table*}

\section{Generalization of Findings}
\label{appendix:generalization}
To demonstrate the generalizability of our findings, we present results under additional setups that vary in task, model alignment type, and sensitive token vocabulary.
We observe similar results across these setups, suggesting that our findings generalize well beyond the main study conditions.

\begin{figure}[h]
    \centering
    \includegraphics[width=0.75\linewidth]{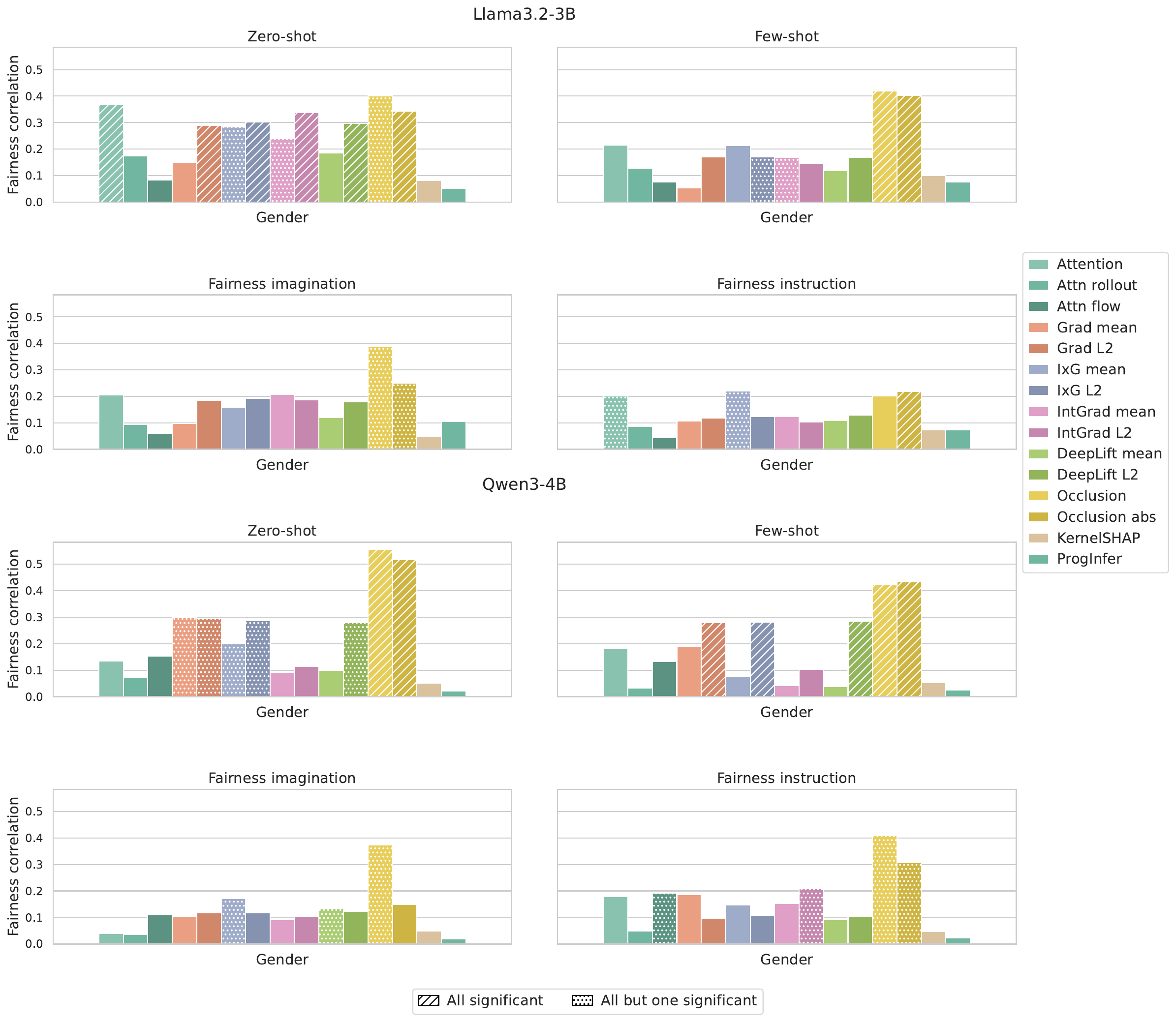}
    \caption{Fairness correlation results for race bias on Twitter Sentiment with Llama3.2-3B-Instruct and Qwen3-4B. Higher values indicate that the method is more effective and reliable in detecting biased predictions at inference time.}
    \label{fig:sentiment_analysis}
\end{figure}

\paragraph{Task} We evaluated explanation-based bias detection (RQ1) on an additional task, namely sentiment analysis, using the Twitter Sentiment dataset\footnote{\href{https://huggingface.co/datasets/shukdevdatta123/twitter_sentiment_preprocessed}{https://huggingface.co/datasets/shukdevdatta123/twitter\_sentiment\_preprocessed}}. 
Specifically, we selected 1000 gender-related examples (500 referencing males and 500 referencing females) and ran explanation-based bias detection on them. 
In Figure~\ref{fig:sentiment_analysis} we report the results on Llama3.2-3B-Instruct and Qwen3-4B models.

In Figure~\ref{fig:sentiment_analysis}, we observe patterns in the sentiment analysis task that are similar to those in our main study: certain explanation methods (e.g., occlusion-based and L2-based approaches) can still achieve high fairness correlation scores. This suggests that our findings could extend beyond the hate speech detection task.

\paragraph{Model Alignment Type}

\begin{figure}[h]
    \centering
    \includegraphics[width=0.75\linewidth]{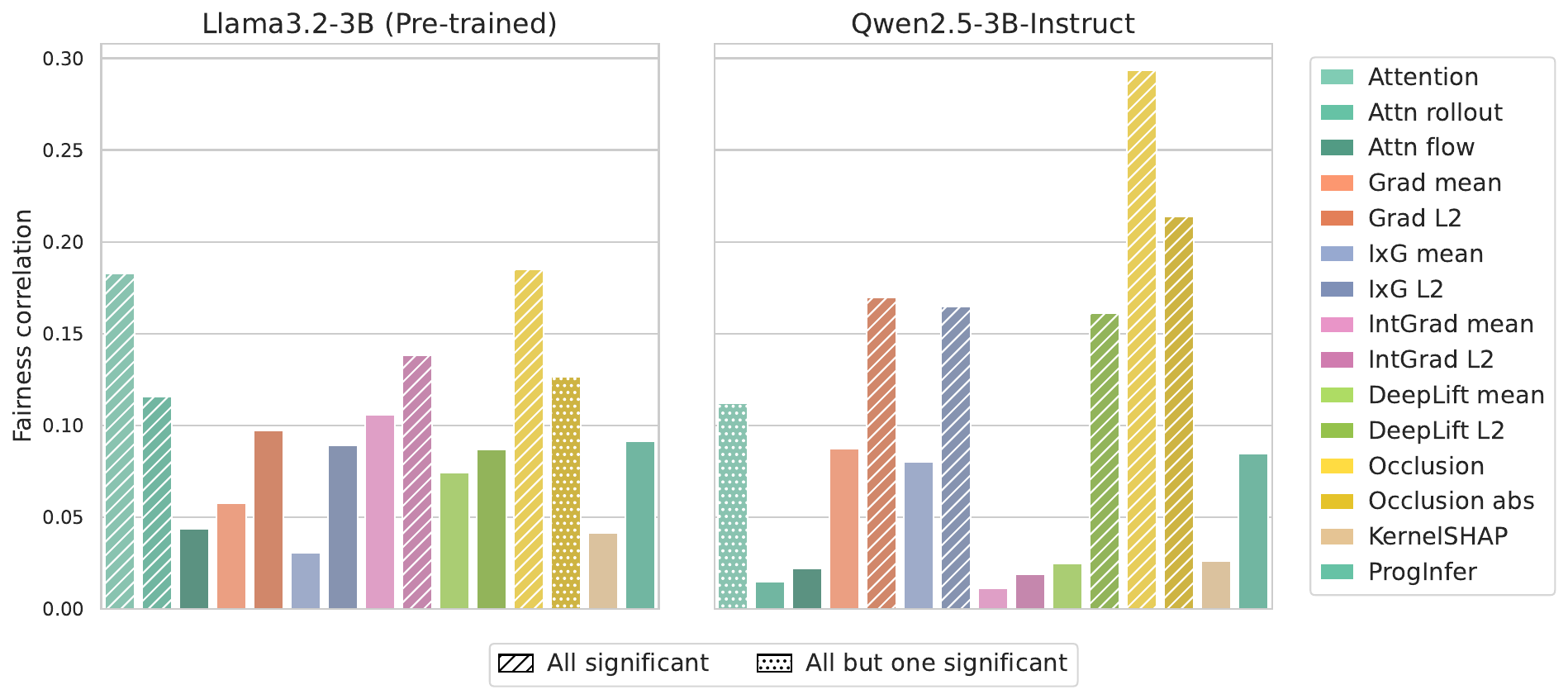}
    \caption{Fairness correlation results for race bias on Civil Comments with Llama3.2-3B and Qwen2.5-3B-Instruct. Both models are differently aligned from models in our main experiments. Higher values indicate that the method is more effective and reliable in detecting biased predictions at inference time.}
    \label{fig:differently_aligned_models}
\end{figure}

We extended our experiments to additional LLMs with different alignment methods. Specifically, we evaluated explanation-based bias detection (RQ1) on two differently aligned LLMs: Llama3.2-3B (pre-trained only, non-instruct, used with few-shot prompting) and Qwen2.5-3B-Instruct (instruction-tuned only). Neither model is aligned to human values, which differs from the models used in our main study. 
The results are computed for race bias on Civil Comments and shown in Figure~\ref{fig:differently_aligned_models}.

We observe that certain explanation methods, such as Occlusion, consistently achieve high fairness correlations. 
This suggests that our findings generalize across LLMs with different alignment settings.

\paragraph{Sensitive Token Vocabulary}

In practice, it is often unrealistic to exhaustively enumerate all vocabulary items that may encode sensitive attributes. To assess the applicability of our findings under such conditions, we analyze how varying the coverage of sensitive tokens affects bias detection and mitigation. Specifically, we focus on gender bias and use a small subset of gendered pronouns (“he/his/him” and “she/her”) as sensitive tokens, while computing fairness metrics with the full gender-related vocabulary (222 words per gender). This setup simulates real-world scenarios where the sensitive vocabulary cannot be fully enumerated.

\begin{figure}[h]
    \centering
    \includegraphics[width=0.75\linewidth]{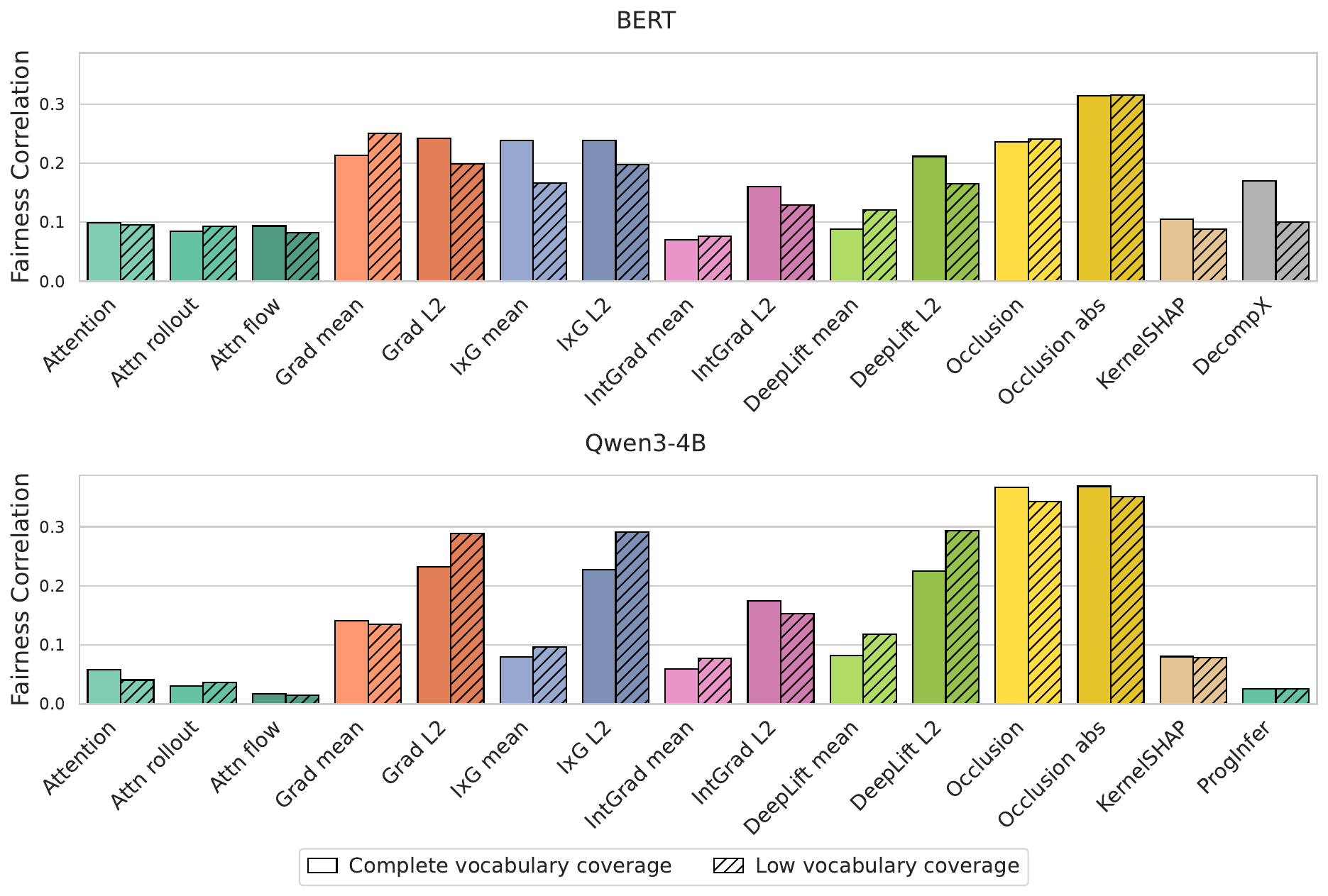}
    \caption{Fairness correlation results when using a reduced sensitive token vocabulary for reliance computation. Results are reported for gender bias on the Civil Comments dataset. Fairness metrics are still computed using the full vocabulary. The reduced vocabulary size has only a marginal effect on fairness correlations.}
    \label{fig:vocab_correlation}
\end{figure}

\begin{figure}[h]
    \centering
    \includegraphics[width=0.75\linewidth]{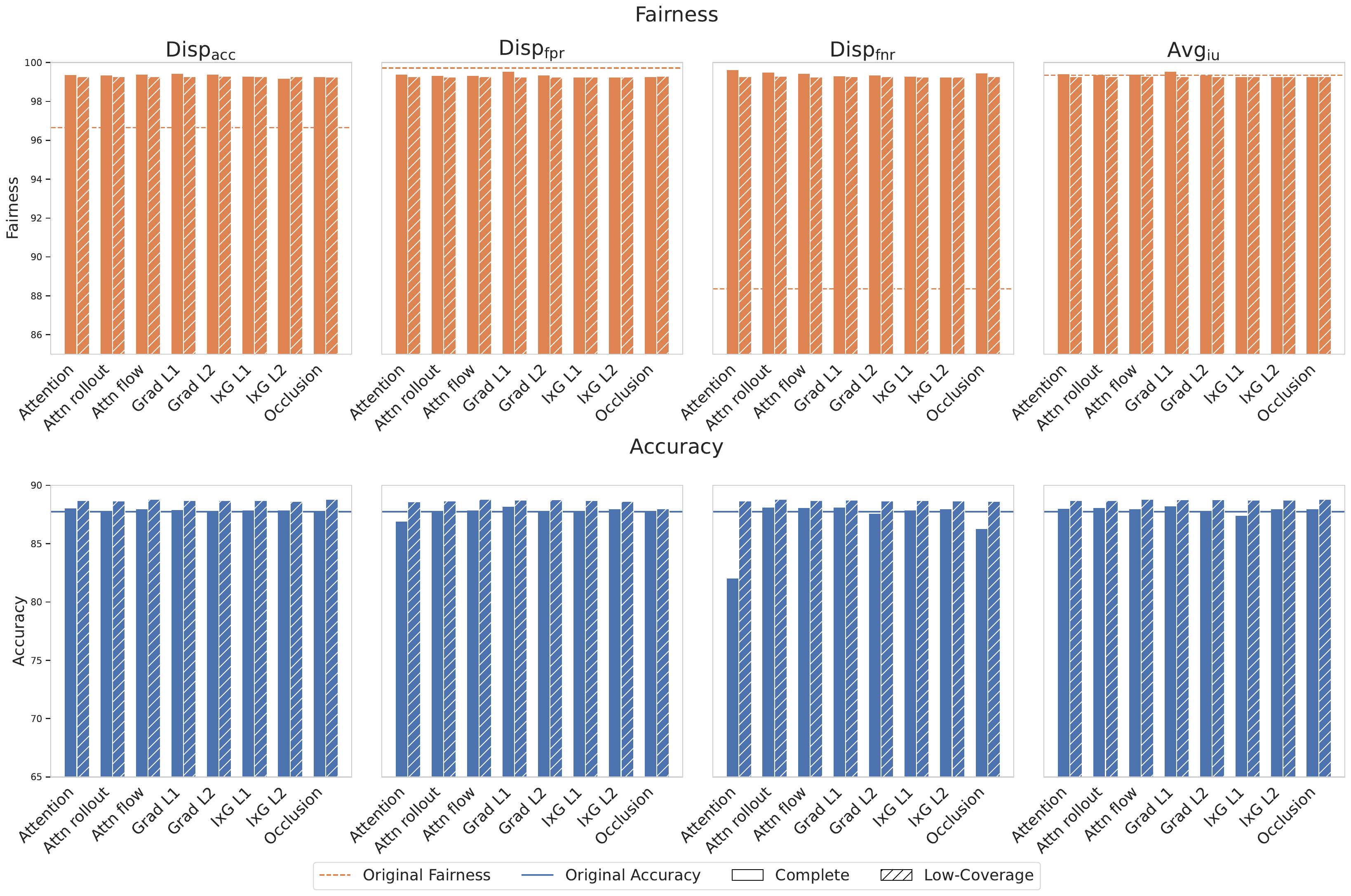}
    \caption{Fairness and accuracy results for gender bias mitigation with a reduced sensitive token vocabulary.
Each column corresponds to models selected by maximizing the fairness–balance metric with respect to the indicated bias metric. 
Using an incomplete vocabulary yields slightly worse debiasing performance than using the complete vocabulary, while preserving task performance more effectively. 
Overall, the impact of reduced vocabulary coverage is minimal.}
    \label{fig:vocab_debiasing}
\end{figure}

As shown in Figures~\ref{fig:vocab_correlation} and~\ref{fig:vocab_debiasing}, reduced vocabulary coverage has minimal impact on explanation-based bias detection and mitigation performance. This result is reassuring, suggesting that explanation methods remain effective in more complex, realistic settings where exhaustive sensitive token coverage is infeasible.

\section{Faithfulness as an Indicator of Bias Detection Ability}
\label{appendix:faithfulness}
What factors influence the reliability of explanations in detecting bias? In this section, we examine the relationship between explanation faithfulness and their ability to identify bias, reflected by fairness correlation scores in RQ1.
We assess the faithfulness of explanation methods using two perturbation-based metrics: comprehensiveness and sufficiency AOPC (Area Over the Perturbation Curve; \citealp{deyoung-etal-2020-eraser}), computed by masking 5\%, 10\%, 20\%, and 50\% of the input tokens.
For substitution, we use the [MASK] token in BERT and the [PAD] token in Qwen3-4B.
Higher comprehensiveness and lower sufficiency scores indicate more faithful explanations.

\begin{figure}[h]
    \centering
    \includegraphics[width=0.5\linewidth]{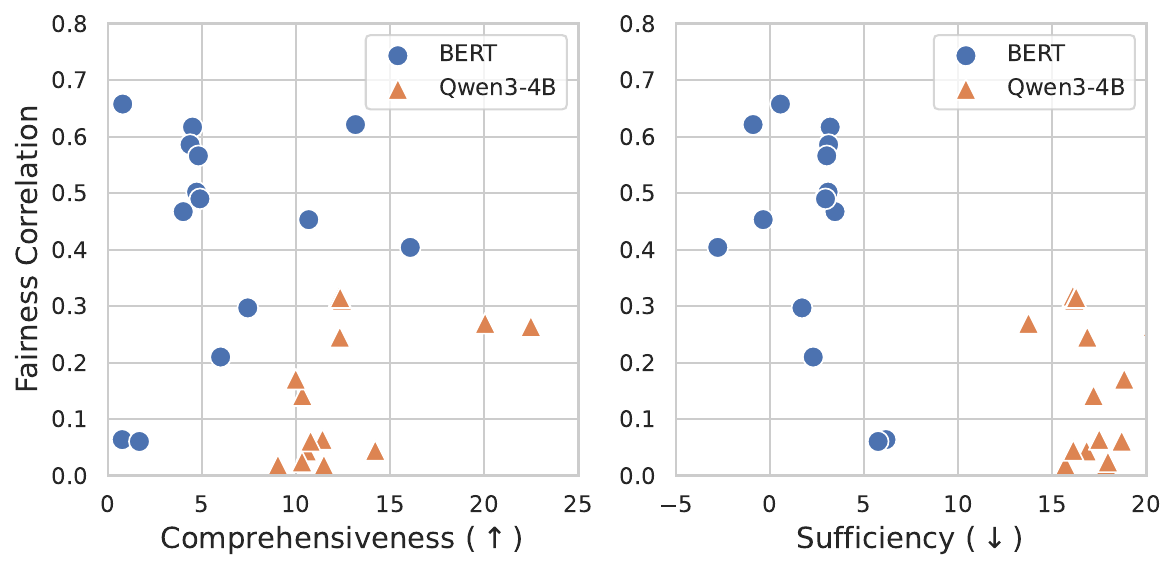}
    \caption{Faithfulness and fairness correlation results of different explanation methods. No clear relationship between explanation faithfulness and their bias detection ability is observed. Each point represents the faithfulness and fairness correlation of one explanation method applied to default/zero-shot models.}
    \label{fig:faithfulness}
\end{figure}

Our results on race bias in Civil Comments (Figure~\ref{fig:faithfulness} and Table~\ref{tab:faithfulness}) reveal no clear link between faithfulness and fairness correlation of explanations. 
In particular, mean-based explanations may achieve better faithfulness scores than their L2-based counterparts, yet they consistently perform significantly worse in identifying bias.
We attribute this discrepancy to two key differences between the faithfulness metrics and our fairness correlation measure.
First, faithfulness evaluates attribution scores across all input tokens, whereas our fairness correlation measure only considers sensitive token reliance.
Second, perturbation-based faithfulness assesses the impact of masking tokens on model predictions, while our individual unfairness metric compares predictions when one social group is substituted for another.
Taken together, these findings suggest that explanation faithfulness is not a reliable indicator of bias detection ability. 
We therefore do not recommend selecting explanation methods for fairness on the basis of faithfulness results alone.

\begin{table}
       \centering
     \caption{Faithfulness results of different explanation methods on BERT and Qwen3-4B models.\\}
     \resizebox{\textwidth}{!}{
     \begin{tabular}{l|ccc|ccc}
     \\
     \toprule
      Explanation & Comp. ($\uparrow$) & Suff. ($\downarrow$) & Fairness Correlation ($\uparrow$) & Comp. ($\uparrow$) & Suff. ($\downarrow$) & Fairness Correlation ($\uparrow$)\\
     \midrule
     & \multicolumn{3}{c|}{BERT} & \multicolumn{3}{c}{Qwen3-4B} \\
     \midrule
Attention & 4.50 & 3.20 & 0.62 & 10.34 & 17.20 & 0.14 \\
Attn rollout & 4.37 & 3.11 & 0.59 & 9.04 & 15.70 & 0.02 \\
Attn flow & 4.01 & 3.46 & 0.47 & 10.57 & 16.82 & 0.04 \\
Grad L2 & 4.82 & 2.99 & 0.50 & 12.30 & 16.09 & 0.32 \\
Grad mean & 0.77 & 6.16 & 0.06 & 11.41 & 17.50 & 0.06 \\
DeepLift L2 & 4.72 & 3.09 & 0.50 & 12.44 & 16.17 & 0.31 \\
DeepLift mean & 1.68 & 5.75 & 0.06 & 10.78 & 18.69 & 0.06 \\
IxG L2 & 4.89 & 2.95 & 0.49 & 12.35 & 16.27 & 0.32 \\
IxG mean & 7.44 & 1.70 & 0.30 & 9.99 & 18.82 & 0.17 \\
IntGrad L2 & 4.81 & 3.02 & 0.57 & 12.33 & 16.86 & 0.25 \\
IntGrad mean & 10.68 & -0.36 & 0.45 & 14.21 & 16.12 & 0.04 \\
Occlusion & 13.16 & -0.90 & 0.62 & 20.05 & 13.73 & 0.27 \\
Occlusion abs & 0.79 & 0.56 & 0.66 & 22.48 & 20.36 & 0.26 \\
KernelSHAP & 5.99 & 2.30 & 0.21 & 11.49 & 17.86 & 0.02 \\
DecompX & 16.08 & -2.77 & 0.40 & - & - & - \\
ProgInfer & - & - & - & 10.32 & 17.96 & 0.025 \\
\bottomrule
\end{tabular}}
     \label{tab:faithfulness}
\end{table}

\section{Explanations for Human Fairness Auditing}
\label{appendix:human fairness auditing}

In addition to evaluating input-based explanations as automatic bias detectors, we also examine their ability to support human auditing of biased model predictions.
To this end, we conducted a small-scale human study following the experimental protocol described below.

We randomly sample 48 correctly predicted examples related to race bias from Civil Comments (4*6=24 from BERT and 24 from Qwen3-4B, balanced across all group–class categories). We evaluate six explanation methods: three directed methods (Occlusion, IxG mean, KernelSHAP) and three undirected methods (Occlusion abs, IxG L2, Attention), chosen to cover diverse explanation families and performance characteristics observed in RQ1.

For each example, annotators first read the input text and provide their own toxicity prediction. They are then shown either the three directed explanations or the three undirected explanations for that example. For each explanation, annotators give two ratings on a 1–5 scale: one assessing its interpretability, and another evaluating how much the model's prediction appears to rely on race-related bias or stereotypes, based on the information conveyed by the explanation.

After annotation, we collect annotators' perceived bias ratings and measure their correlation with the ground-truth individual unfairness scores. Higher fairness correlation indicates greater effectiveness of an explanation method for human fairness auditing.

\begin{table}[h]
\caption{Fairness correlation of explanation methods for human fairness auditing and their interpretability.
Best scores in each explanation type are marked in \textbf{bold}.
Higher fairness correlation scores indicate that explanations can better assist humans to detect bias.
\\}
\resizebox{\linewidth}{!}{
    \centering
    \begin{tabular}{lcccccc}
    \toprule
       & \multicolumn{3}{c}{Undirected}  & \multicolumn{3}{c}{Directed}  \\
       & Attention & IxG L2 & Occlusion abs & KernelSHAP & IxG mean & Occlusion \\
       \midrule
       Fairness correlation & 0.402  & 0.123 & \textbf{0.433} & 0.254  & -0.078 & \textbf{0.342} \\
       Interpretability & 2.256 & \textbf{3.179} & 2.920 & 2.518 & 2.439 & \textbf{2.780} \\
    \bottomrule
    \end{tabular}
    }
    \label{tab:human_study}
\end{table}

\begin{table}[h]
\caption{Fairness correlation of explanation methods for human fairness auditing under different conditions. 
Higher fairness correlation scores indicate that explanations can better assist humans to detect bias. 
\textcolor{forestgreen}{Green} (\textcolor{red}{red}) indicates the results are \textcolor{forestgreen}{better} (\textcolor{red}{worse}) than the baseline (all examples).
\\}
\resizebox{\linewidth}{!}{
    \centering
    \begin{tabular}{lcccccc}
    \toprule
       & \multicolumn{3}{c}{Undirected}  & \multicolumn{3}{c}{Directed}  \\
       & Attention & IxG L2 & Occlusion abs & KernelSHAP & IxG mean & Occlusion \\
       \midrule
       All examples & 0.402  & 0.123 & 0.433 & 0.254  & -0.078 & 0.342 \\
       Correct predictions Only & \textcolor{forestgreen}{0.572} & \textcolor{forestgreen}{0.202} & \textcolor{forestgreen}{0.602} & \textcolor{red}{0.217} & \textcolor{forestgreen}{0.029} & \textcolor{forestgreen}{0.451} \\
       Toxic examples only & \textcolor{forestgreen}{0.637} &  \textcolor{red}{0.118} & \textcolor{red}{0.374} & \textcolor{red}{0.231} & \textcolor{forestgreen}{0.134} & \textcolor{red}{0.315} \\
       High interpretability (score $\geq$ 3) & \textcolor{red}{0.138} & \textcolor{forestgreen}{0.227} & \textcolor{red}{0.288} & \textcolor{red}{0.154} & \textcolor{forestgreen}{-0.043} & \textcolor{forestgreen}{0.404}\\
    \bottomrule
    \end{tabular}
    }
    \label{tab:human_study_comparison}
\end{table}

Table~\ref{tab:human_study} shows that certain explanation methods, such as Attention, Occlusion, and Occlusion abs, achieve high fairness correlations, suggesting that they can effectively assist humans in detecting bias. 
Across explanation types, undirected explanations appear more helpful. 
For example, Occlusion and Occlusion abs produce the same attribution patterns that differ only in directional encoding, yet participants were better able to identify bias using the undirected variant (Occlusion abs). 
Furthermore, while some methods support both human and automatic bias detection consistently (e.g., Attention and Occlusion abs), others, such as IxG L2, show substantial gaps in performance. This highlights a potential discrepancy between how humans interpret explanations and how our automatic pipeline evaluates them.

We also observe that high interpretability alone does not guarantee better support for human fairness auditing: methods with strong interpretability scores (e.g., IxG L2) still fail to effectively help humans detect bias.
Finally, 4 out of 6 annotators reported that undirected explanations helped them detect bias more effectively, noting that they introduce less noise and make annotation easier.

Table~\ref{tab:human_study_comparison} further analyzes explanation-assisted human fairness auditing under different conditions. 
For correctly predicted examples, explanations generally provide stronger support for bias detection. 
However, the effects of label type and explanation interpretability appear more nuanced and vary across methods. 
Overall, these results suggest that explanation-assisted human fairness auditing is a promising and interesting direction for future work and warrants further investigation.

\section{Hybrid Bias Mitigation Techniques}
\label{appendix:hybrid debiasing}
we conducted preliminary experiments that combine several pre-processing techniques (group balance, group–class balance, and CDA) with an effective explanation-based debiasing method (IxG L1/L2). The resulting individual fairness outcomes, along with comparisons to traditional and explanation-only methods, are presented in Table~\ref{tab:hybrid_debiasing}.

\begin{table}[h]
    \centering
        \caption{Each cell shows the $\text{Avg}_\text{iu}$ score after applying a combination of pre-processing and explanation-based debiasing methods. Lower values indicate reduced bias. Values in parentheses denote the change relative to using only the corresponding traditional/explanation-based method, where \textcolor{forestgreen}{negative values} indicate improved debiasing.
    We observe that hybrid approaches consistently achieve stronger bias mitigation than either method used in isolation.\\}
    \begin{tabular}{cccc}
      \toprule
      \multicolumn{4}{c}{Race} \\
      \midrule
    & Group balance & Group-class balance & CDA \\
      IxG L1 & 0.012 (\textcolor{forestgreen}{-4.492}/\textcolor{forestgreen}{-1.461}) & 0.000 (\textcolor{forestgreen}{-3.048}/\textcolor{forestgreen}{-1.473}) & 0.001 (\textcolor{forestgreen}{-0.547}/\textcolor{forestgreen}{-1.473}) \\
      IxG L2 & 2.162 (\textcolor{forestgreen}{-2.342}/\textcolor{forestgreen}{-0.598}) & 2.110 (\textcolor{forestgreen}{-0.938}/\textcolor{forestgreen}{-0.650}) & 0.210 (\textcolor{forestgreen}{-0.338}/\textcolor{forestgreen}{-2.550}) \\
      \midrule
      \multicolumn{4}{c}{Gender} \\
      \midrule
          & Group balance & Group-class balance & CDA \\
      IxG L1 & 0.005 (\textcolor{forestgreen}{-0.594}/\textcolor{forestgreen}{-0.548}) & 0.001 (\textcolor{forestgreen}{-0.836}/\textcolor{forestgreen}{-0.552}) & 0.001 (\textcolor{forestgreen}{-0.488}/\textcolor{forestgreen}{-0.551}) \\
      IxG L2 & 0.308 (\textcolor{forestgreen}{-0.291}/\textcolor{forestgreen}{-0.331}) & 0.546 (\textcolor{forestgreen}{-0.292}/\textcolor{forestgreen}{-0.093}) & 0.368 (\textcolor{forestgreen}{-0.122}/\textcolor{forestgreen}{-0.271}) \\
      \bottomrule
        
    \end{tabular}

    \label{tab:hybrid_debiasing}
\end{table}


We observe that the hybrid method achieves better bias mitigation effects than using each debiasing method alone, with consistent improvements for both race and gender bias. 
Based on this, we believe exploring hybrid methods for more effective bias mitigation could be a promising future direction.

\section{Fairness Correlations in Explanation-Debiased Models}
\label{appendix: fair-washing}

\begin{figure}
    \centering
    \includegraphics[width=0.75\linewidth]{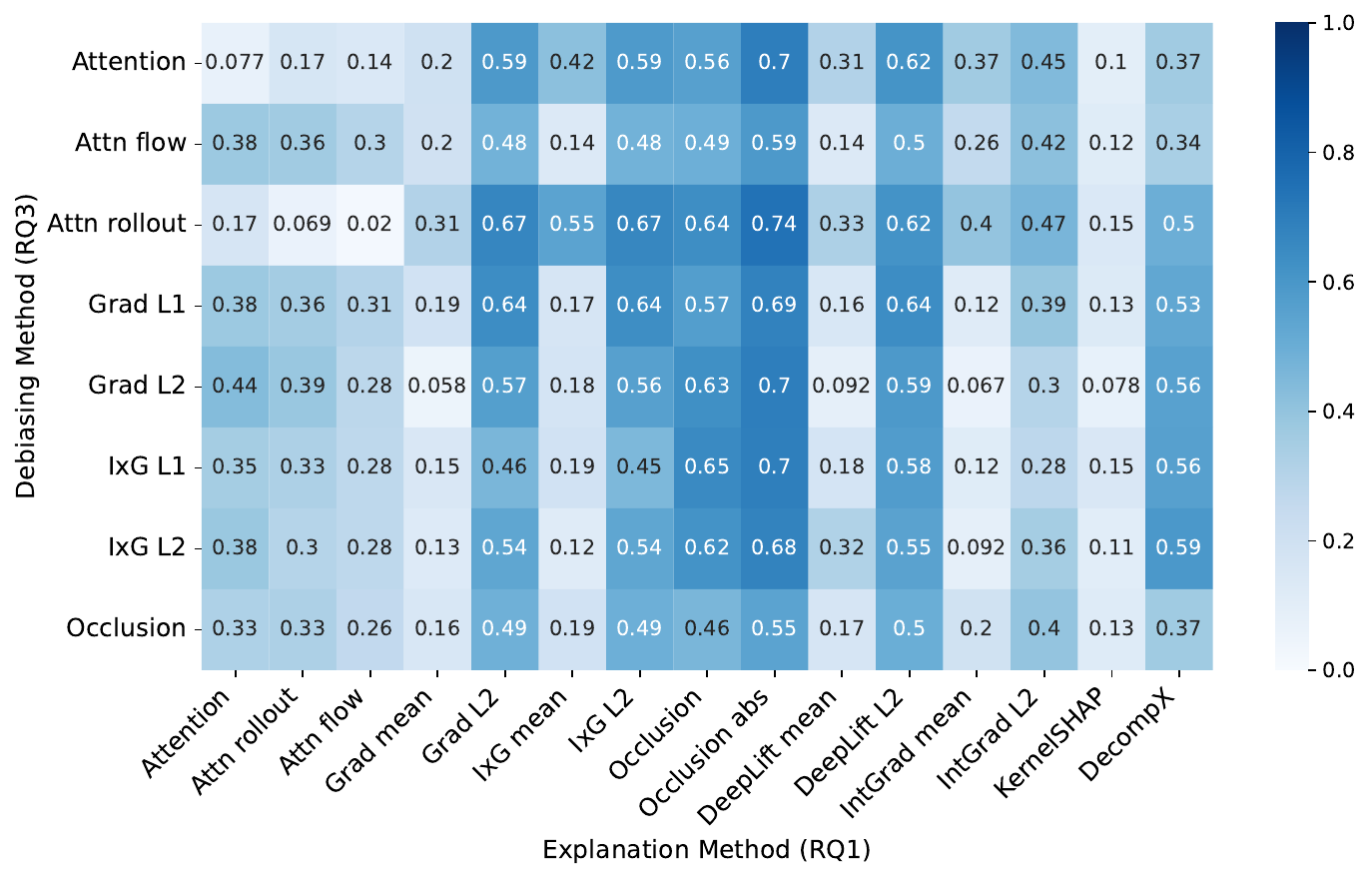}
    \caption{Fairness correlation results on BERT models with race bias mitigated through explanation-based methods on Civil Comments.}
    \label{fig:appendix fairwashing race}
\end{figure}

Figure \ref{fig:appendix fairwashing race} presents the fairness correlation scores computed on explanation-debiased models. 
We find that Grad L2, IxG L2, DeepLift L2, and Occlusion-based explanations still show strong bias mitigation ability in the debiased models.

\section{Limitations and Future Work}
\label{appendix:limitations}

Our study has several limitations that we acknowledge and aim to address in future work.

First, as the first quantitative investigation of this topic, our study focuses solely on hate speech detection and uses a limited set of experimental setups. Although the results are consistent across these setups and preliminary experiments (Appendix~\ref{appendix:generalization}) suggest good generalization across tasks, models and sensitive token vocabulary, broader validation is still needed. Future work could extend this evaluation to additional domains and high-stakes applications.

Second, several findings are derived under the specific experimental setups used in this work. For instance, in RQ2, we conclude that the proposed attribution-based metrics are not reliable fairness indicators. However, it remains possible that other metrics could be effective. Similarly, our fairness-balanced metric in RQ3 may not be the optimal validation strategy in all settings. As it is infeasible to exhaustively enumerate and evaluate all potential configurations, we believe our conclusions nonetheless offer valuable guidance and highlight important methodological considerations for the community.

Third, our work focuses on evaluating standalone explanation-based strategies for improving fairness. Ensembles of multiple explanation methods, or hybrid approaches that combine explanation techniques with established debiasing methods, may yield better outcomes. Additionally, incorporating human oversight may further enhance the effectiveness and robustness of explanation-based fairness auditing. Our preliminary experiments show promising results in using hybrid debiasing techniques~\ref{appendix:hybrid debiasing}, and demonstrates the possibility for human fairness auditing based on explanations~\ref{appendix:human fairness auditing}.
Based on that, we believe that a systematic investigation of such hybrid or human-in-the-loop approaches represents an interesting avenue for future work.

Fourth, we do not identify any explanation method that consistently outperforms others across all research questions, which prevents us from offering a single recommendation. We therefore encourage future researchers to choose explanation methods that align with their specific tasks and constraints. Future work could further investigate why certain methods are better suited to particular settings and, ideally, develop practical guidelines for selecting effective methods without requiring extensive empirical comparisons.

Finally, although we consider both group and individual fairness, this work provides a more in-depth analysis of individual fairness (in RQ1 and RQ2), driven by the conceptual alignment between input-based explanations and individual fairness notions. We encourage future work to more thoroughly examine how explanation methods relate to group fairness.

\section{LLM Usage}
\label{appendix: llm usage}
Apart from the models evaluated in our experiments and analyses, we used LLMs (ChatGPT) solely to polish the writing in this work.

\end{document}